\RequirePackage{fix-cm}
\documentclass[smallextended]{svjour3}       
\smartqed  
\usepackage{amssymb, amsmath, comment}
\usepackage{float,epsfig,graphicx,subfig, booktabs,pgfplotstable}
\usepackage{mathptmx} 
\usepackage{kotex}

\newcommand{\inputSingle}[3]{$L = #1[m]$, $c = #2[kg\cdot m/s]$, $\dot{\theta}^0 = #3[rad/s]$}

\newcommand{\inputDouble}[4]{$L_1 = #1[m]$, $L_2 = #2[m]$, $\dot{\theta}^0_1 = #3[rad/s]$,  $\dot{\theta}^0_2 = #4[rad/s]$}

\newcommand{\figsize}{0.459}
\newcommand{\figsmall}{0.45}
\newcommand{\ra}[1]{\renewcommand{\arraystretch}{#1}}

\newcommand{\yl}{y}
\newcommand{\yp}{\hat{y}} 

\newcommand{\M}{\mathcal{M}}
\newcommand{\A}{\mathcal{A}}

\newcommand{\Sfixed}{$S_{fixed}$}
\newcommand{\Sfull}{$S_{full}$}

\newcommand{\R}{\mathcal{R}^2}
\newcommand{\MSE}{\mathcal{E}_{mse}}

\newcommand{\q}{\mathbf{q}}
\newcommand{\ddq}{\ddot{\q}}

\newcommand{\Cq}{\mathbf{C_q}}
\newcommand{\CqT}{\mathbf{C}^T_{\q}}
\newcommand{\T}{\mathbf{T}\,}

\journalname{Multibody System Dynamics}

\begin{document}
\title{
Data-driven simulation for general purpose multibody dynamics using deep neural networks
}


\titlerunning{Data-driven multibody dynamics simulation} 

\author{
Hee-Sun Choi
\and
Junmo An
\and 
Jin-Gyun Kim
\and
Jae-Yoon Jung 
\and 
Juhwan Choi
\and
Grzegorz Orzechowski 
\and
Aki Mikkola
\and
Jin Hwan Choi 
}


\institute{
Hee-Sun Choi, Jin-Gyun Kim, Jin Hwan Choi (Corresponding author) 
\at Department of Mechanical Engineering, Kyung Hee University 1732, Deogyeong-daero, Giheung-gu, Yongin-si, Gyeonggi-do 17104, Republic of Korea \\
\email{mdefptheia@gmail.com, jingyun.kim@khu.ac.kr, jhchoi@khu.ac.kr}
\and
Junmo An, Jae-Yoon Jung  
\at Department of Industrial and Management Systems Engineering, Kyung Hee University 1732, Deogyeong-daero, Giheung-gu, Yongin-si, Gyeonggi-do 17104, Republic of Korea \\
\email{ajm9306@khu.ac.kr, jyjung@khu.ac.kr}             
\and
Juhwan Choi  
\at R\&D Center, FunctionBay, Inc. 5F, Pangyo Seven Venture Valley 1 danji 2 dong, 15, Pangyo-ro 228 beon-gil, Bundang-gu, Seongnam-si, Gyeonggi-do, Republic of Korea \\
\email{juhwan@functionbay.co.kr } 
\and
Grzegorz Orzechowski, Aki Mikkola  
\at Department of Mechanical Engineering, LUT University, Yliopistonkatu 34, 53850 Lappeenranta, Finland\\
\email{grzegorz.orzechowski@lut.fi, aki.mikkola@lut.fi}
}

\date{Received: date / Accepted: date}

\maketitle

\begin{abstract}
In this paper, a machine learning based simulation framework of general purpose multibody dynamics is introduced. The aim of the framework is to generate a well trained meta-model of multibody dynamics (MBD) systems. 
To this end, deep neural network (DNN) is employed to the framework so as to construct data based meta model representing multibody systems. Constructing well defined training data set with time variable is essential to get accurate and reliable motion data such as displacement, velocity, acceleration, and forces. As a result of the introduced approach, the meta-model provides motion estimation of system dynamics without solving the analytical equations of motion. The performance of the proposed DNN meta-modeling was evaluated to represent several MBD systems. 
\keywords{
Multibody dynamics \and 
Meta-model \and 
Deep neural network \and 
Feed forward network \and 
Data-driven simulation
}
\end{abstract}

\section{Introduction}
\label{sec:intro}
Using Machine Learning (ML) with big data is an important subject matter in science and engineering. This is because ML is effective to handle and interpret big data sets for the purpose of finding certain patterns from the data.

In particular, Deep Neural Network (DNN), which is based on an Artificial Neural Network (ANN) with multiple hidden layers between input and output layers allows to handle complex shapes with nonlinear functions with multi-dimensional input data. DNN has been successfully used in a large number of practical applications. Well trained neural network then provides precise pattern recognition based on data sets in real time.

These features, big data recognition and real time estimation of nonlinear functions, of ML approaches are attractive to dynamics and control engineers who are handling nonlinear system dynamics with real world data. There have been several previous studies on applying ML, DNN, or other big-data handling techniques to rigid multibody system problems. 
For example, Bayesian formulation \cite{Lanz06,BlaTorGim15,TinMisPetSch07}
in combination with Markov random field approximation, Kalman filter, or particle filter has been applied to various multibody dynamics (MBD) problems to handle noise data effectively in real-life applications, generate reliable modeling with efficient computational cost, estimate multibody system in probabilistic sense, or identify nonlinear parameters in governing equations. 
ML approaches \cite{LiWuTedTenTor19,TutBroWan12,AnsTupDatNeg18,LinHafQueFre10,HalErdBog09} such as regression methods, reinforcement learning algorithms, and surrogate models have also been employed.
Regression methods have many different types that can be performed in ML. In addition to the simple linear regression model, one can select and use techniques such as polynomial regression, support vector regression, decision tree regression, and random forest regression to suit a given problem. 
Based on the investigated input-label values, surrogate models perform a probabilistic estimate for an unknown objective function. This is an approach that uses an interpretable model to describe complex models. The most commonly used model in surrogate models is the Gaussian process. 
The proposed method has enhanced accuracy of prediction, especially in the long time scales, and increased computational efficiency in simulating dynamic response of multibody system. 
Moreover, neural networks \cite{AnsTupDatNeg18,KraCauMar18,FalMalMel11,MarZaaWhiTaj07,ByrFox17} have been suggested as effective alternatives to multibody dynamics simulation in comparison with conventional algorithms. The approaches have been proved to be fast and reliable to describe and predict characteristics of multibody systems.

It is important to note that previous studies \cite{Lanz06,LinHafQueFre10,HalErdBog09,KraCauMar18,FalMalMel11,MarZaaWhiTaj07,ByrFox17} are focused on particular MBD problems, mainly on contact, railways, vehicles, gaits, robotics, or tracking. Accordingly, a general MBD problem has not been introduced and analyzed through DNN technique. 

To address these shortcomings, this study introduces a procedure to generate a solver based on {\textit{DNN meta-model}} for {\textit{general}} purpose multibody system, which allows us to predict MBD with high accuracy in real time. 
Among the various ML methods, a supervised learning technique is used for the mathematical and/or numerical data set of the MBD model in the training process.
Data preparation and training process are called {\textit{off-line}} stage, and its trained result is known as {\textit{meta-model}}. 
Using the meta-model, the time varying results can be estimated such as displacement, velocity, and acceleration of the multibody system without directly solving the governing equations of MBD, and then this estimation process is called {\textit{on-line}} stage.
In particular, {\textit{the feed forward networks}} (FFN) with hidden layers and non-linear activation functions are employed among the various DNN methods since it can efficiently represent continuous functions.
Three representative MBD problems, single pendulum, double pendulum, and slider crank mechanism, were considered to evaluate the performance of the proposed DNN based meta-modeling framework.
To get the reliable meta-model, sufficient and accurate training data set of MBD is prerequisite, and random search is also important to define appropriate hyper-parameters of MBD problems such as the number of hidden layers, the size of batches, the number of epochs, optimizer, etc.
In particular, numerical results imply that a position of time variable as input or output data is crucial to get the usable transient response of MBD.

In Section \ref{sec:mbd formulation}, the governing equations of MBD is briefly reviewed.  
In Section \ref{sec:dnn}, the overview of neural networks of MBD and its meta-modeling process is presented. It should be noted that the framework of the proposed meta-modeling provides fundamental ideas of handling experimental or real-world data and exploiting their structures and relations to understand dynamics of general multibody systems. Not depending on complexity of MBD systems, the present meta-modeling helps us to achieve real-time and robust simulations with accurate motion results. In addition, high level of engineering simulations can be employed for not only engineering designs, but also motion related Internet of Things (IoT).
Section \ref{sec:numerical} describes the case studies of the meta-modeling process using single pendulum, double pendulum, and slider crank mechanism.
Conclusions are given in Section \ref{sec:conclusions}.

\section{Brief Review on Common General Purpose MBD Governing Equations}
\label{sec:mbd formulation}

Multibody system dynamics offers a straightforward approach to construct and solve equations of motion for mechanical systems. Multibody system dynamics includes a large number of procedures those can be categorized based on the used coordinates. In topological approaches, such like semi-recursive formulation, relative coordinates between the bodies are used. In the global approaches, in turn, the set of coordinates defines each body of the system. It is important to note that although topological and global approaches both lead to identical dynamic responses, the numerical performance differs. In this section often used global methods are briefly reviewed. 

In {\it{the augmented formulation}},  constraint equations are accounted in the equation of motion by employing Lagrange multipliers. In this approach the equations of motion can be written as
\begin{equation}
\label{eq:governing:augmented}
  \begin{bmatrix}
  \vec{M} & ~\CqT \\ 
  \Cq & \vec{0}
  \end{bmatrix}
  \begin{bmatrix}
  \ddq \\ \vec{\lambda}
  \end{bmatrix}
  = \begin{bmatrix}
  \vec{F}_a \\ \vec{F}_c
  \end{bmatrix}
\end{equation}
where $\vec{M}$ is the mass matrix, $\vec{C}$ is the constraint vector, $\Cq$ is the the Jacobian matrix of the constraint vector $\vec{C}$, $\vec{F}_a$ is the vector of applied generalized forces, and $\vec{F}_c$ the vector can be obtained by differential constraint twice with respect to time. The equation of motion is solved to obtain the generalized coordinates $\vec{q}$ and the Lagrange multipliers $\vec{\lambda}$. 

The other commonly used form of equations of motion for multibody system can be achieved from applying {\it{the embedding technique}} to global coordinates \eqref{eq:governing:augmented}. The embedding technique reduces the generalized coordinates to be solved from $\ddq$ to a set of independent generalized $\ddq_{ind}$. In practice, this reduction can be accomplished using a transformation matrix $\vec{T}$:
\begin{equation}
\label{eq:transformation}
  \ddq = \vec{T}\,\ddq_{ind} + \vec{r},
\end{equation}
where $\vec{r}$ is a remainder vector. Substituting \eqref{eq:transformation} into the augmented system \eqref{eq:governing:augmented} yields
\begin{equation}\begin{aligned}
\label{eq:augmented:modi}
  \begin{cases}
  \vec{M} \T \ddq_{ind} + \vec{M}\vec{r} + \CqT \vec{\lambda} = \vec{F}_a,
  \\
  \Cq \T \ddq_{ind} + \Cq\vec{r} = \vec{F}_c.
  \end{cases}
\end{aligned}\end{equation}
By applying an identity $ \T^T\CqT = \vec{0} $, the equation \eqref{eq:augmented:modi} can be simplified into
\begin{equation}
\label{eq:governing:embedded}
  \tilde{\vec{M}} \ddq_{ind} 
  = \tilde{\vec{F}},
\end{equation}
where
\begin{equation}\begin{aligned}
  &\tilde{\vec{M}} := \T^T \vec{M} \T,
  \\
  &\tilde{\vec{F}} := \T^T\vec{F}_a - \T^T\vec{M}\vec{r}.
\end{aligned}\end{equation}

\section{Deep Neural Network for Multibody Dynamics Systems}
\label{sec:dnn}
In this section, a brief introduction to DNN that will be used in numerical examples is presented, and training of the DNN for MBD systems is also described.
\\

Machine Learning (ML) aims to develop technologies and algorithms that enables computers to analyze and predict mechanisms of a system by learning structures of big amount of data. ML allows important tasks to be performed by generalizing from examples \cite{Dom12}. ML has already powered many aspects of modern society from web searches and item recognition to image classification, speech recognition \cite{LeBen15}, and cyber-physical systems (CPS)."

\begin{figure}[H]
  \centering
  \includegraphics[width=0.7\textwidth]
  {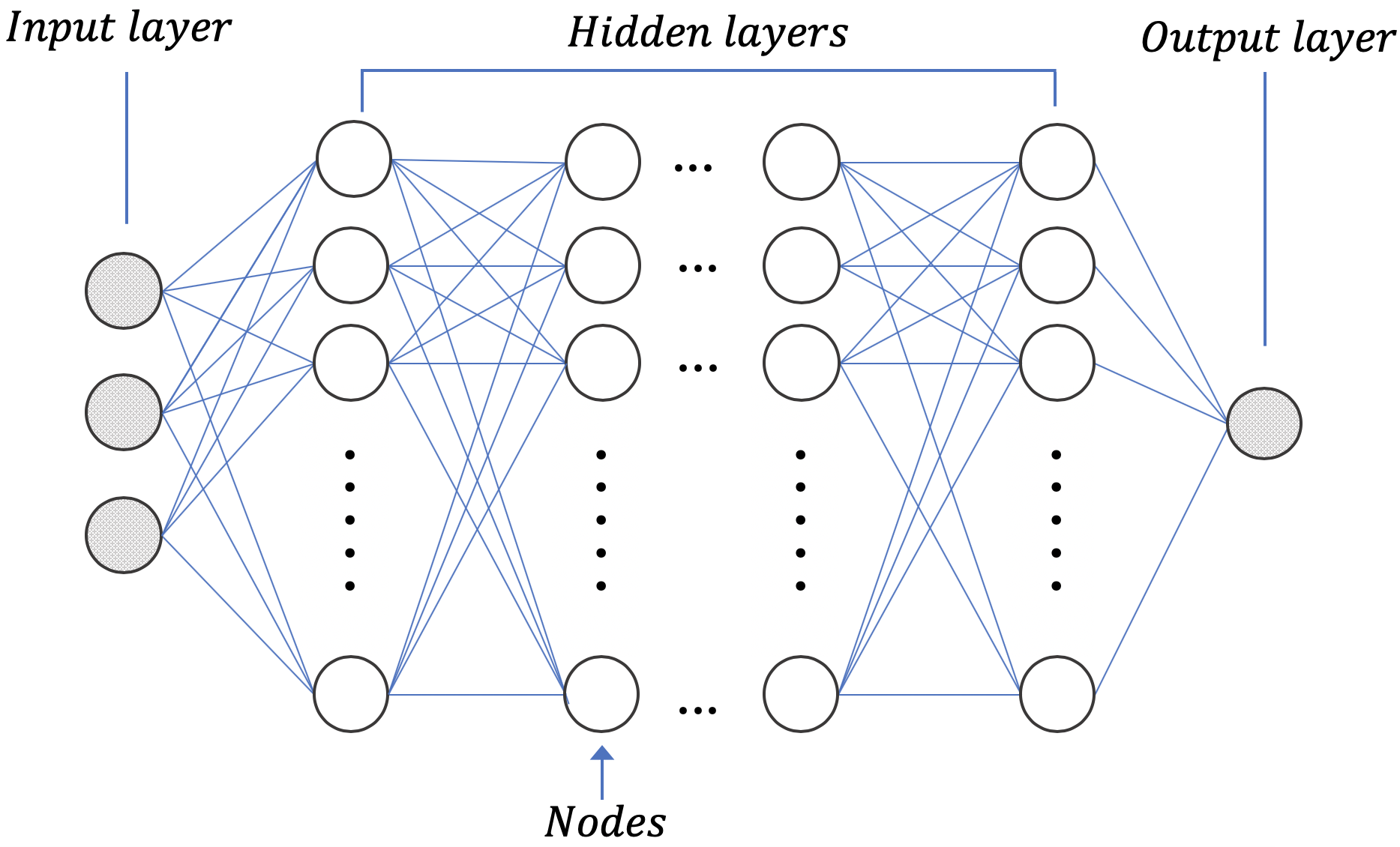}
\caption{
Structure of Artifical Neural Networks (ANN). If there are multiple hidden layers, ANN is referred to as Deep Neural Networks (DNN). 
}
\label{fig:ann_dnn}
\end{figure}

Being a part of ML, {\it{Artificial Neural Networks (ANN)}} are clusters of {\it{nodes}} (or {\it{neurons}}), which is designed to mimic the decision-making process of human brain, see Fig. \ref{fig:ann_dnn}. Nodes form {\it{layers}}, i.e. the input layer, the hidden layer, and the output layer. The input and output layers consist input and output parameters, respectively, of a meta-model. Containing information, nodes of each layer interchange the information through {\it{weights}}. One of the main purposes of ANN is to find the best weights to maximize the performance of a given neural network. Rumelhart et al. \cite{RumHinWil86} developed an {\it{error back-propagation algorithm}} to find weights and improve neural networks efficiently.

To describe and represent more complicated and intricate data, more than one hidden layer can be considered. In this case, the ANN is referred to as {\it{Deep Neural Networks (DNN)}}. The increased number of hidden layers increases the number of nodes and weights, which requires an expensive computational cost and makes it difficult to train a model. Despite the shortcomings, DNN yields better meta-models for solving complex {\it{nonlinear}} problems. 
\\

Structure of DNN can be specified in more details by the {\it{hyper-parameters}} such as the number of layers, the number of nodes for each layer, the batch size, the activation functions, the regulatory method, and the optimizers. The performance of DNN highly relies on the proper choice of hyper-parameters. 
Some important hyper-parameters mentioned in the numerical tests (Section \ref{sec:numerical}) are briefly summarized as follows:
\begin{enumerate}
\item [$\bullet$]
{\textit{Batch size}}\\
The batch size is the number of training data samples in one pass for updating weights. Due to memory limitations, it is not recommended to perform training with all available data samples at once. The larger the batch size, the less computational cost a training requires. 
\item [$\bullet$]
{\textit{Activation function}}\\
In DNN, values specified to nodes of a layer are not transferred directly to the next layer, but transformed through a nonlinear function, called {\it{activation function}}. It helps the values of nodes not to diverge during training and allows to solve complex problems with a small number of nodes. If an unsuitable activation function is chosen, gradients of DNN (in the error back-propagation process) can be vanishing, which makes learning speed severely slow. Activation functions such as {\it{tanh}}, {\it{sigmoid}}, and {\it{ReLU}}, are known to appropriate choices. 
\item [$\bullet$]
{\textit{Optimizers}}\\
Weights of DNN are found by error back-propagation process, which sequentially updates the weights to minimize a {\it{loss function}} defined by a given error, such as $\MSE$ described in \eqref{eq:MSE}. In this process, a local minimum problem needs to be solved and an efficient {\it{optimizer}} helps to reduce solution time. Representative techniques are stochastic gradient descent, Adam \cite{KinBa14}, RMSprop \cite{Hin12}.
\end{enumerate}

\subsection{Overview of Neural Networks for MBD} 
\label{sec:dnn:general}

\paragraph{Meta-model using Neural Networks}~\\


ML methods can be categorized in viewpoint of learning styles into three: supervised, unsupervised, and reinforcement learning. 
Supervised learning trains a meta-model by considering both reference response features called {\it{labels}} and predictive features, and by gradually improving the model to fit the given training data. There are mainly classification and regression methods in supervised learning.
In unsupervised learning, in contrary to supervised learning, label (or reference) features are not designated. It focuses on how training data is structured. 
Reinforcement learning is an effective algorithm for optimization analysis. It learns data by making decisions to maximize user-specified reward. Users need to design appropriate model conditions such as environments, actions, rewards.

MBD problems can be mainly dealt with supervised or reinforcement learning techniques since many MBD problems aim to seek robust and optimal design considering a set of design parameters.

To apply supervised learning, training data need to be prepared afore-hand for learning the model. The training data for MBD meta-models can be obtained in a few manners, usually by computational methods. In case of reinforcement learning, a multibody systems simulation environment is requisite to train an agent according to cumulated reward for each action. Both learning approaches require time-consuming tasks to learn the meta-models of MBD: data preparation task for supervised learning and simulation task for reinforcement learning. However, once the meta-model is built, it resolves MBD problems in real-time and yields dynamics responses.

In this research, the supervised learning of MBD meta-model based on training data is mainly considered. Supervised learning finds an approximation function $\M$ that minimizes a loss $L(x; \M)$ over samples $x$. An algorithm $\A_{\boldsymbol{\alpha}}$ produces $\M$ for a training set $\vec{X}^{train}$ through the optimization of a training criterion with respect to a set of parameters, given hyper-parameters $\boldsymbol{\alpha}$ \cite{BerBen12}. The built function 
$$
	\M = \A(\vec{X}^{train}; \boldsymbol{\alpha})
$$ 
is called a {\textit{meta-model}} in this research.  

A neural network algorithm is one of the powerful machine learning algorithms of minimizing the loss 
$$
	L\left(x;\, \A(\vec{X}^{train}; \boldsymbol{\alpha})\right).
$$ 

Specifically, the algorithm uses a network structure and optimizes the parameters of the networks, weights and biases, by utilizing the back-propagation algorithms, which is an extension of the gradient descent method for neural network structures.

In this research, neural networks are adopted to build the meta-models of MBD problems, since it is subject to be generalized to fit various shapes of nonlinear functions with multi-dimensional input data. In particular, {\textit{the feed forward networks (FFN)}} with hidden layers and nonlinear activation functions are considered, which are the universal approximators that can represent effectively continuous functions. Owing to the characteristics of FFN, it is a powerful candidate of implementing the meta-models of general purposed MBD problems. Moreover, many techniques for DNN including accelerated activation functions such as ReLU, dropout, regularization, and batch normalization have strengthened the potential of FFN with deep layers for modeling general purpose MBD problems. 

\begin{figure}[h]
  \centering
  \includegraphics[width=0.95\textwidth]
  {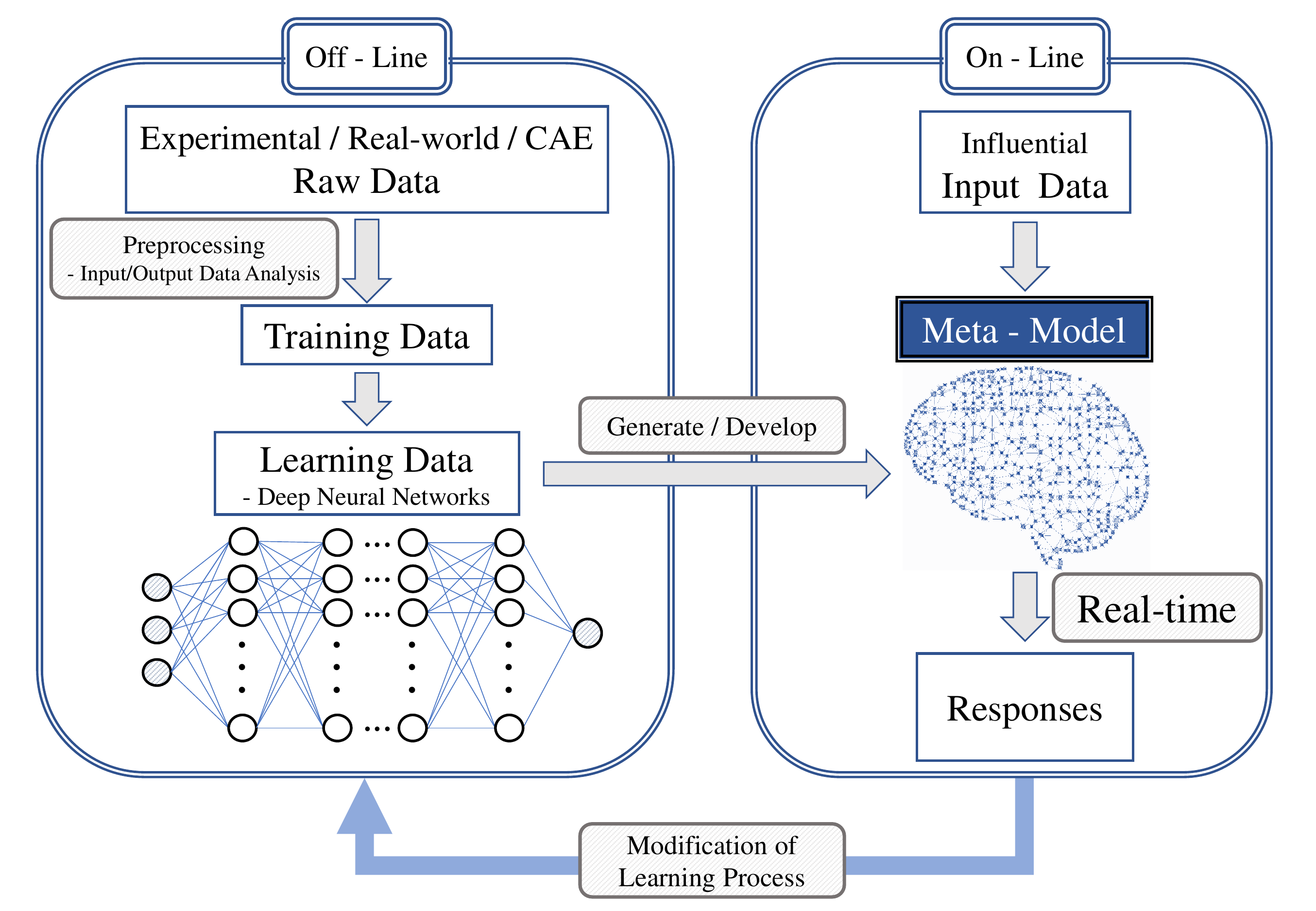}
\caption{
Flows of meta-modeling for MBD. By analyzing and learning data on MBD, a meta-model can be generated. The meta-model is intended to yield {\it{real-time}} dynamic responses of given MBD problems. Performance of the meta-model can be evaluated by comparing its results with experimental or real-world data. The evaluation helps to reconstruct or improve the off-line learning algorithm.
}
\label{fig:flowchart}
\end{figure}

The flowchart in Fig. \ref{fig:flowchart} shows brief outlines of meta-modeling of MBD problems and its benefits.

\paragraph{Design of Neural Networks for Meta-models}~\\
MBD problems rarely have high dimensionality of input or output data, compared to common DNN applications such as image, speech, and text data. Rather than high dimensionality, in general, MBD considers complicated nonlinear functions and requires accurate and robust solutions. 

If an MBD problem is given, the design of input and output layers is typically decided. For example, each variable of input (or output) data is mapped to a single node of the input (or output) layer in case that the variable is numeric one, but if the variable is nominal one, it should be mapped to multiple nodes through one-hot encoding. In one-hot encoding, each value of the nominal variable is transformed to one of one-hot vectors, 
$$
	\left\{(1,0,\cdots,0), (0,1,\cdots,0), \cdots, (0,0,\cdots,1)\right\}.
$$

Different from input and output layers, the design of hidden layers is volatile. The number of hidden layers and the number of nodes are the most critical hyper-parameters, and their best design must be decided along with other hyper-parameters at the step of hyper-parameter tuning. Empirically, it is known that deeper hidden layers are more effective than larger nodes of shallower hidden layers if two FFN models have similar numbers of parameters such as weights and biases.

To build expressive MBD meta-models, FFN models with enough width and depth are necessary. However, proper regularization methods such as $L_1$ and $L_2$ regularization, dropout and batch-normalization are required to achieve the generalized meta-models because FFN models with too many parameters are often overfitted to the given training data \cite{GodBenCou16}. 

\paragraph{Hyper-parameters Optimization of meta-models}~\\
Similar to typical ML algorithms, the neural network algorithm does not provide a method to find the optimal hyper-parameters $\boldsymbol{\alpha}$. 
Hyper-parameters of DNN are critical to the accuracy and robustness of the meta-model. Unfortunately, there is no perfect scheme of building the most accurate and robust DNN model from a given training data. One must search the best set of hyper-parameters such as the number of hidden layers, the number of nodes in each hidden layers, activation function, optimization function, learning rate, and the number of epochs.

Generally, two kinds of search methods are often used for the purpose of hyper-parameter optimization; a given set of candidate values for each hyper-parameter are investigated with the {\textit{grid search}} method, or randomly selected values for hyper-parameters are evaluated with the {\textit{random search}} method. It is known that random search is more efficient to find optimal hyper-parameters than grid search \cite{BerBen12}. Recently, AutoML is actively researched in academic and practical fields to find the best design of DNN. When the AutoML techniques are mature, it is expected that the optimal design of the DNN-based meta-models can be found in easier and faster manners \cite{FeuEggKalLinHut18}. 

\paragraph{Generation of MBD Training Data}~\\
In this paper, it is assumed that one can obtain as many MBD sample data as is need to train the meta-model and achieve a reliable model. In other words, a case with an insufficient training data set is not considered. Nevertheless, since the process of MBD data collection takes so long time in case of complex multibody systems, a more efficient manners of collecting training data is needed. 

First, the amount of training samples can be determined according to some criteria. Incremental learning methods can be applied to learn the meta-models. For instance, a certain level of performance measures such as the root mean squared errors or the mean absolute percentage errors can be adopted for the criteria to stop feeding more samples to the meta-model. In case of the random search method, simply more random samples can be provided to the less trained meta-model, and in case of the grid search method, finer-grained grid samples can be done \cite{HuaLeeLinHua07}. 

Second, the range of each design parameter for more training samples can be adjusted after seeking less accurate ranges of design parameters of the meta-model. It is under an assumption that model complexity is often different in many ranges of nonlinear hyperplanes. In such cases, adaptive sampling methods such as focused grid search can be less exhaustive than uniform design of the typical grid search method \cite{PonAmoBalPaiFer16}.

\subsection{Detailed Assumptions and Conditions for Meta-modeling Process} 
\label{sec:dnn:details}
The followings are some assumptions and comments on the meta-modeling that is developed for MBD problems. The same conditions are applied to the numerical tests in Section \ref{sec:numerical}. 

\begin{enumerate}
\item
{\textit{Training Data}}
\\[1.5mm] $\bullet$
{\textit{Sufficiently many sets of training data}}\\
As mentioned in Section \ref{sec:dnn:general}, it is assumed that there are as many sets of data for training and tests as one wants. Since the most important objective of this research is to achieve a highly accurate meta-model, the other issues such as computational efficiency and problems of insufficient training data are not mainly concerned. 
\\[1.5mm] $\bullet$ 
{\textit{Uniform Meshes}}\\
Training data for input parameters are uniformly meshed in a given finite range. 
\\[1.5mm] $\bullet$ 
{\textit{Data without Noise}}\\
Training data for output responses such as displacements, veloicities, or acclerations are {\it{exactly}} calculuated from governing equations for MBD problems. In other words, training data are artifically generated without any noise. 
\\[1.5mm] $\bullet$
{\textit{Time Variable and Structurues of Training Data}}\\
An important question in meta-modeling for dynamic problems is whether time variable $t$ needs to be handled as an input parameter or not. 

Table \ref{tab:train:time_input} shows an example of training data set, where time variable $t$ is considered as an input. All the discrete time instants are contained in the set of training data. On the other hand, if time variable is not considered as an input parameter, there are $\#\{t_n\}$ sets of training data, where time is fixed to $t = t_n$, as shown in Table \ref{tab:train:time_fixed}. The two types of training data structures are referred to as \Sfull\, and \Sfixed.
\\

It may seem that \Sfixed\, is simpler than \Sfull, in that the former considers a fixed time instant $t = t_n$ and has a much smaller size of training data set compared to \Sfull, especially when the number of discrete time instants is very large. However, handling time variable as a non-input (\Sfixed) is not adequate for MBD analysis in two following major aspects:
\\
\begin{enumerate}
\item
It requires to make as many meta-models as the number of discrete time instants $t_n$, $n = 0, 1, \cdots$. 
Moreover, if grid search is performed for each meta-model to find out the best hyper-parameters, this approach can be computationally infeasible. 
\item
Each resulting meta-model provides predictions only for a specific time $t = t_n$, which makes it difficult to figure out time-varying tendency of MBD.\\
\end{enumerate}

Thus, in this research, it is concluded that a meta-model for MBD problems need to be generated from training data of form \Sfull, where time variable is considered as an input. More details on training data structure and its results are described in Section \ref{sec:numerical}. 
\\
\item
{\textit{Test Data}}
\\[1.5mm] 
$\bullet$ 
{\textit{Unseen Data}}\\
The performance of a resulting meta-model is evaluated with some sets of test data which are unseen from training process. 
\\[1.5mm] 
$\bullet$ 
{\textit{Randomly Distributed Data}}\\
Unlike training data, input parameters for test are not uniformly meshed. They are randomly distributed in the same given range. 
\\
\item
{\textit{Grid Search and Hyper-parameters}}
\\[1.5mm] 
Grid search is performed to find out appropriate hyper-parameters for each MBD example, which helps to yield a highly accurate meta-model. From grid search, the number of hidden layers, the number of nodes for each layer, the size of batches, the number of epochs, optimizer, and loss functions need to be decided. Still, there can be other sets of hyper-parameters that result in similar or better performance. 
\\
\item
{\textit{Evaluation of Performance}}
\\[1.5mm]
The performance of a resulting meta-model $\M$ is evaluated in terms of two measures: $R$-squared value and absolute mean-squared error (MSE), denoted by $\R$ and  $\MSE$, respectively. 
When an output label $\yl$ is given for a set of test data, and the mata-model $\M$ yields a prediction $\yp$ for the test set, the performance measures are defined by 
\begin{equation}\begin{aligned}
  \R(\yl, \yp) 
  := 1 - \dfrac{\sum_{i=1}^{N}\left(\yl_i - \yp_i\right)^2}{\sum_{i=1}^{N}\left(\yl_i - \bar{\yl}\right)^2},
\end{aligned}\end{equation}
\begin{equation}\begin{aligned}
\label{eq:MSE}
  \MSE(\yl, \yp) 
  := \dfrac{1}{N}\sum_{i=1}^{N} \left(\yl_i - \yp_i\right)^2,
\end{aligned}\end{equation}
where $\yl = (\yl_1, \cdots, \yl_N)$, $\yp = (\yp_1, \cdots, \yp_N)$, and $\bar{\yl} := \sum_{i=1}^{N}\yl_i/N$. 
As the solution $\yp$ of the meta-model predicts the label $\yl$ more accurately, the value of $\R$ closes to $1$, and the error $\MSE$ closes to $0$. 
\end{enumerate}

\begin{table}
\centering \ra{1.3}
\begin{tabular}{@{}cccccc@{}}
\toprule
\multicolumn{3}{c}{Input} 
& \phantom{abc}
& \multicolumn{2}{c}{Output} 
\\
\cmidrule{1-3} \cmidrule{5-6}
$L$ & $c$ & $t$
&& $\theta$ & $\dot{\theta}$ 
\\ \midrule 
    0.0  & 0.00  & 0.00  &   & 1.57080 & 0.00000 \\
    0.0  & 0.00  & 0.01  &   & 1.56589 & -0.98100 \\
    0.0  & 0.00  & 0.02  &   & 1.55118 & -1.96192 \\
    \midrule
    0.10  & 0.05  & 0.00  &  & 1.57080 & 0.00000 \\
    0.10  & 0.05  & 0.01  &  & 1.56590 & -0.97936 \\
    0.10  & 0.05  & 0.02  &  & 1.55122 & -1.95540  \\
    \midrule
    0.20  & 0.10  & 0.00  &  & 1.57080 & 0.00000 \\
    0.20  & 0.10  & 0.01  &  & 1.56590 & -0.97773 \\
    0.20  & 0.10  & 0.02  &  & 1.55126 & -1.94890 \\
    \bottomrule
\end{tabular}
\caption{Structure of training data set for DNN, where time variable $t$ is considered as an {\it{input}}. This type of training data structure is denoted by \Sfull. In this case, a single meta-model is generated.}
\label{tab:train:time_input}
\vspace{1cm}
\begin{tabular}{@{}cccccc@{}}
\toprule
\phantom{abc}
& \multicolumn{2}{c}{Input} 
& \phantom{abc}
& \multicolumn{2}{c}{Output} 
\\
\cmidrule{2-3} \cmidrule{5-6}
& $L$ & $c$
&& $\theta$ & $\dot{\theta}$ 
\\ \midrule 
    $t = 0.00$ 
    & 0.0  & 0.00   &  & 1.57080 & 0.00000 \\
    & 0.10  & 0.05  &  & 1.57080 & 0.00000 \\
    & 0.20  & 0.10  &  & 1.57080 & 0.00000 \\
    \bottomrule
\end{tabular}
\\[2mm]
\begin{tabular}{@{}cccccc@{}}
\toprule
\phantom{abc}
& \multicolumn{2}{c}{Input} 
& \phantom{abc}
& \multicolumn{2}{c}{Output} 
\\
\cmidrule{2-3} \cmidrule{5-6}
& $L$ & $c$
&& $\theta$ & $\dot{\theta}$ 
\\ \midrule 
    $t = 0.01$ 
    & 0.0  & 0.00  &  & 1.56589 & -0.98100 \\
    & 0.10  & 0.05 &  & 1.57080 & 0.00000 \\
    & 0.20  & 0.10 &  & 1.56590 & -0.97773 \\
    \bottomrule
\end{tabular}
\\[2mm]
\begin{tabular}{@{}cccccc@{}}
\toprule
\phantom{abc}
& \multicolumn{2}{c}{Input} 
& \phantom{abc}
& \multicolumn{2}{c}{Output} 
\\
\cmidrule{2-3} \cmidrule{5-6}
& $L$ & $c$
&& $\theta$ & $\dot{\theta}$ 
\\ \midrule 
    $t = 0.02$ 
    & 0.0  & 0.00   &  & 1.55118 & -1.96192 \\
    & 0.10  & 0.05  &  & 1.55122 & -1.95540  \\
    & 0.20  & 0.10  &  & 1.55126 & -1.94890 \\
    \bottomrule
\end{tabular}
\caption{Structure of training data set for DNN, where time variable $t$ is fixed and not considered as an input. This type of training data structure is denoted by \Sfixed. 
In this case, $\#\{t_n\}$ numbers of meta-models are generated corresponding to $\#\{t_n\}$ sets of training data. 
}
\label{tab:train:time_fixed}
\end{table}
\section{Case Studies}
\label{sec:numerical}
In this section, three fundamental MBD examples, single pendulum, double pendulums, and slider crank mechanisms, are investigated. For each example, a data-driven meta-model is generated through FFN, and its performance is evaluated in various ways, as described in Section \ref{sec:dnn:details}. 

\subsection{Damped Single Pendulum}
\label{sec:single}
\begin{figure}[H]
  \centering
  \includegraphics[width = 0.27\textwidth]
  {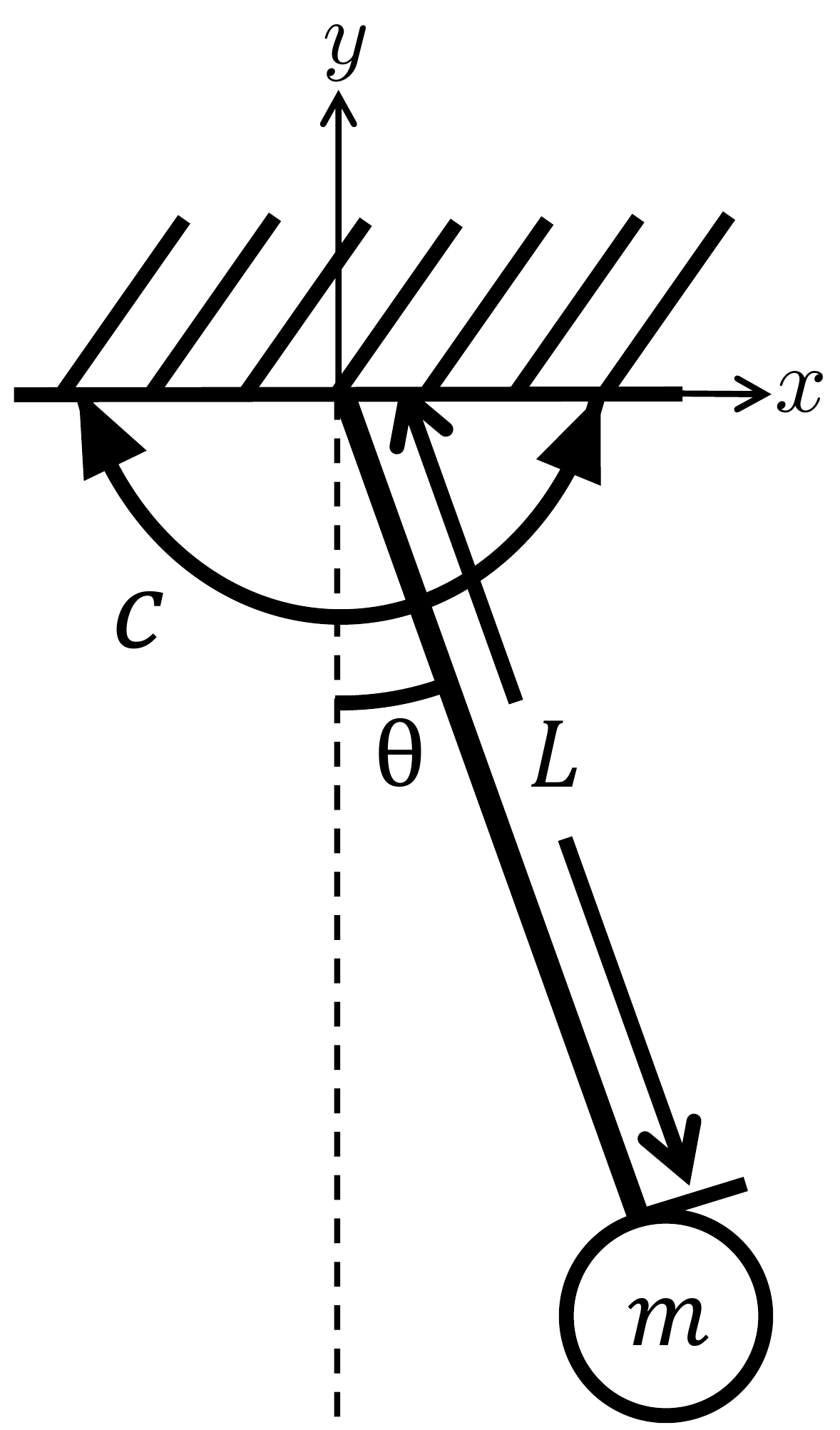}
\caption{
Dampled single pendulum problem. Gravity acceleration $g$, the mass $m$, and the initial angle $\theta^0$ are fixed to $g = 9.81[m/s^2], m = 0.3[kg],$ and $\theta^0 = \pi/2[rad]$. The length of the massless rod $L[m] \in [0.1, 0.2]$, the damping coefficient $c[kg\cdot m/s] \in [0,1]$, and the initial angular velocity $\dot{\theta}^0[rad/s] \in [0, 5]$ are arbitrarily determined within the given ranges. 
}
\label{fig:single:diagram}
\end{figure}

A damped single pendulum problem shown in Fig. \ref{fig:single:diagram} can be expressed in the following mathematical governing equation: 
\begin{equation}\begin{cases}
\label{eq:single:governing}
  \ddot{\theta} 
  + \cfrac{g}{L} \sin(\theta) 
  + \cfrac{c}{mL} \dot{\theta} = 0,
  & \text{where}~~ \theta = \theta(t), 
  \quad t \in [0, t_f],
  \\
  \theta(t) = \theta^0, 
  \dot{\theta}(t) = \dot{\theta}^0, 
  & \text{where}~~  t = 0,
\end{cases}\end{equation}
where $g$ is the gravity acceleration, $L$ is the length of the massless rod, $m$ is the mass, and $c$ is the damping coefficient, respectively.
The variables $\theta$ and $\dot{\theta}$ are time-varying angle and its velocity, whose initial values are specified as $\theta^0$ and $\dot{\theta}^0$, respectively. 

Although all the input parameters $(g, L, m, c, \theta^0, \dot{\theta}^0)$ affect dynamics of the single pendulum in Fig. \ref{fig:single:diagram}, it is empirically noticed that the parameters $(L, c, \dot{\theta}^0)$ make a major influence on the dynamic response characteristics. Thus, it is assumed that the relatively insignificant parameters $(g, m, \theta^0)$ are fixed to values $(9.81[m/s^2]$, $0.3[kg]$, $\pi/2[rad])$, while the parameters $(L, c, \dot{\theta}^0)$ are not determined specifically. It is the objective of this example to generate a meta-model which yields the dynamics of damped single pendulum as outputs when a particular set of input parameters $(L, c, \dot{\theta}^0)$ are given.

For an efficient learning, it is assumed that $(L, c, \dot{\theta}^0)$ are chosen within finite ranges:
\begin{equation*}\begin{aligned}
  L[m] 
  &\in [0.1, 0.2] ~~(\Delta{L} = 0.01)
  , 
  \\
  c[kg\cdot m/s] 
  &\in [0, 0.15] ~~(\Delta{c} = 0.01)
  , 
  \\
  \dot{\theta}^0[rad/s] 
  &\in [0, 5] ~~(\Delta{\dot{\theta}^0} = 0.5).
\end{aligned}\end{equation*}
Here, $(\Delta{L}, \Delta{c}, \Delta{\dot{\theta}^0})$ denote uniform meshsizes for training data. In evaluating a meta-model, the uniform meshes are not applied, and arbitrarily chosen input values are used. 

To describe dynamics of the damped single pendulum, the time-varying solutions $\theta(t)$, $\dot{\theta}(t)$, and $\ddot{\theta}(t)$ are achieved as outputs of a meta-model. 
For time variable $t$, discrete time instants $\{t_n\}$ with a uniform meshsize $\Delta{t}$ is considered in an interval $[0,t_f]$, where $t_f=2$:
\begin{align}
  t_n[s] := n \,\Delta{t} \in [0, 2] 
  \quad \left(\Delta{t} = 10^{-2}\right), 
\end{align}
for $n = 0, 1, \cdots, 200$. 
\\

As described in Section \ref{sec:dnn:details}, time variable $t$ can be handled as an input (\Sfull) or fixed to a certain instant (\Sfixed). Results from the two structures are compared. \Sfull\, case generates only one meta-model, while \Sfixed\, case $\#\{t_n\} = 201$ meta-models. 
Thus, for \Sfull, the input and output of meta-model are four and three dimenional, repectively. The total number of training data is $267,531$. 
\Sfixed has three dimensional input and the number of its training data is $1,331$ for each model. 
\\

Hyper-parameters found from grid search are shown in Table \ref{tab:single:hyper-param}.
\begin{table*}[h]
\centering \ra{1.15}
\begin{tabular}{@{}lc@{}}
\toprule
~~Hyper-parameters~
& ~~Choice~~~
\\
\midrule
The number of hidden layers & 2\\
The number of nodes in each layer & 128\\
The size of batch & 64\\
The number of epochs & 400\\
Loss function & $\MSE$\\
Optimizer & Adam\\
\bottomrule
\end{tabular}
\caption{Hyper-parameters for the damped single pendulum problem }
\label{tab:single:hyper-param}
\end{table*}

\begin{figure}
  \centering
  \includegraphics[width = \figsize\textwidth]
  {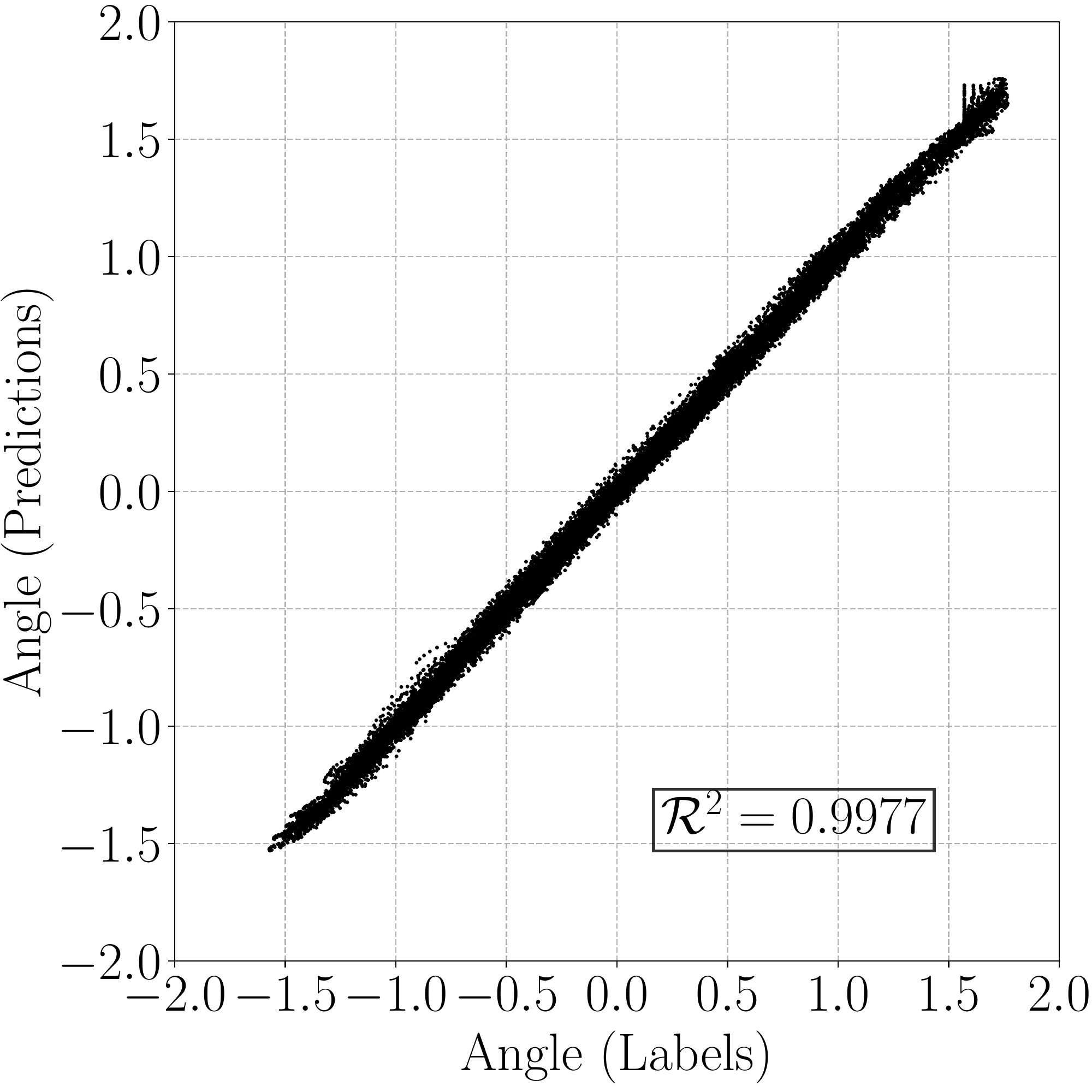}
  \includegraphics[width = \figsize\textwidth]
  {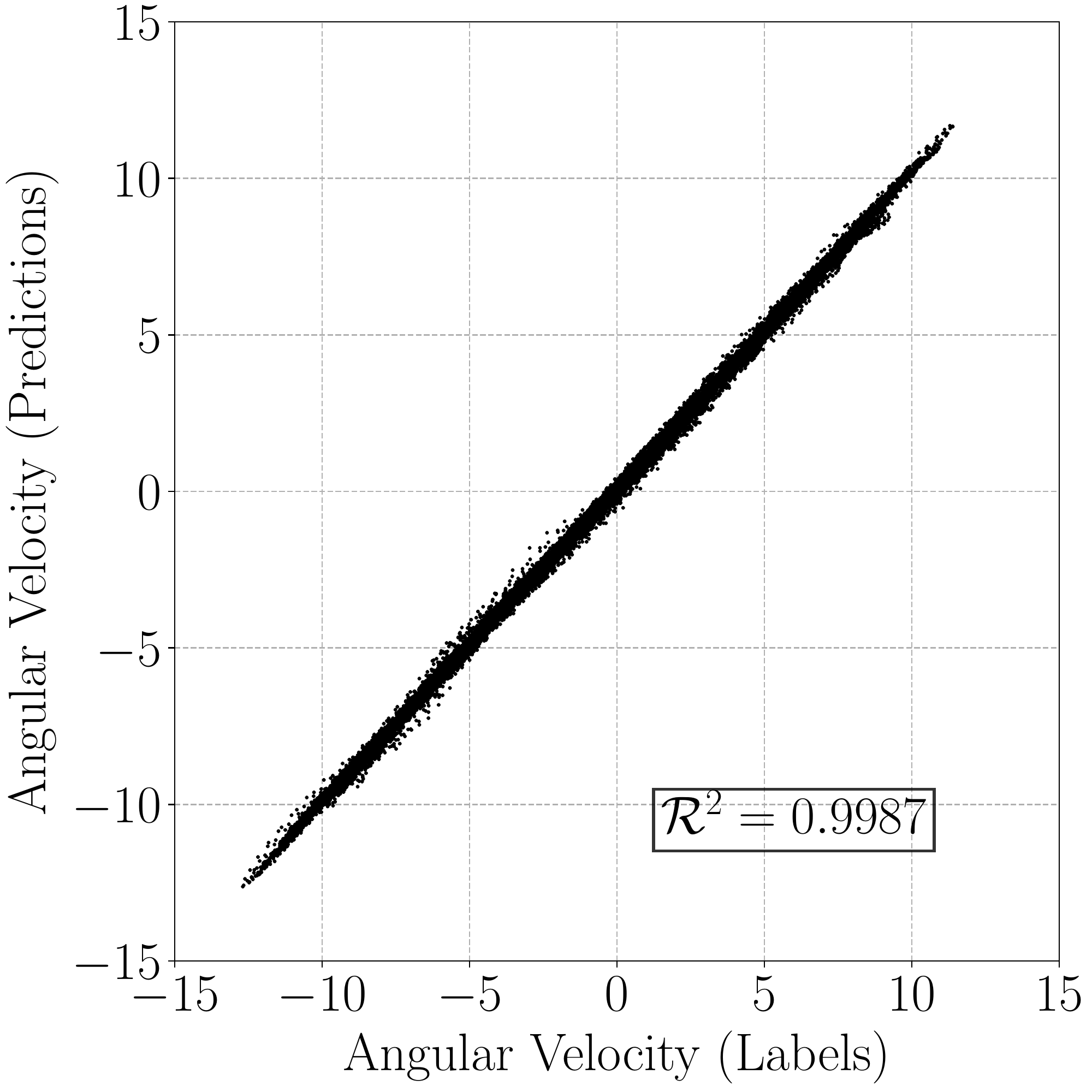}
  \includegraphics[width = \figsize\textwidth]
  {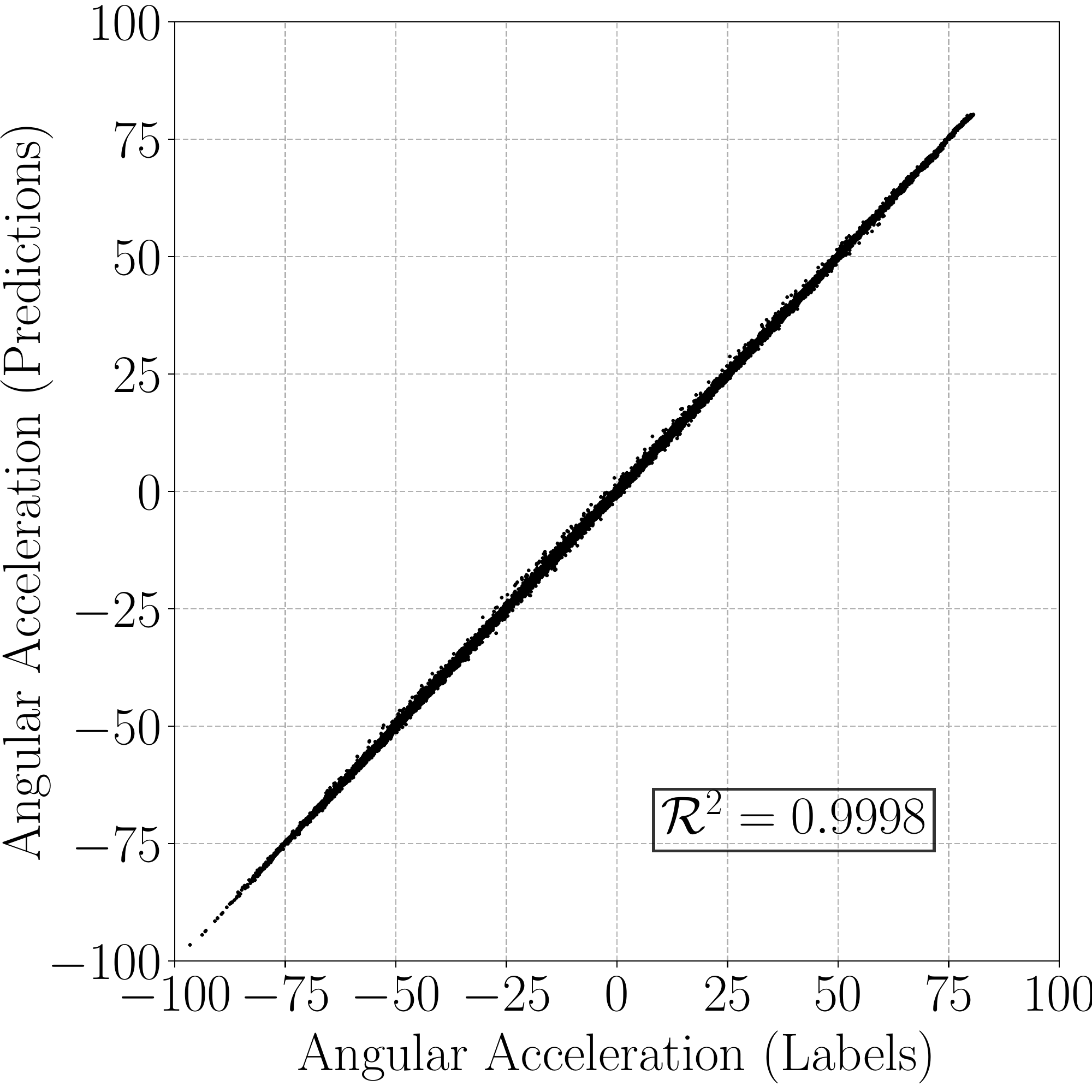}
\caption{
Labels vs. Predictions for test data. The meta-model for the damped single pendulum problem is generated from \Sfull\, type of training set. Test data are {\it{unseen}} from training. The $\R$ values are almost 1, which implies that the meta-model predicts output solutions with high accuracy.
}
\label{fig:single:scatter}
\end{figure}

Fig. \ref{fig:single:scatter} displays the scatter plots where {\it{labels}}, i.e. reference solutions, and predictions of outputs $(\theta, \dot{\theta}, \ddot{\theta})$ are compared. The results are achieved from a set of test data, which are unseen from training. The $\R$ scores are around $0.997$, which implies that the DNN model predicts the outputs with high accuracy. 

\begin{figure}
  \centering
  \includegraphics[width =\figsize\textwidth]
  {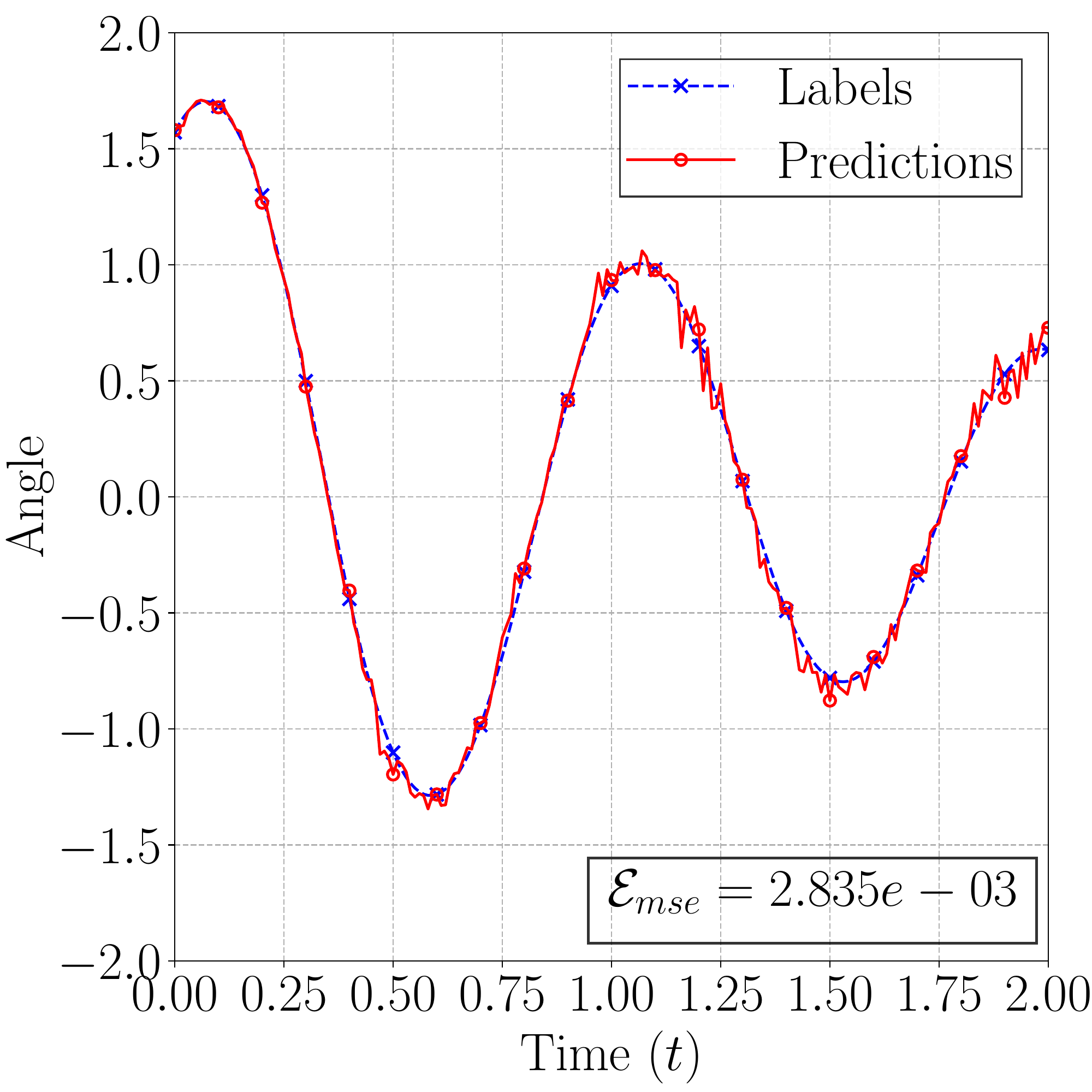}
  \includegraphics[width = \figsize\textwidth]
  {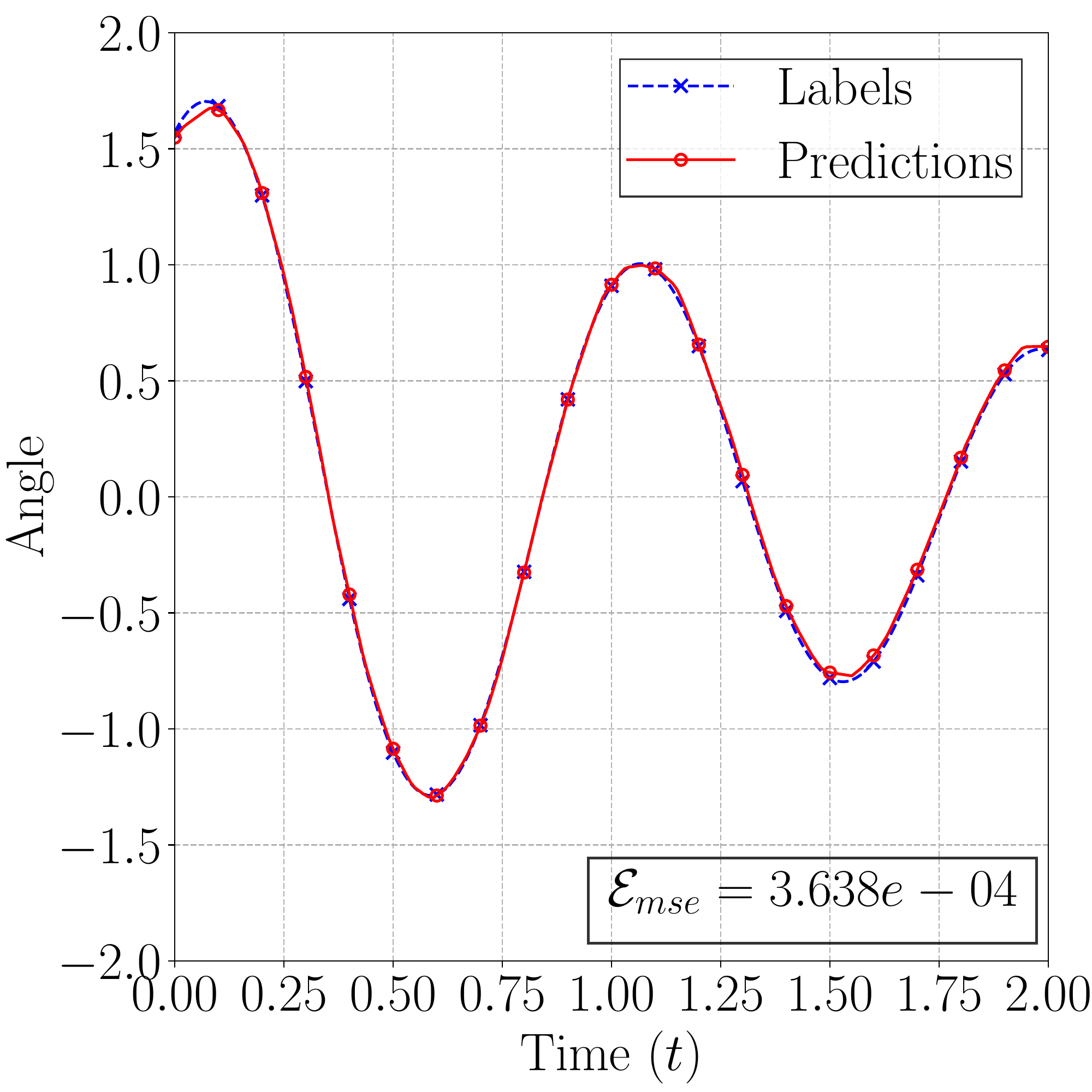}
  \includegraphics[width = \figsize\textwidth]
  {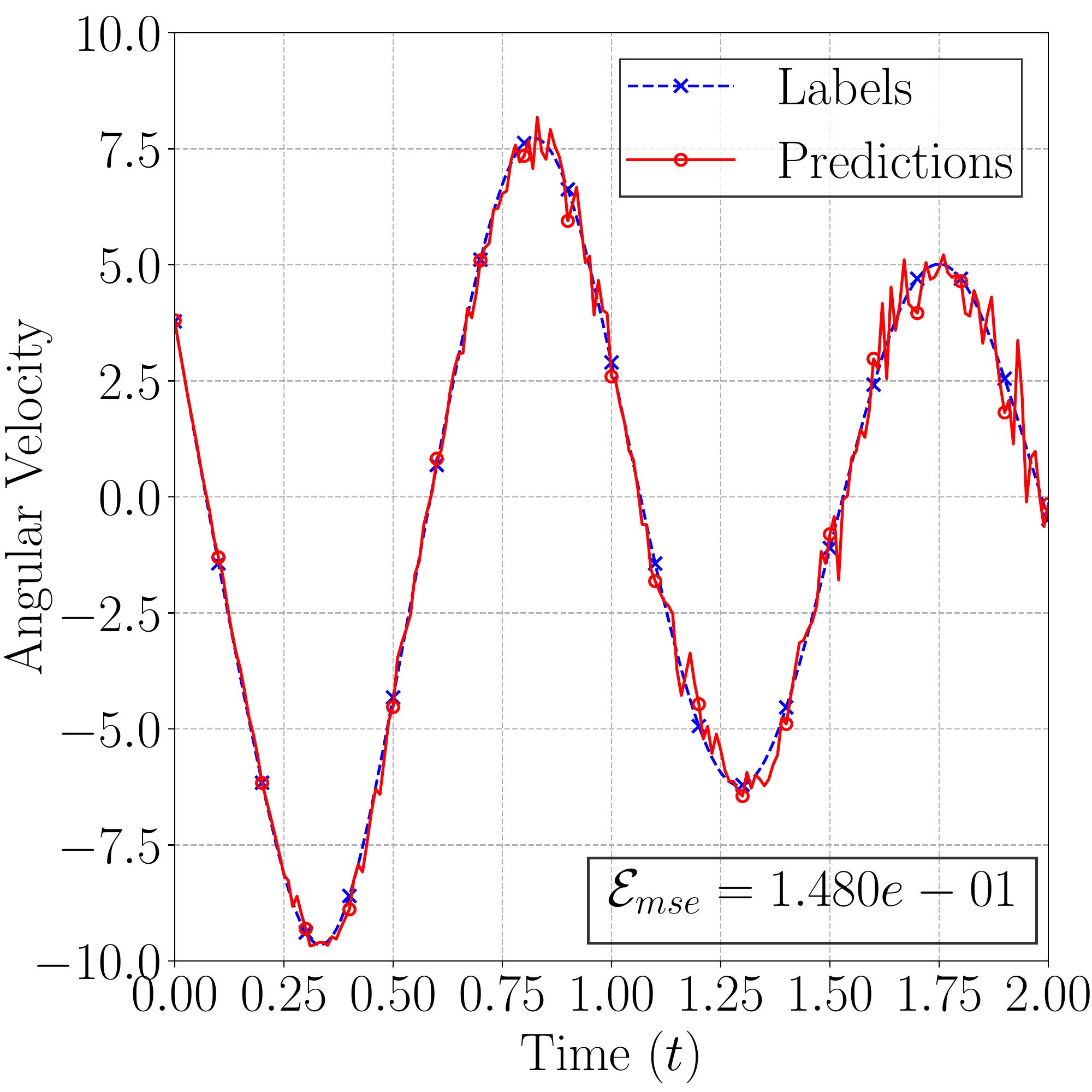}
  \includegraphics[width = \figsize\textwidth]
  {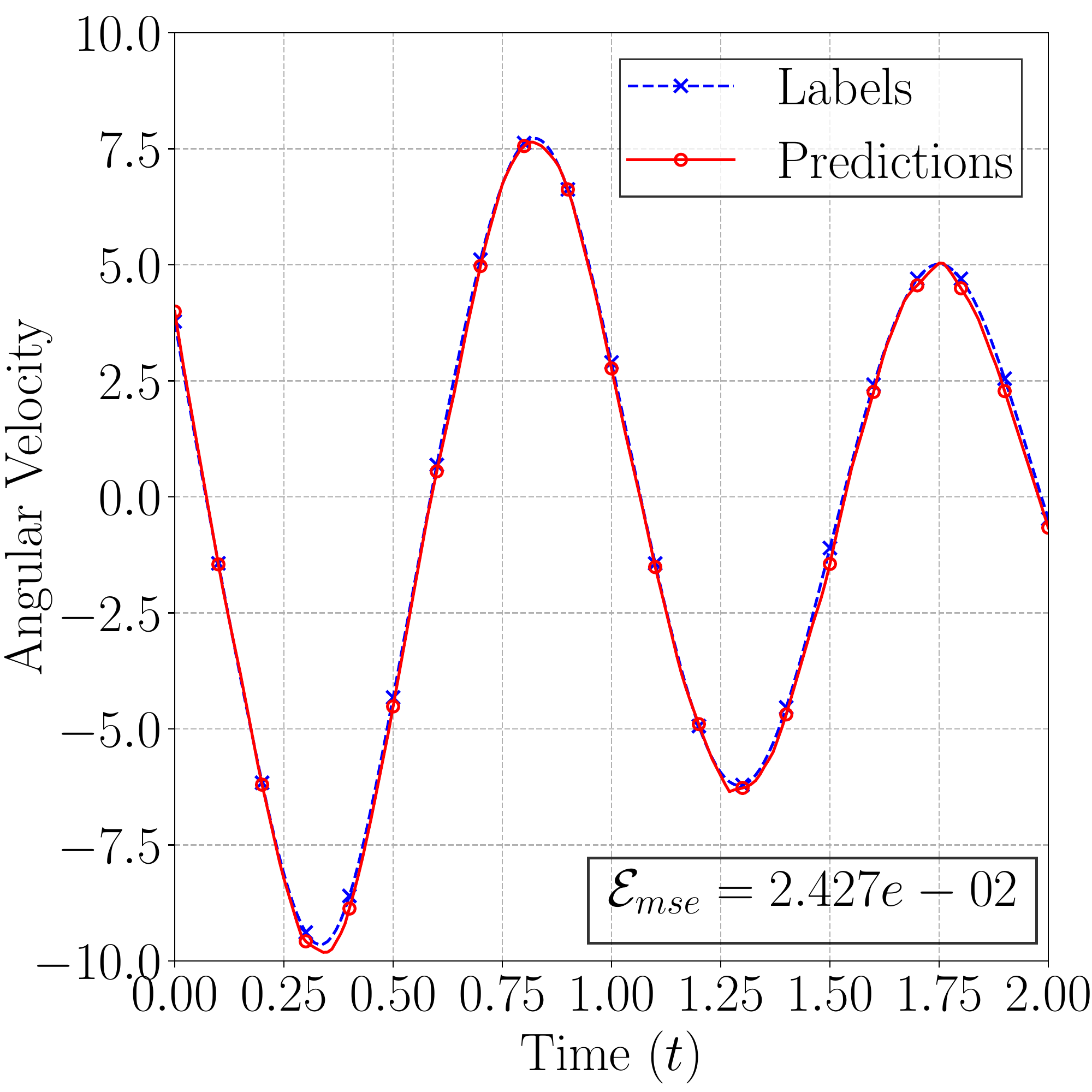}
  \includegraphics[width = \figsize\textwidth]
  {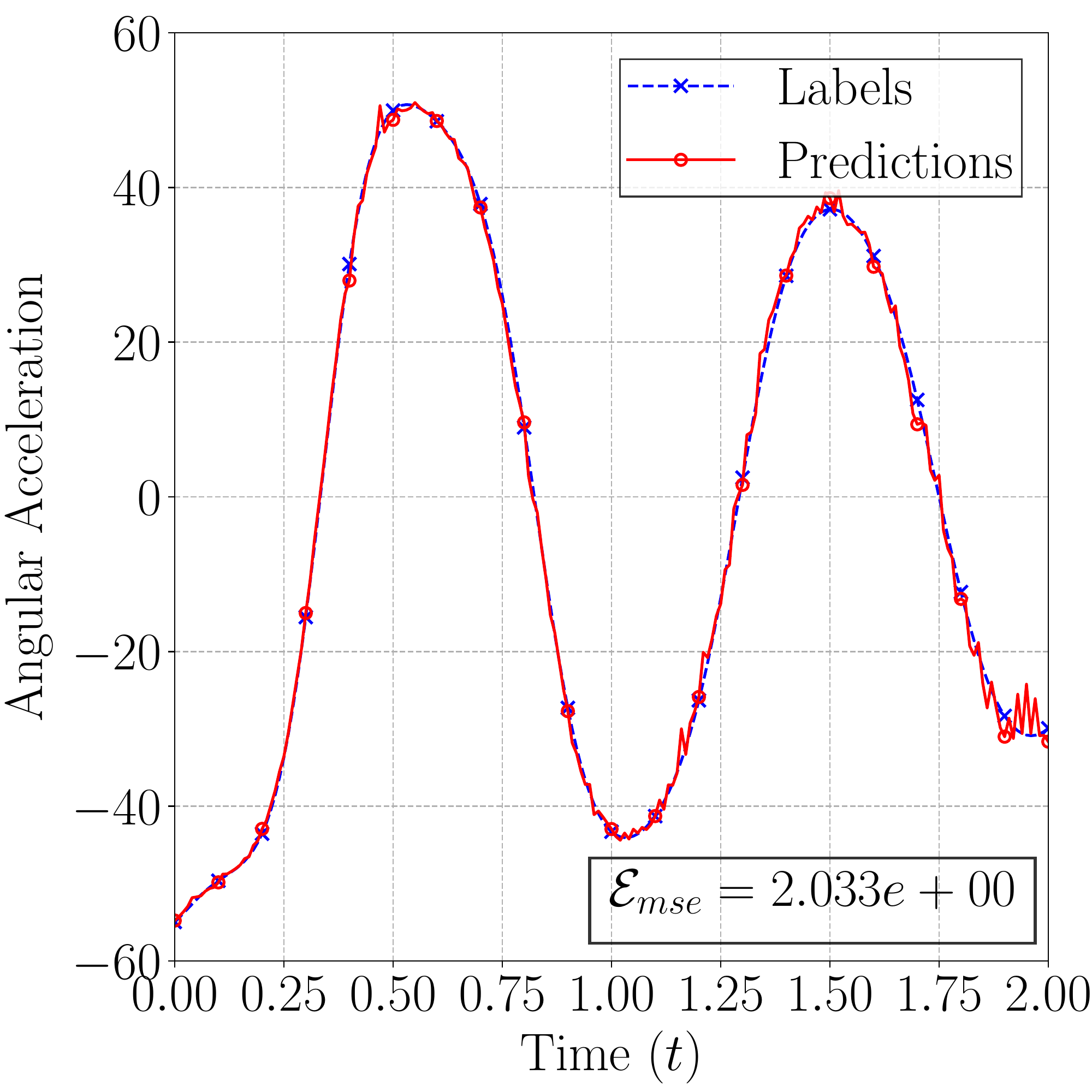}
  \includegraphics[width = \figsize\textwidth]
  {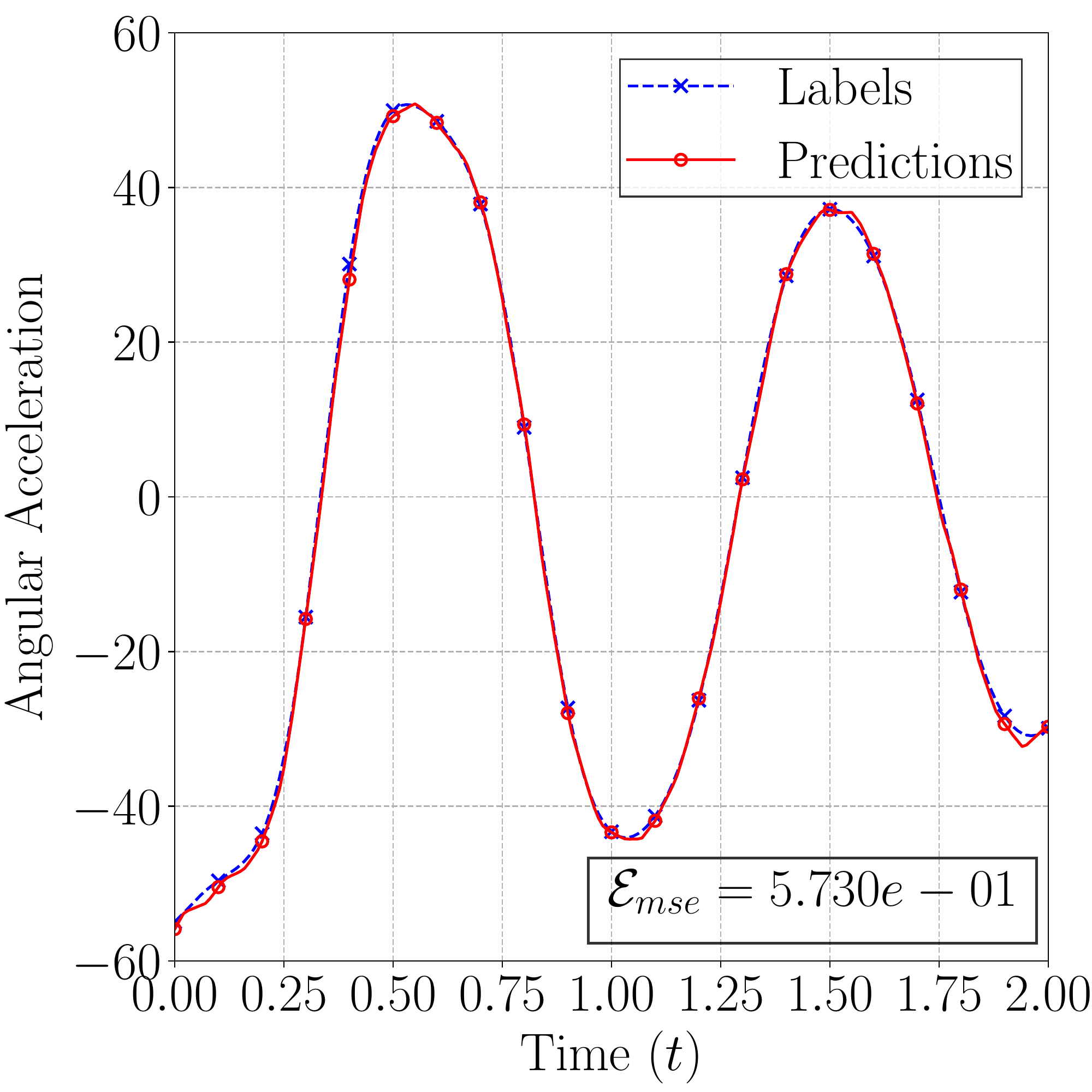}
\caption{Dynamic responses of the damped single pendulum for specific input 
\inputSingle{0.1911}{3.78}{0.055}. 
Labels(blue dashed, crosses) and predictions(red solid, circles) are compared for test data. Left:\, $\#\{t_n\}$ numbers of meta-models are generated for each fixed time $t = t_n$ (\Sfixed). Some oscillations are observed. Right:When time variable $t$ is considered as an input parameter (\Sfull). Relatively smooth solutions are achieved.
}
\label{fig:single:specific}
\end{figure}

Fig. \ref{fig:single:specific} shows dynamics of angle($\theta$) (Top), angular velocity($\dot{\theta}$) (Middle), and angular acceleration($\ddot{\theta}$) (Bottom), for a specific case: 
\inputSingle{0.1911}{3.78}{0.055}.
Labels (blue dashed, crosses) and predictions (red solid, circles) are shown for each solution. Results of \Sfixed (Left) and \Sfull (Right) are compared. Although both \Sfixed\, and \Sfull\, yields highly accurate results, some oscillations are observed in case of \Sfixed(Left). On the other hand, \Sfull(Right) gives relatively smooth solutions. 

\begin{figure}
  \centering
  \includegraphics[width = \figsize\textwidth]
  {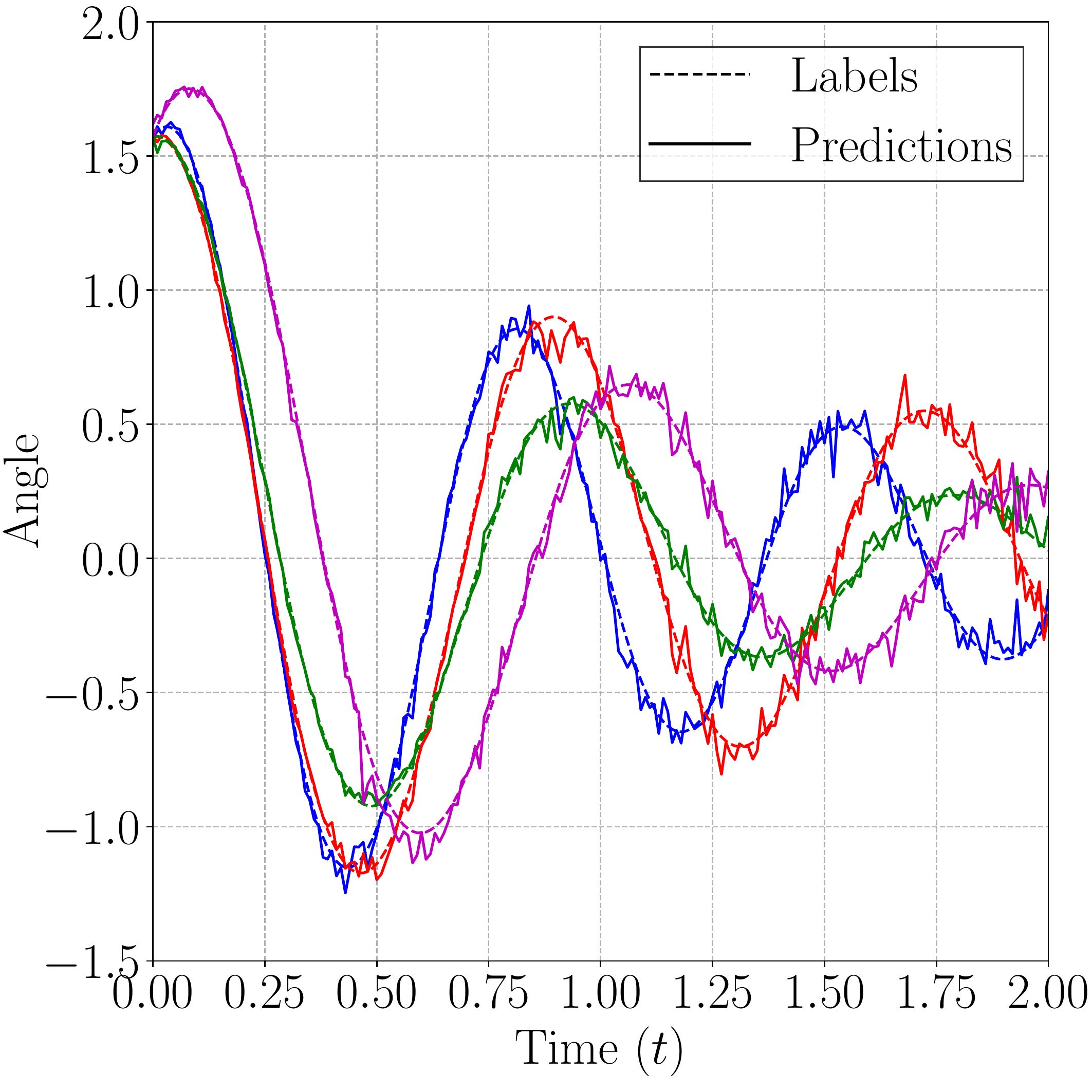}
  \includegraphics[width = \figsize\textwidth]
  {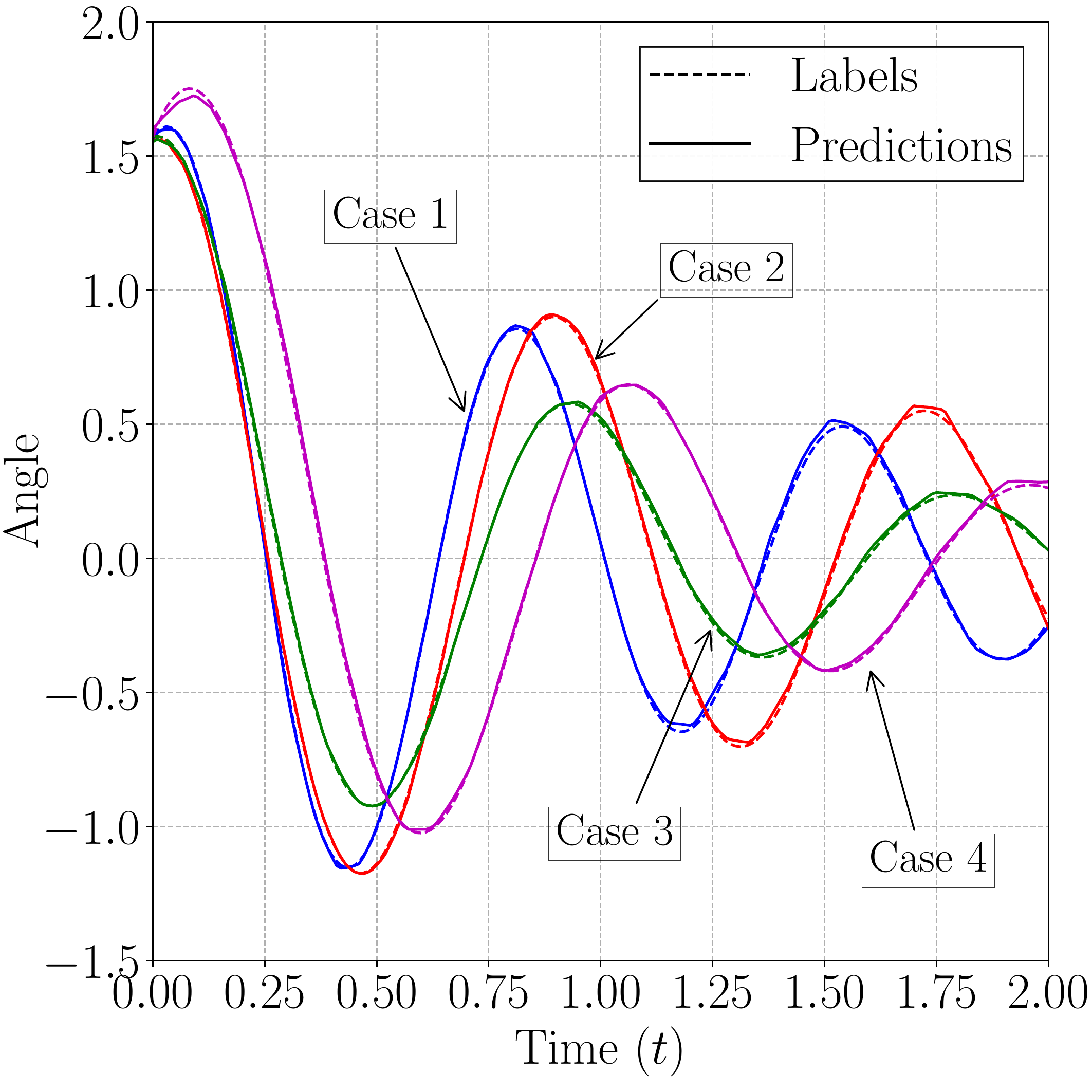}
  \includegraphics[width = \figsize\textwidth]
  {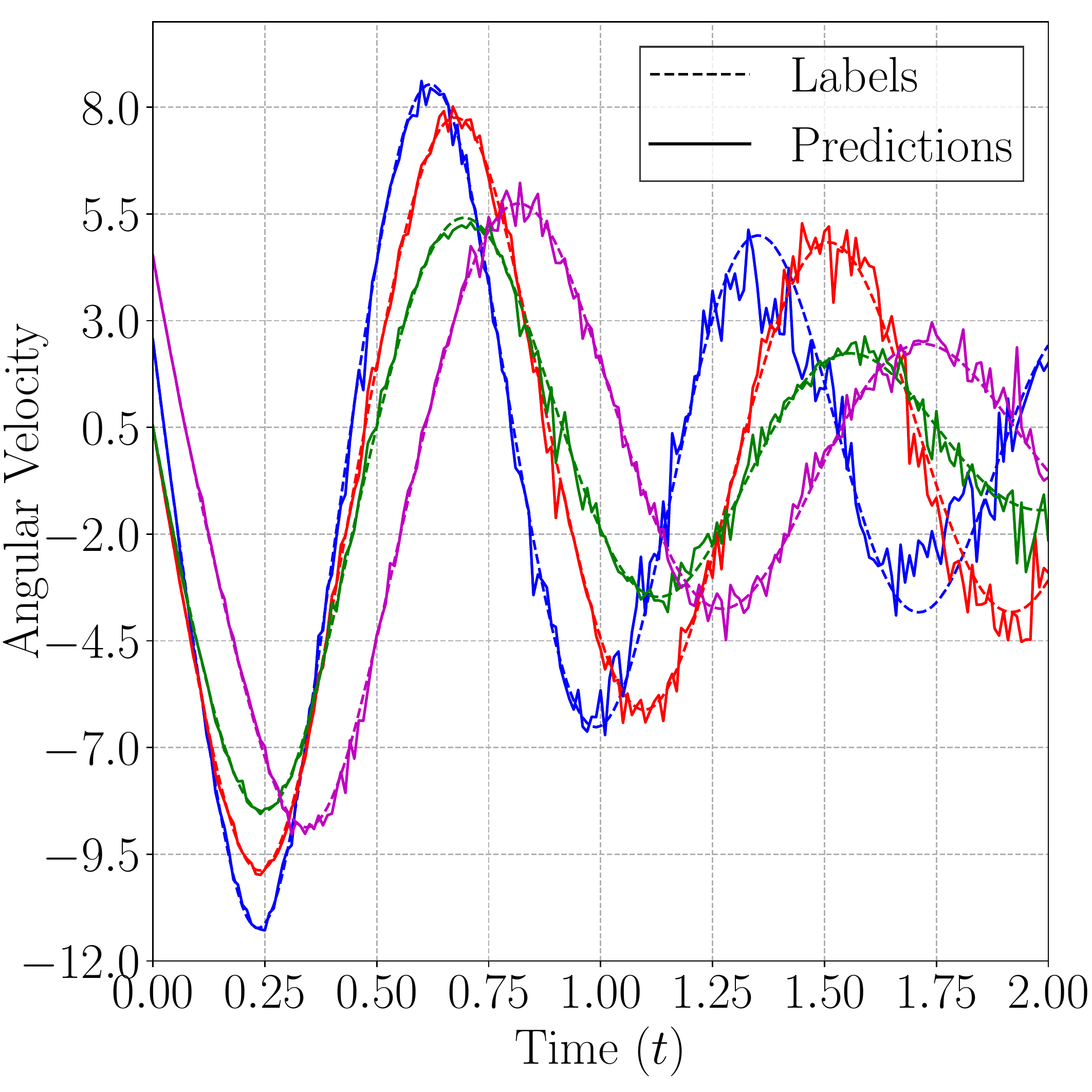}
  \includegraphics[width = \figsize\textwidth]
  {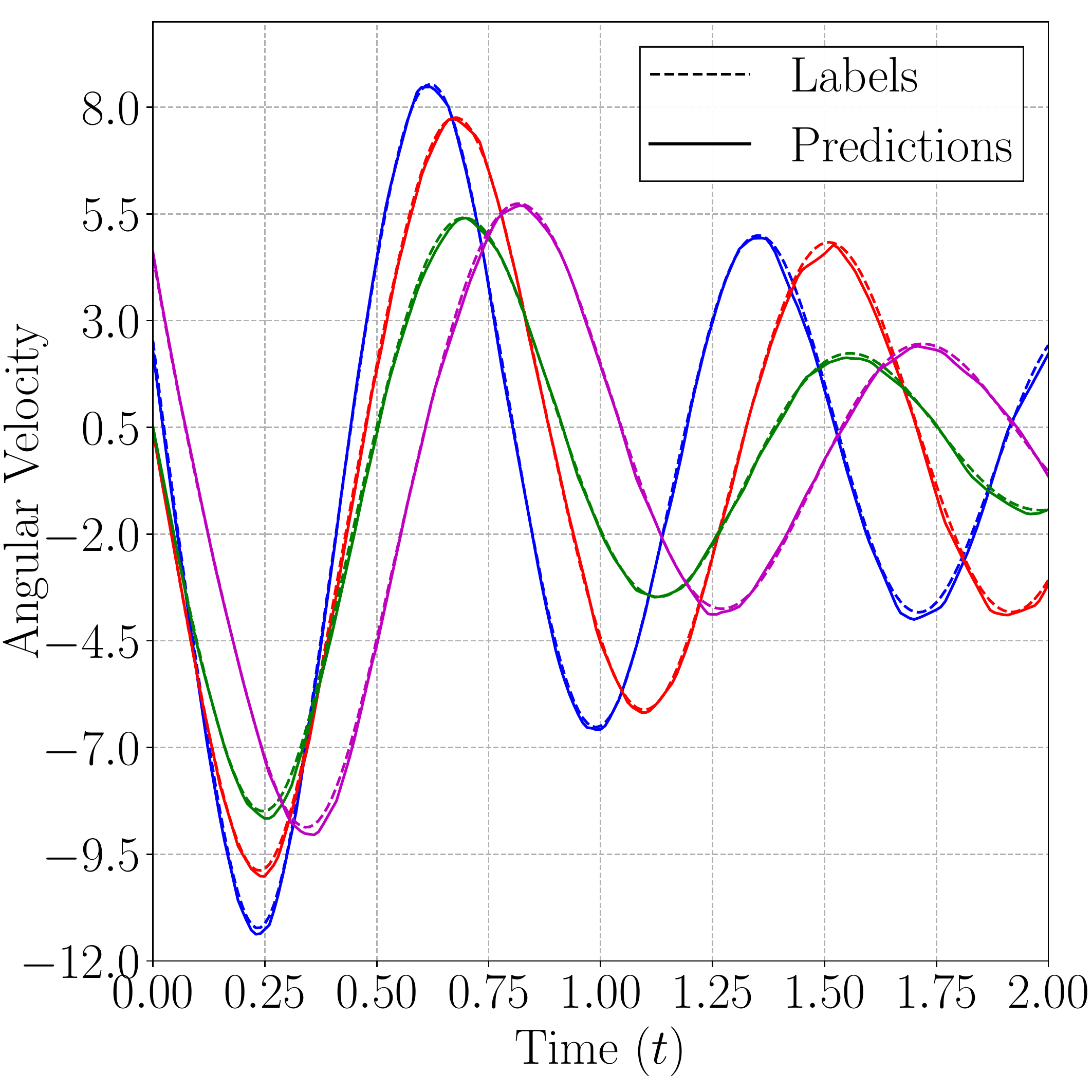}
  \includegraphics[width = \figsize\textwidth]
  {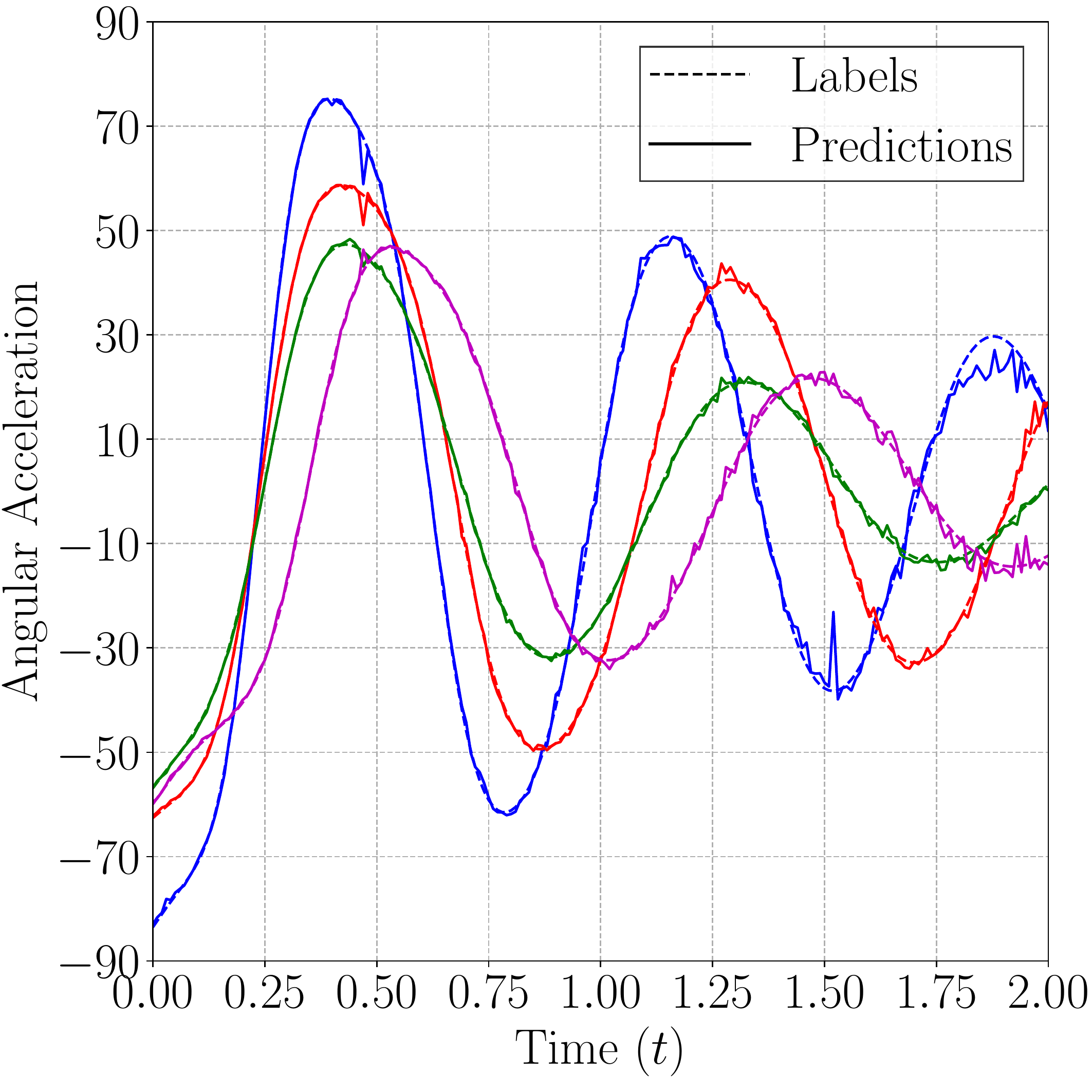}
  \includegraphics[width = \figsize\textwidth]
  {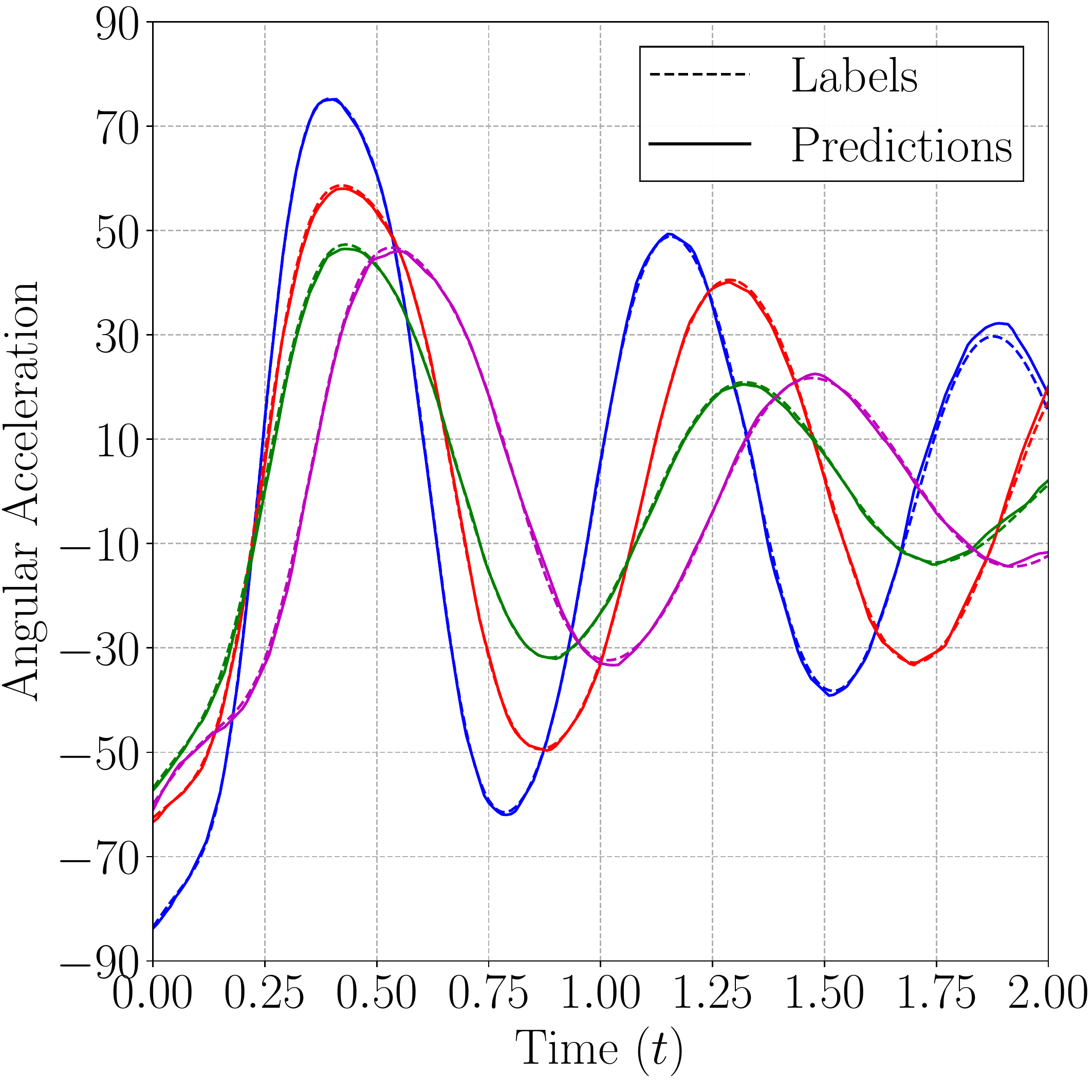}
\caption{Dynamic responses of single pendulum for multiple inputs 
\inputSingle{0.123}{2.53}{0.055} (blue), 
\inputSingle{0.1583}{0.52}{0.055} (red), 
\inputSingle{0.1758}{0.52}{0.109} (green), 
\inputSingle{0.1911}{4.52}{0.109} (magenta).
Labels(dashed) and predictions(solid) are compared for test data.
Left:$\#\{t_n\}$ numbers of meta-models are generated for each fixed time $t = t_n$.(\Sfixed). Some oscillations are observed. Right:When time variable $t$ is considered as an input parameter (\Sfull). Relatively smooth solutions are achieved.
}
\label{fig:single:multiple}
\end{figure}

In Fig. \ref{fig:single:multiple}, performance comparison of \Sfixed(Left) and \Sfull(Right) for other input parameters are summarized in Table \ref{tab:single:multiple}:
Similarly as in Fig. \ref{fig:single:specific}, oscillatory waves are observed in case of \Sfixed. Some are more severe than others, which makes prediction error greater. On the other hand, \Sfull\, yields smooth and accurate predictions for all cases.

\begin{table*}
\centering \ra{1.3}
\begin{tabular}{@{}ccccc@{}}
\toprule
& $L~[m]$
& $c~[kg \cdot m/s]$
& $\dot{\theta}^0~[rad/s]$
\\
\midrule
Case 1
& 0.123 & 2.53 & 0.055\\
Case 2
&0.1583 & 0.52 & 0.055\\
Case 3
& 0.1758 & 0.52 & 0.109\\
Case 4
& 0.1911 & 4.52 & 0.109\\
\bottomrule
\end{tabular}
\caption{Input parameters of multiple cases for Fig. \ref{fig:single:multiple}}
\label{tab:single:multiple}
\end{table*}

\subsubsection{Hyper-parameters for \Sfull\, and \Sfixed}
\label{sec:single:newgrid}

In the damped single pendulum problem, 
the same hyper-parameters are used to both types of training data \Sfull\, and \Sfixed, where the hyper-parameters are found from a grid search for \Sfull. 
Since the data structures of \Sfull\, and \Sfixed\, are different, it would be the best to carry out independent grid search for each structure, in comparing results of \Sfull\, and \Sfixed. Obviously, the performance of \Sfixed\, will be improved if more appropriate hyper-parameters are applied. 
To clarify positives and negatives of employing better hyper-parameters for \Sfixed, independent grid searches for \Sfixed\, models are performed. Since there are $\#\{t_n\}=201$ models in \Sfixed, $\#\{t_n\}$ grid searches are required. The hyper-parameters found for \Sfixed\, are listed in Table \ref{tab:single:Sfixed:hyper-parameters}.

\begin{figure}
  \centering
  \includegraphics[width =\figsize\textwidth]
  {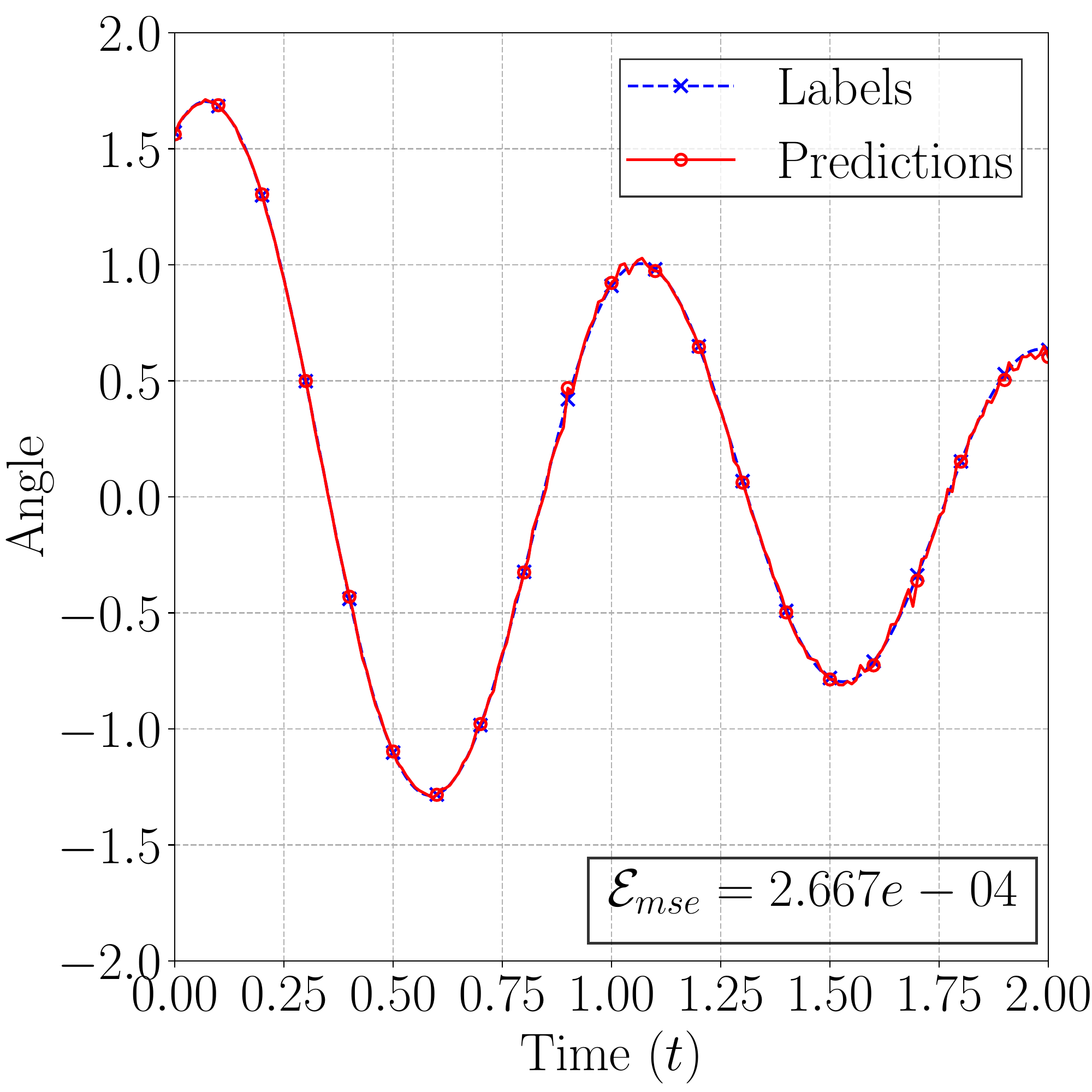}
  \includegraphics[width = \figsize\textwidth]
  {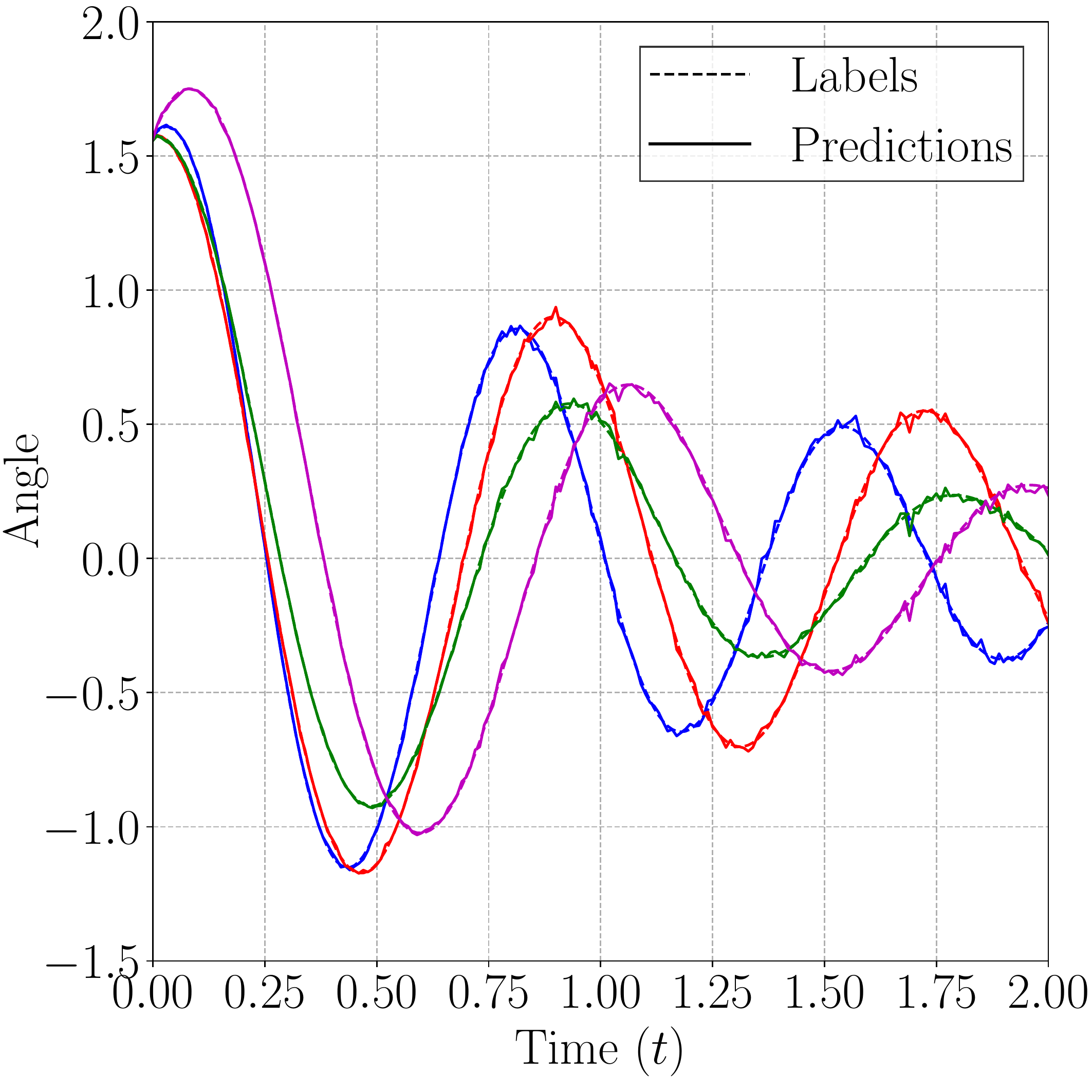}
  \includegraphics[width = \figsize\textwidth]
  {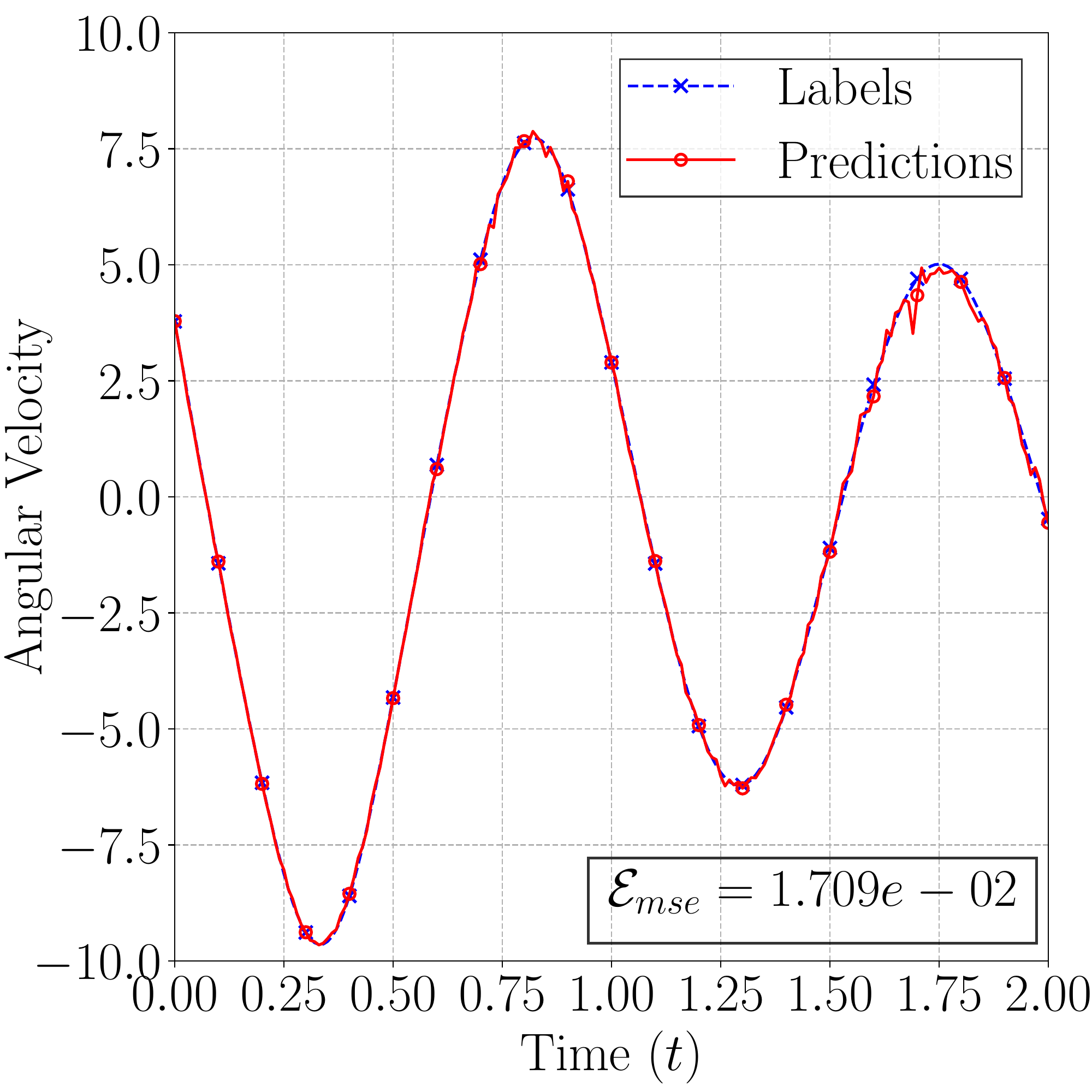}
  \includegraphics[width = \figsize\textwidth]
  {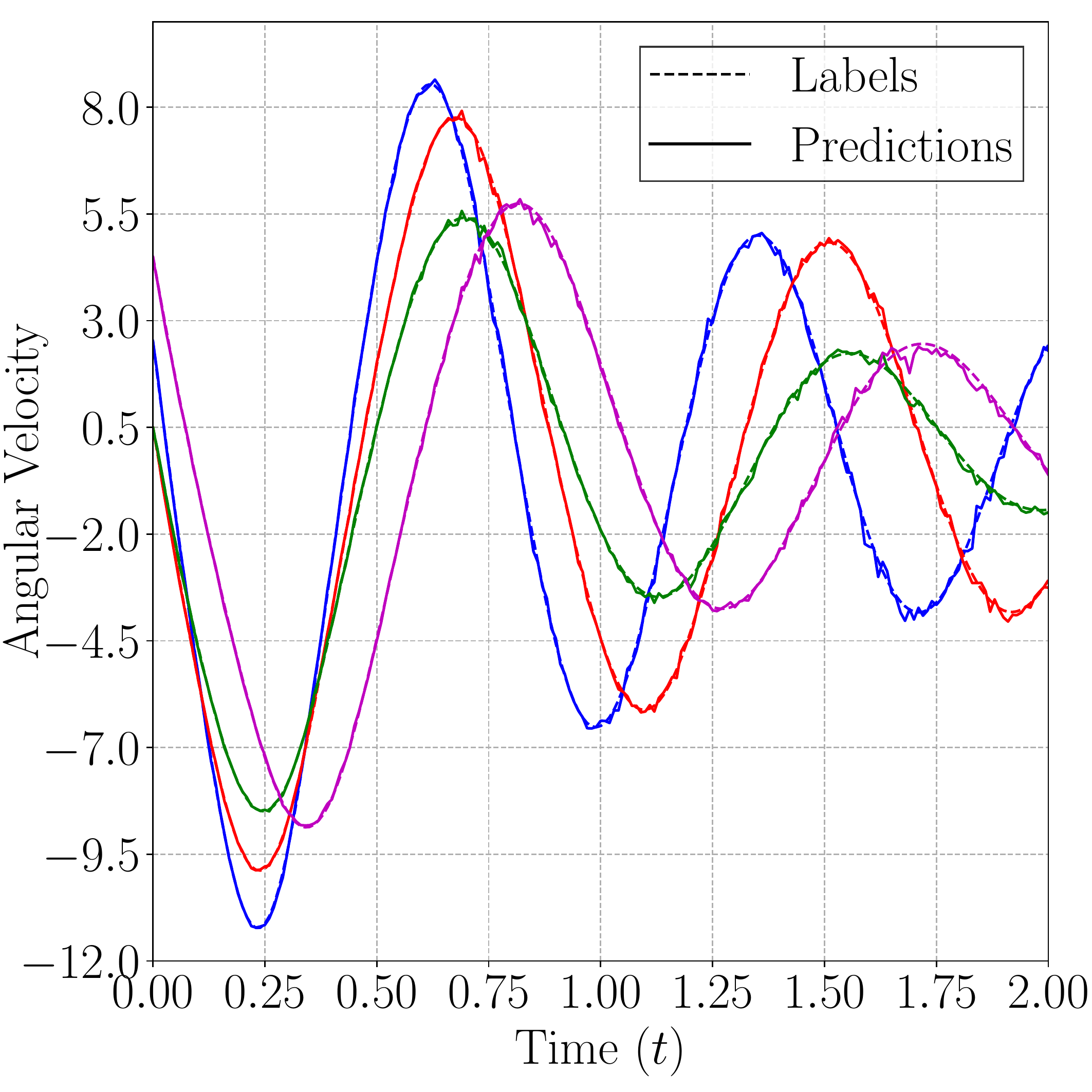}
  \includegraphics[width = \figsize\textwidth]
  {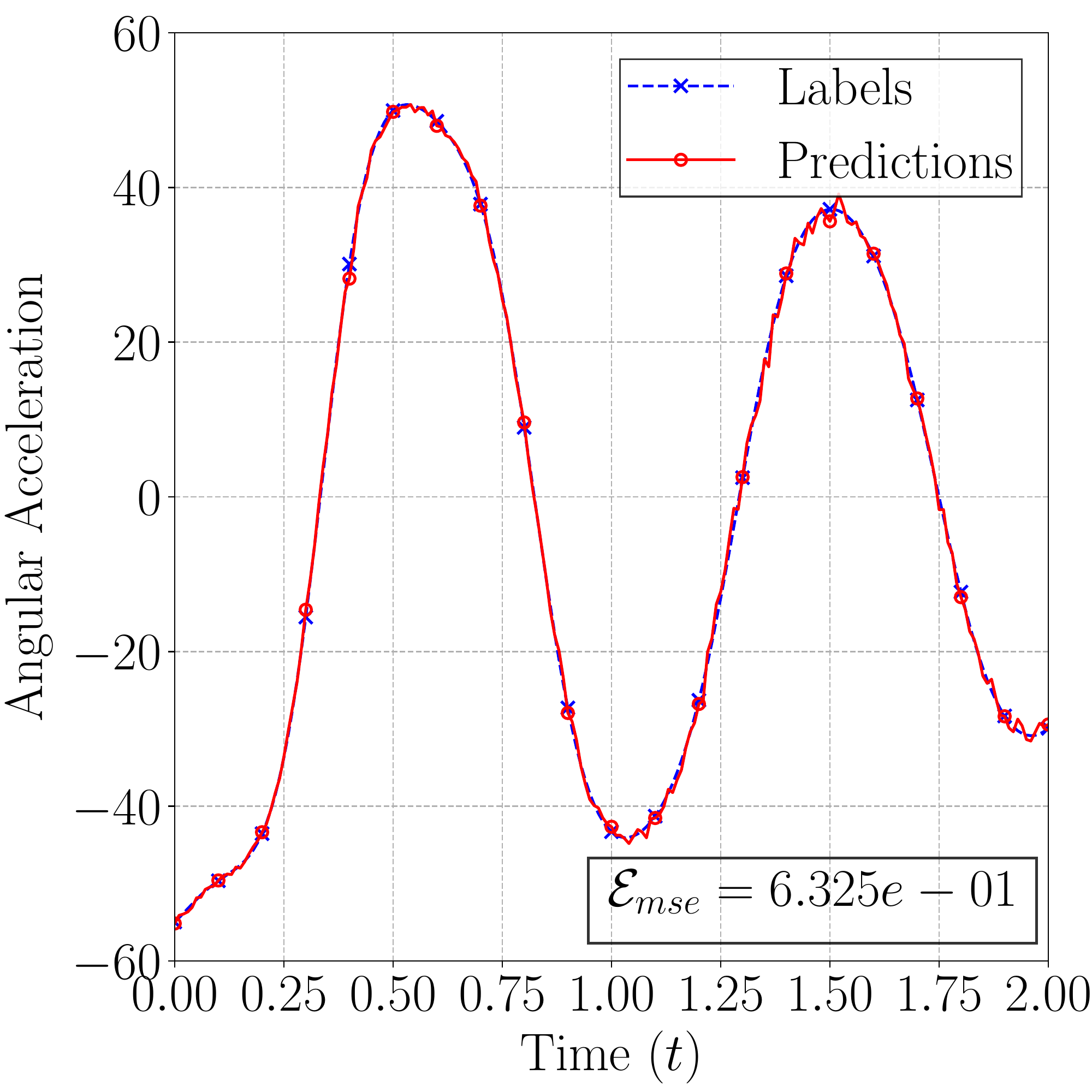}
  \includegraphics[width = \figsize\textwidth]
  {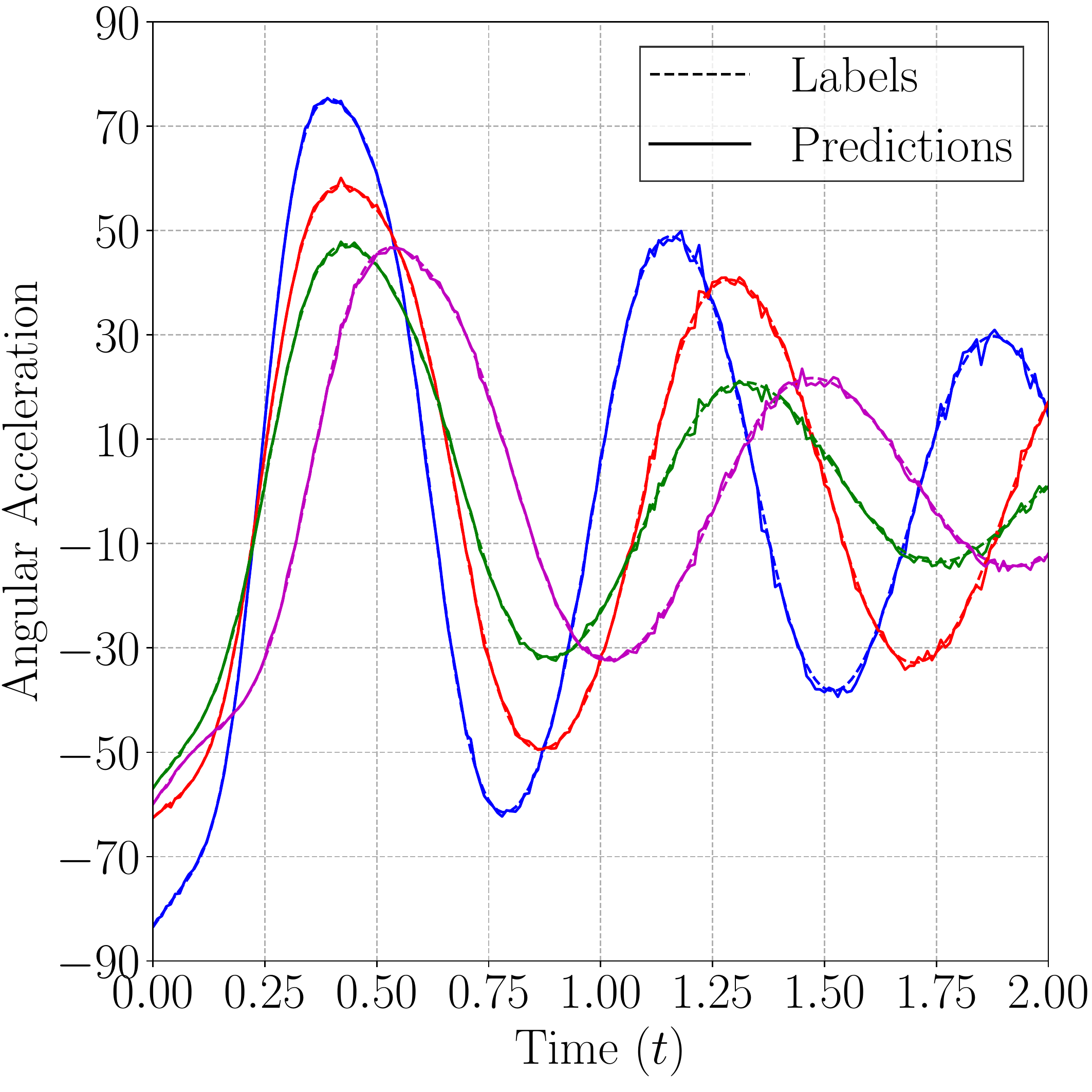}
\caption{
Dynamic responses of the damped single pendulum achieved from \Sfixed\, training data with hyper-parameters in Table \ref{tab:single:Sfixed:hyper-parameters}. While the results in Fig. \ref{fig:single:specific} (Left) and \ref{fig:single:multiple} (Left) employs the hyper-parameters of \Sfull, the present results uses the hyper-parameters from independent grid searches on $\#\{t_n\}=201$ numbers of \Sfixed\, models. While the accuracies of solutions are improved, the oscillations are still observed. 
}
\label{fig:single:newgridsearch}
\end{figure}

Obviously, compared to the hyper-parameters for \Sfull\, in Table \ref{tab:single:hyper-param}, those in Table \ref{tab:single:Sfixed:hyper-parameters} improves the performance of \Sfixed. The improved results corresponding to Fig. \ref{fig:single:specific} (Left) and \ref{fig:single:multiple} (Left) are shown in Fig. \ref{fig:single:newgridsearch} (Left) and Fig. \ref{fig:single:newgridsearch} (Right), respectively. Compared to the results shown in Fig. \ref{fig:single:specific} (Left) and \ref{fig:single:multiple} (Left), the accuracies of solutions from independent grid searches are clearly enhanced, which can be confirmed by the orders of $\MSE$. 

However, the oscillations are still observed, which yield less smooth solutions compared to the results of \Sfull, shown in Fig. \ref{fig:single:specific} (Right) and \ref{fig:single:multiple} (Right). In addition, $\#\{t_n\}$ numbers of grid searches for \Sfixed\, requires a heavy computational burden. The normalized clock time for grid search for \Sfull\, and \Sfixed\, are compared in Table \ref{tab:single:comp.cost}. 

Thus, the usage of the same hyper-parameters to both \Sfull\, and \Sfixed\, is not a serious hindrance to comparing performance of the two types of training data sets. For simplicity and computational feasibility, the hyper-parameters found from \Sfull\, for both \Sfull\, and \Sfixed\, are employed, in the numerical examples in Sections \ref{sec:double} and \ref{sec:slider}. 

\begin{table*}[h]
\centering \ra{1.3}
\begin{tabular}{@{}cccc@{}}
\toprule
\begin{tabular}[c]{@{}c@{}}
Model \\ for $t = t_n$ \end{tabular} 
& \begin{tabular}[c]{@{}c@{}}
The number of \\ hidden layers\end{tabular} 
& \begin{tabular}[c]{@{}c@{}}The number of nodes \\ per a hidden layer\end{tabular} 
& \begin{tabular}[c]{@{}c@{}}The size of \\ batch\end{tabular} 
\\
\midrule
$t=0.00$ & 2 & 256 & 128\\
$t=0.01$ & 2 & 256 & 64\\
$t=0.02$ & 2 & 256 & 128\\
$t=0.03$ & 2 & 256 & 64\\
$t=0.04$ & 2 & 256 & 128\\
$t=0.05$ & 2 & 256 & 128\\
$\vdots$ & $\vdots$ & $\vdots$ & $\vdots$ \\
$t=1.00$ & 3 & 256 & 128\\
$t=1.01$ & 4 & 256 & 128\\
$t=1.02$ & 4 & 256 & 64\\
$t=1.03$ & 3 & 256 & 128\\
$t=1.04$ & 4 & 256 & 128\\
$t=1.05$ & 2 & 256 & 128\\
$\vdots$ & $\vdots$ & $\vdots$ & $\vdots$ \\
$t=1.95$ & 4 & 128 & 128\\
$t=1.96$ & 4 & 128 & 128\\
$t=1.97$ & 3 & 256 & 64\\
$t=1.98$ & 3 & 256 & 128\\
$t=1.99$ & 3 & 256 & 128\\
$t=2.00$ & 4 & 256 & 128\\
\bottomrule
\end{tabular}
\caption{
Hyper-parameters for \Sfixed\, training data, which are achieved from independent grid searches for $\#\{t_n\}=201$ \Sfixed models.  
}
\label{tab:single:Sfixed:hyper-parameters}
\end{table*}

\begin{table*}
\centering \ra{1.3}
\begin{tabular}{@{}cccc@{}}
\toprule
Training data
& \begin{tabular}[c]{@{}c@{}}
The number of \\ models \end{tabular}  
& \begin{tabular}[c]{@{}c@{}}
The number of training data\\ per model \end{tabular} 
& \begin{tabular}[c]{@{}c@{}}
Normalized clock time \\ for grid searches\end{tabular} 
\\
\midrule
\Sfull & 1 & 267,531 & 1 \\
\Sfixed & 201& 1,331 & 18.3458\\
\bottomrule
\end{tabular}
\caption{
Comparison of data structures \Sfull\, and \Sfixed, and normalized clock times taken for independent grid searches. Grid searches for \Sfixed requires a heavy computational cost. 
}
\label{tab:single:comp.cost}
\end{table*}

\subsection{Double Pendulum}
\label{sec:double}
\begin{figure}
  \centering
  \includegraphics[width = 0.37\textwidth]
  {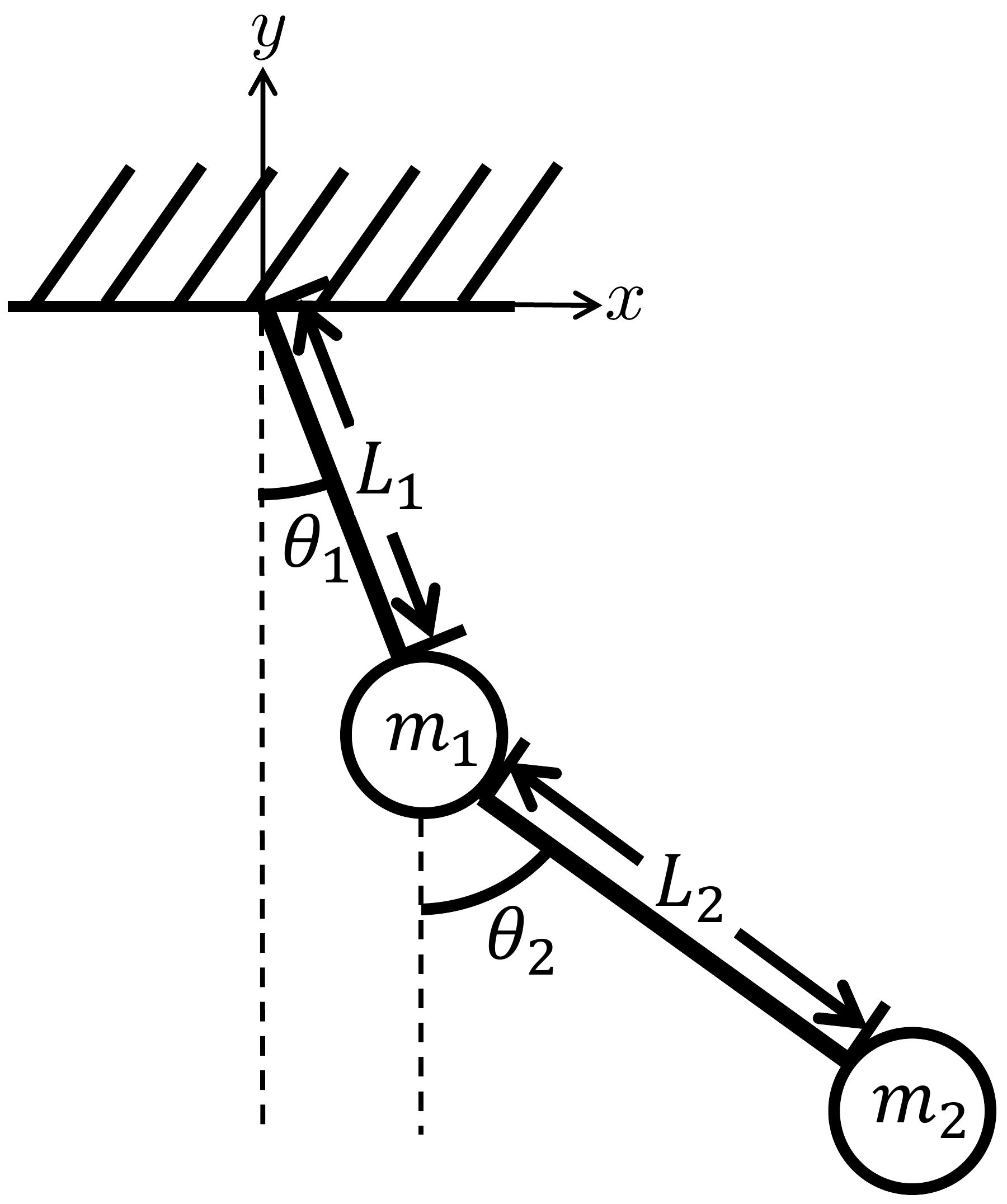}
\caption{Double pendulum problem. Gravity acceleration $g$, the masses $m_1$,$m_2$, and the initial angles $\theta^0_1$,$\theta^0_2$,are fixed to $g = 9.81[m/s^2]$, $m_1=2.0[kg],$ $m_2 = 1.0[kg]$, $\theta^0_1 = 1.6[rad]$, and $\theta^0_2 = 1.6[rad]$. The lengths of the massless rods $L_1[m] \in [1,2]$, $L_2[m] \in [2,3]$, and the initial angular velocities $\dot{\theta}^0_1[rad/s] \in [0, 0.1]$, $\dot{\theta}^0_2[rad/s] \in [0.3, 0.5]$ are arbitrarily determined within the given ranges. 
}
\label{fig:double:diagram}
\end{figure}

A double pendulum problem in Fig. \ref{fig:double:diagram} follows the given mathematical governing equation: 
\begin{equation}\begin{aligned}
\label{eq:double:governing}
\begin{cases}
  (m_1 + m_2)L_1\ddot{\theta}_1 
  + m_2 L_2 \ddot{\theta}_2 \cos(\theta_1 - \theta_2)
  + m_2 L_2 \dot{\theta}_2^2 \sin(\theta_1 - \theta_2)
  + (m_1 + m_2)g \sin(\theta_1) 
  = 0, 
  \\
  m_2 L_2 \ddot{\theta}_2
  + m_2 L_1 \ddot{\theta}_1
  \cos(\theta_1 - \theta_2)
  - m_2 L_1 \dot{\theta}_1^2 \sin(\theta_1 - \theta_2)
  + m_2 g \sin(\theta_2)
  = 0,
  \\
  \theta_i(t) = \theta_i^0,~
  \dot{\theta}_i(t) = \dot{\theta}_i^0, 
  \quad \text{where}\quad  t = 0 ,
  \quad i = 1, 2. 
\end{cases}
\end{aligned}\end{equation}
where $\theta_i = \theta_i(t)$ and $t \in [0,t_f]$ $,i = 1, 2,$ represent the time-varying angles of the links as shown in Fig. \ref{fig:double:diagram}. 
Parameters $g$ is the gravity constant, $L_i$ is the length of the massless rod $i$, $m_i$ is the mass, $\theta^0_i$ is the initial angle, $\dot{\theta}^0_i$ is the the initial angular velocity, and $i = 1,2,$ body notation, respectively.

In the meta-modeling, it is assumed that $(L_1, L_2, \dot{\theta}^0_1, \dot{\theta}^0_2)$ are independent input parameters and $(\theta_1, \theta_2, \dot{\theta}_1, \dot{\theta}_2)$ are output parameters. As in the single pendulum problem \eqref{eq:single:governing}, inputs are chosen within some ranges. The other parameters are fixed to given constants.
More details on ranges and mesh sizes of parameters are summarized in Table \ref{tab:double:param}. 

As in the previous numerical example, two types of training data, i.e. \Sfixed\, and \Sfull\, are compared. For \Sfixed, there are $\#\{t_n\} = 501$ meta-models, where each model is trained from $14,641$ numbers of data set. For \Sfull, there is only one meta-model trained from $14,641\times 501 = 7,335,141$ numbers of data set. 
For both \Sfixed\, and \Sfull\, types of training data, hyper-parameters are found as in Table \ref{tab:double:hyper-param}. 
\begin{table*}[h]
\centering \ra{1.15}
\begin{tabular}{@{}lc@{}}
\toprule
~~Hyper-parameters~
& ~~Choice~~~
\\
\midrule
The number of hidden layers & 4\\
The number of nodes in each layer & 64\\
The size of batch & 1024\\
The number of epochs & 400\\
Loss function & $\MSE$\\
Optimizer & Adam\\
\bottomrule
\end{tabular}
\caption{Hyper-parameters for the double pendulum problem }
\label{tab:double:hyper-param}
\end{table*}

\begin{figure}
  \centering
  \includegraphics[width = \figsize\textwidth]
  {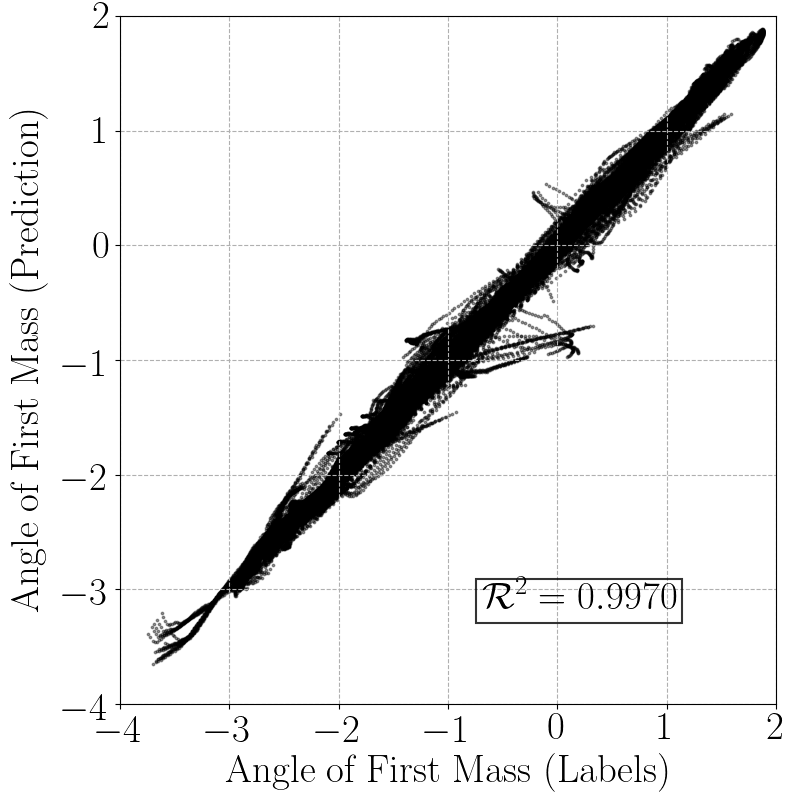}
  \includegraphics[width = \figsize\textwidth]
  {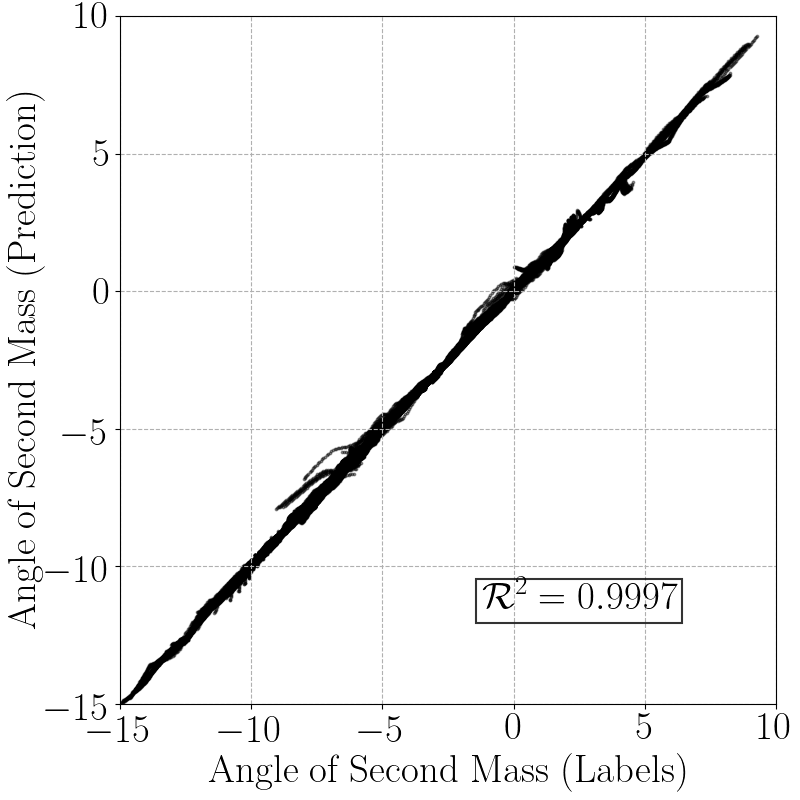}
  \\
  \includegraphics[width = \figsize\textwidth]
  {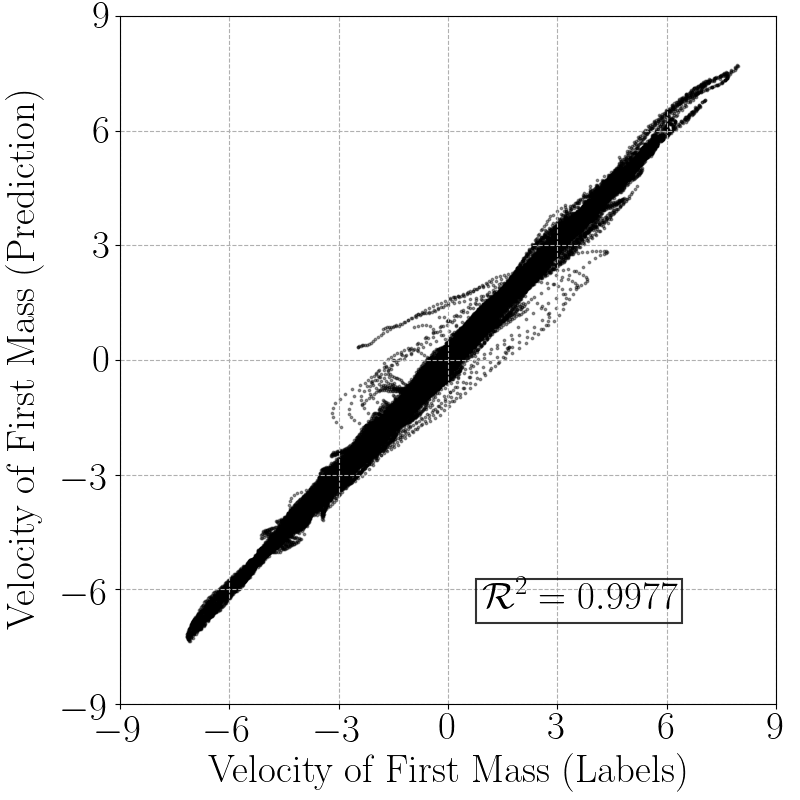}
  \includegraphics[width = \figsize\textwidth]
  {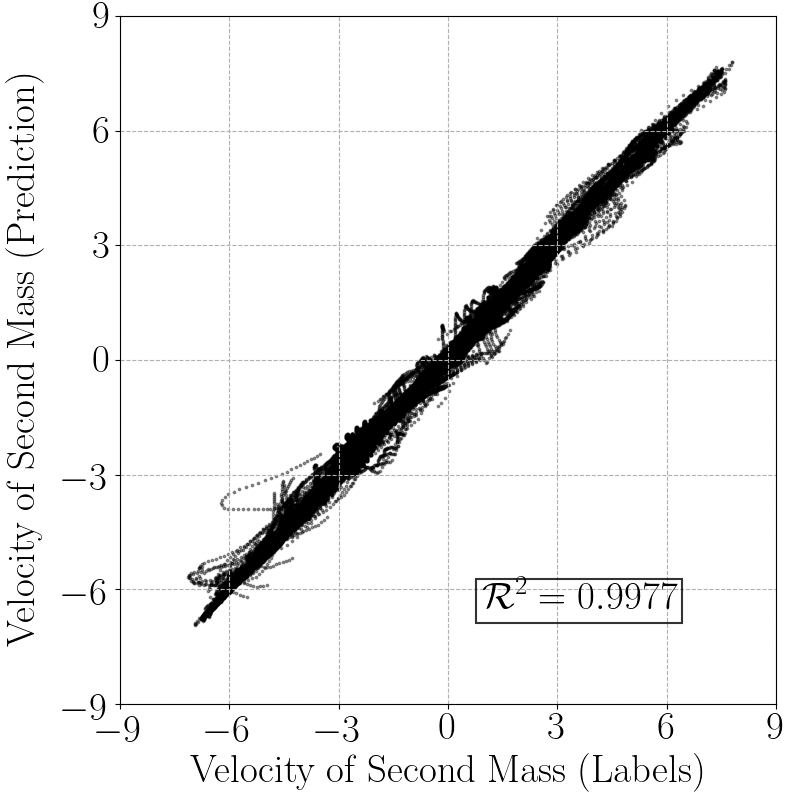}
\caption{Labels vs. Predictions for test data. The meta-model for the double pendulum problem is generated from \Sfull\, type of training set. Test data are {\it{unseen}} from training. The $\R$ scores are almost 1, which implies that the meta-model yields accurate solutions.}
\label{fig:double:scatter}
\end{figure}

The scatter plots in Fig. \ref{fig:double:scatter} show 
that a meta-model from \Sfull\, predicts output parameters $(\theta_1, \theta_2, \dot{\theta}_1, \dot{\theta}_2)$ with a great accuracy. The $\R$ values are over $0.997$ in all cases of solutions. 
\\

\begin{figure}
  \centering
  \includegraphics[width = \figsize\textwidth]
  {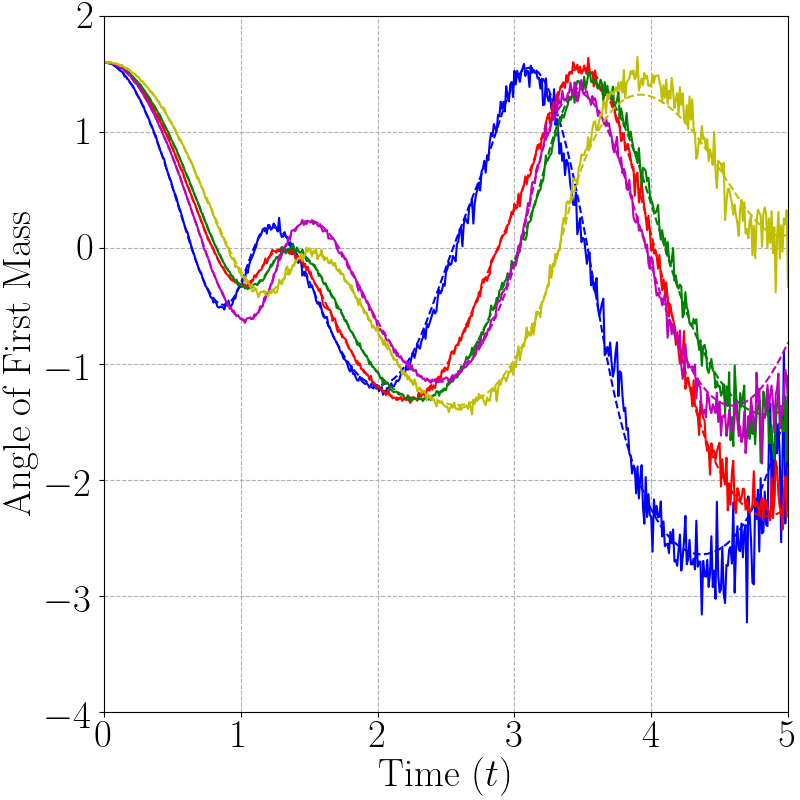}
  \includegraphics[width = \figsize\textwidth]
  {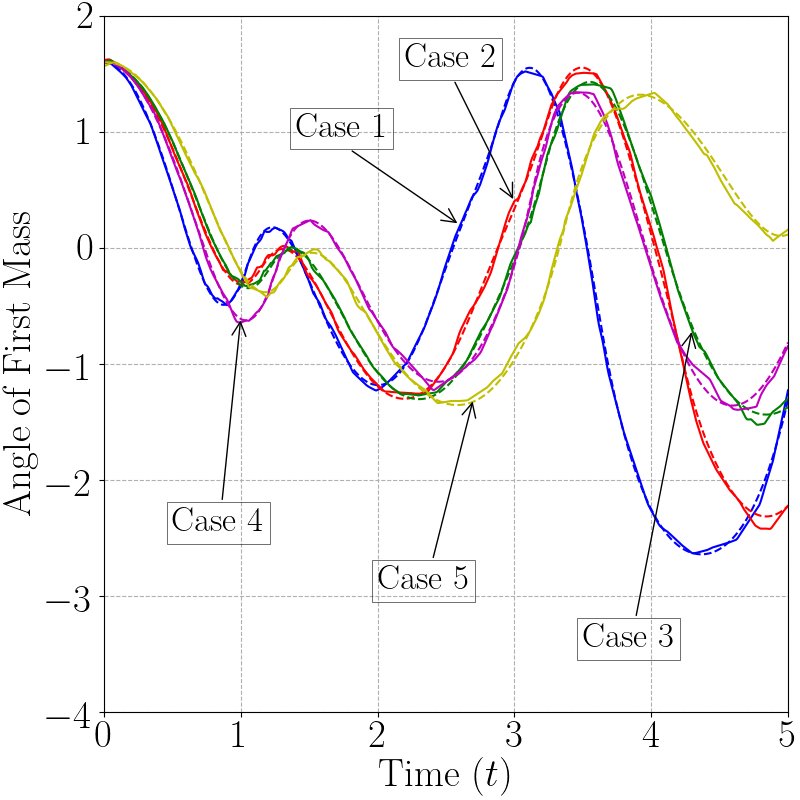}
  \includegraphics[width = \figsize\textwidth]
  {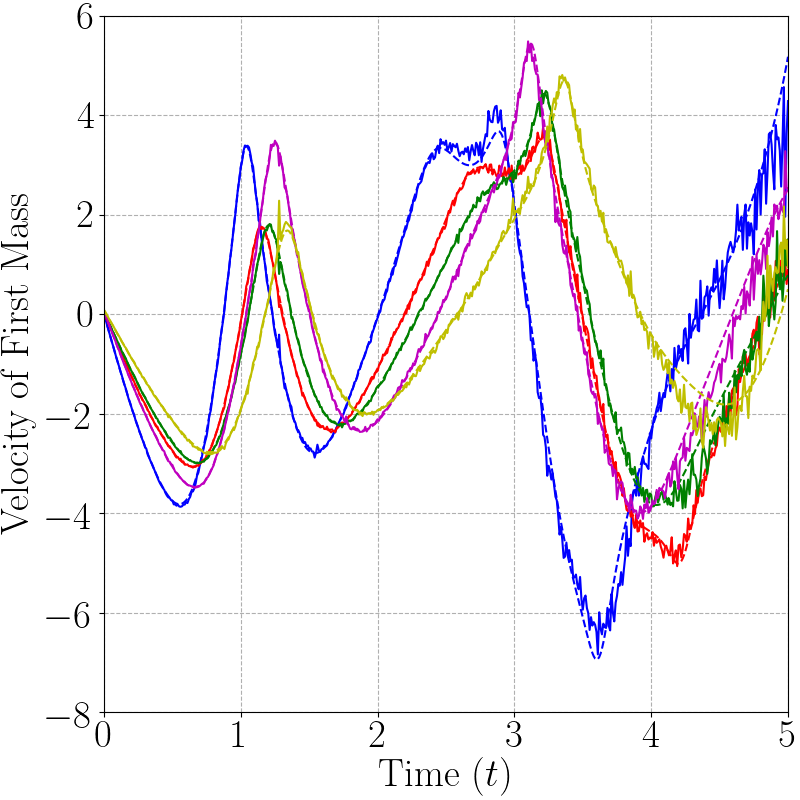}
  \includegraphics[width = \figsize\textwidth]
  {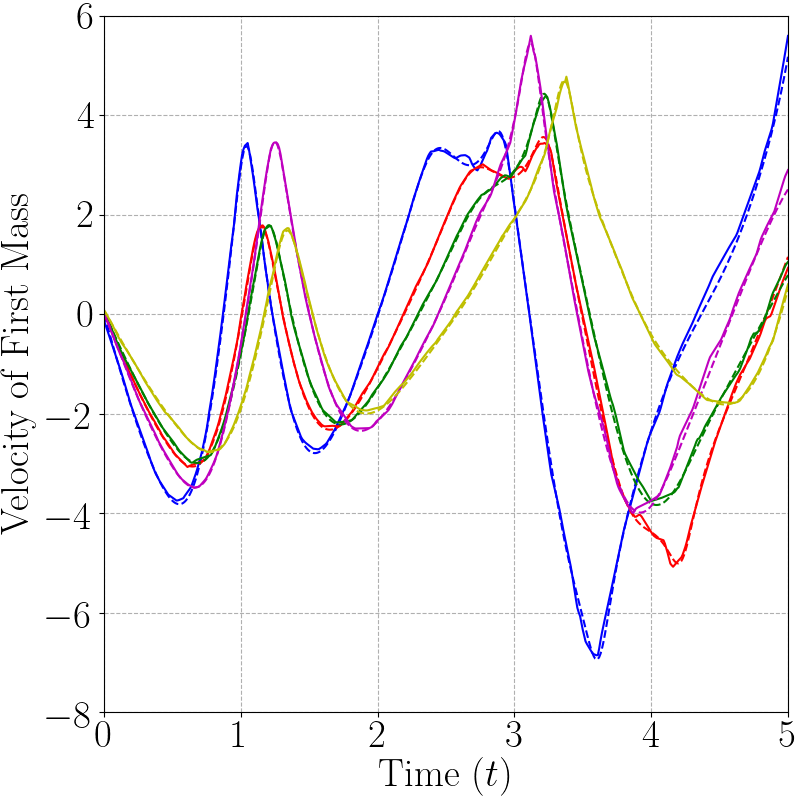}
\caption{Dynamic responses of double pendulum for multiple cases of input parameters:
\inputDouble{1.010}{2.130}{0.0}{0.3} (blue), 
\inputDouble{1.500}{2.410}{0.03}{0.330} (red), 
\inputDouble{1.620}{2.560}{0.044}{0.384} (green), 
\inputDouble{1.330}{2.820}{0.062}{0.412} (magenta),
\inputDouble{1.980}{2.940}{0.087}{0.470} (yellow).
Labels(dashed) and predictions(solid) are given for test data. 
Results from two types of training set \Sfixed (Left) and \Sfull (Right) are compared. Oscillations from \Sfixed\, becomes more severe than the case of single pendulum shown in Fig. \ref{fig:single:multiple}.
}
\label{fig:double:multiple:mass 1}
\end{figure}

\begin{figure}
  \centering
  \includegraphics[width = \figsize\textwidth]
  {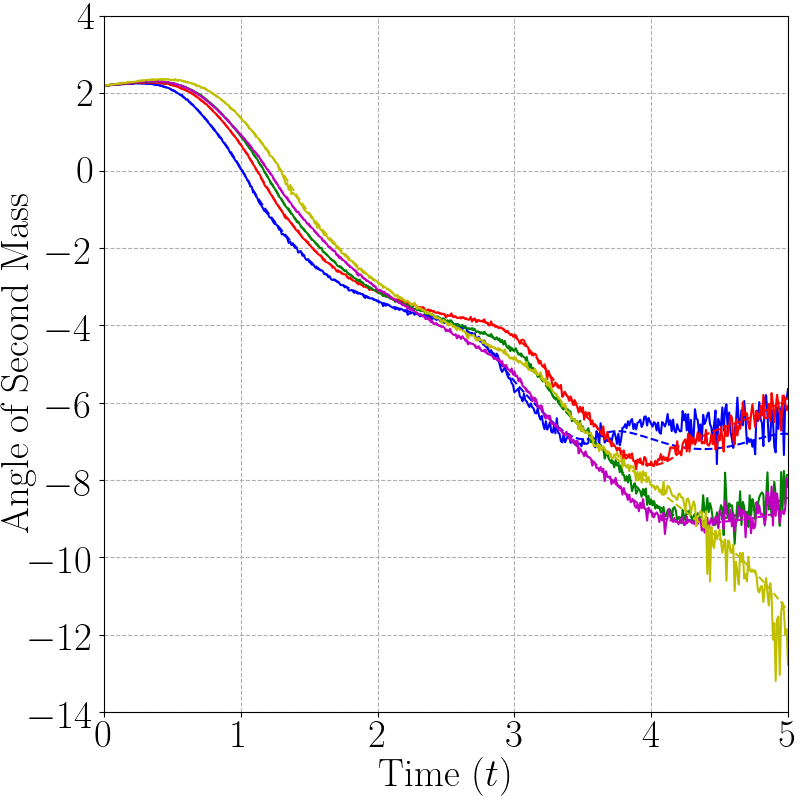}
  \includegraphics[width = \figsize\textwidth]
  {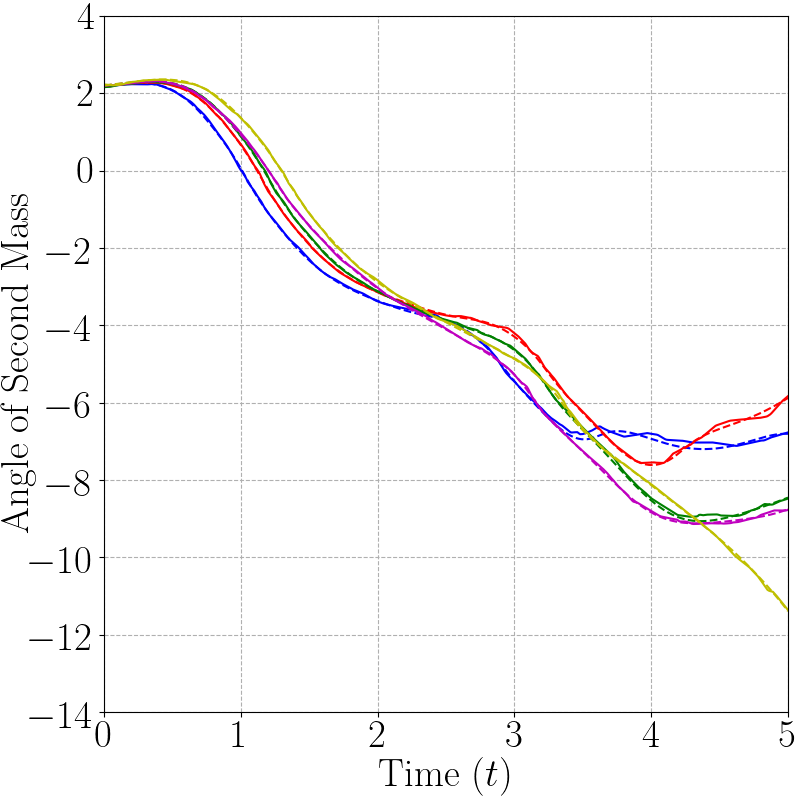}
  \includegraphics[width = \figsize\textwidth]
  {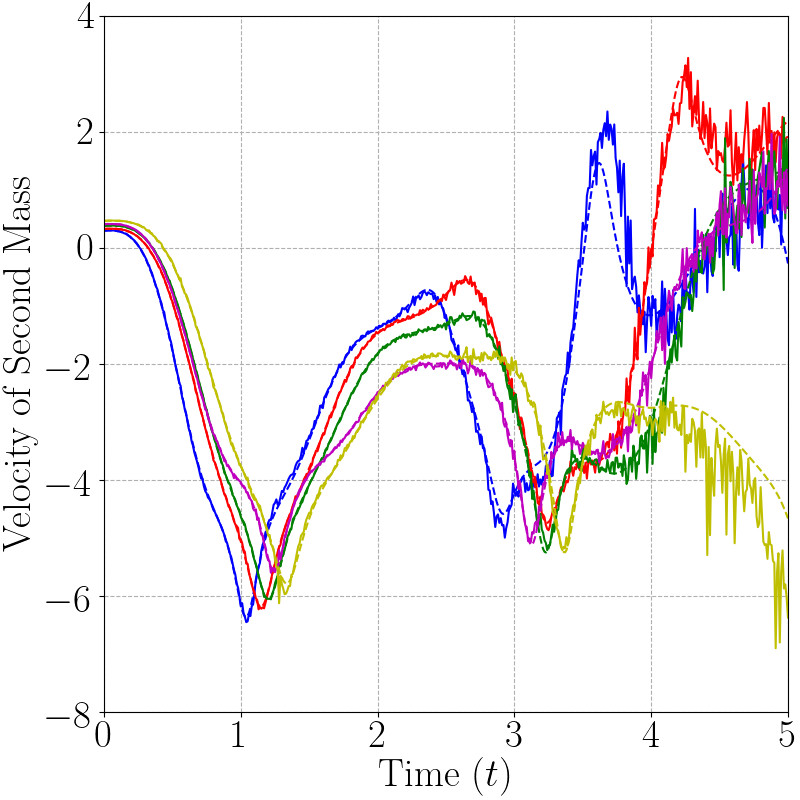}
  \includegraphics[width = \figsize\textwidth]
  {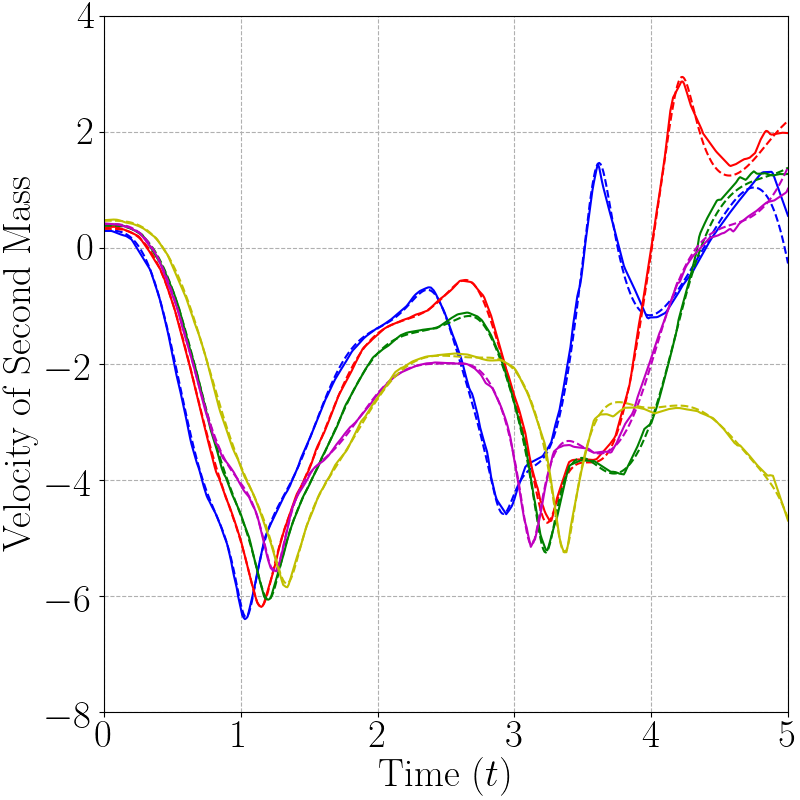}
\caption{Dynamic responses of double pendulum for multiple cases of input parameters:
\inputDouble{1.010}{2.130}{0.0}{0.3} (blue), 
\inputDouble{1.500}{2.410}{0.03}{0.330} (red), 
\inputDouble{1.620}{2.560}{0.044}{0.384} (green), 
\inputDouble{1.330}{2.820}{0.062}{0.412} (magenta),
\inputDouble{1.980}{2.940}{0.087}{0.470} (yellow).
Labels(dashed) and predictions(solid) are given for test data. 
Results from two types of training set \Sfixed (Left) and \Sfull (Right) are compared. Oscillations from \Sfixed\, becomes more severe than the case of single pendulum shown in Fig. \ref{fig:single:multiple}
}
\label{fig:double:multiple:mass 2}
\end{figure}

Performances of meta-models from \Sfixed\, and \Sfull\, types of training data are compared in Fig. \ref{fig:double:multiple:mass 1} and \ref{fig:double:multiple:mass 2}. It shows dynamic changes of predictions (solid) from meta-models in comparison with their labels (dashed), for multiple cases as shown in Table \ref{tab:double:multiple}. 
\begin{table*}[ht]
\centering \ra{1.1}
\begin{tabular}{@{}cccccc@{}}
\toprule
& $L_1~[m]$
& $L_2~[m]$
& $\dot{\theta}^0_1~[rad/s]$
& $\dot{\theta}^0_2~[rad/s]$
\\
\midrule
Case 1
& 1.010 & 2.130 & 0.00 & 0.300 \\
Case 2
& 1.500 & 2.410 & 0.03 & 0.330 \\
Case 3
& 1.620 & 2.560 & 0.044 & 0.384 \\
Case 4
& 1.330 & 2.820 & 0.062 & 0.412 \\
Case 5
& 1.980 & 2.940 & 0.087 & 0.470 \\
\bottomrule
\end{tabular}
\caption{Input parameters of multiple cases for Fig. \ref{fig:double:multiple:mass 1} and \ref{fig:double:multiple:mass 2}. }
\label{tab:double:multiple}
\end{table*}

As observed in single pendulum cases shown in Fig. \ref{fig:single:specific} and \ref{fig:single:multiple}, the meta-model from \Sfixed\, shows lots of oscillations in its dynamic responses. Here the oscillations are quite severe, especially when $t$ is large. Though these results can be improved if more appropriate hyper-parameters are employed for each of $\#\{t_n\}$ number of meta-models, the grid searches are computationally infeasible. 
On the other hand, meta-model from \Sfull\, yields more accurate and smooth dynamic responses. 
\\

\begin{figure}
  \centering
  \includegraphics[width = \figsmall\textwidth]
  {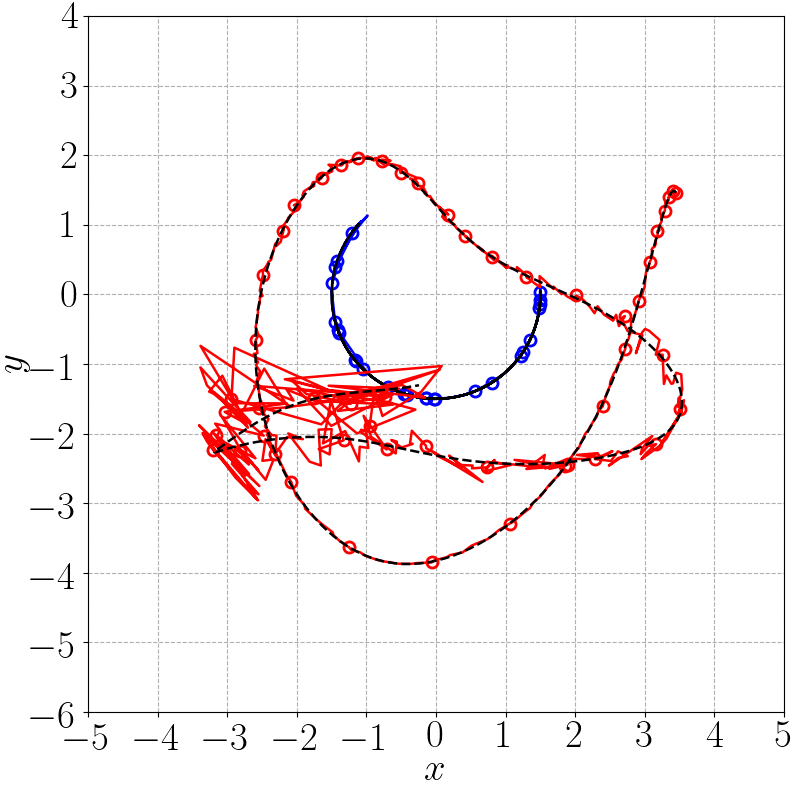}
  \includegraphics[width = \figsmall\textwidth]
  {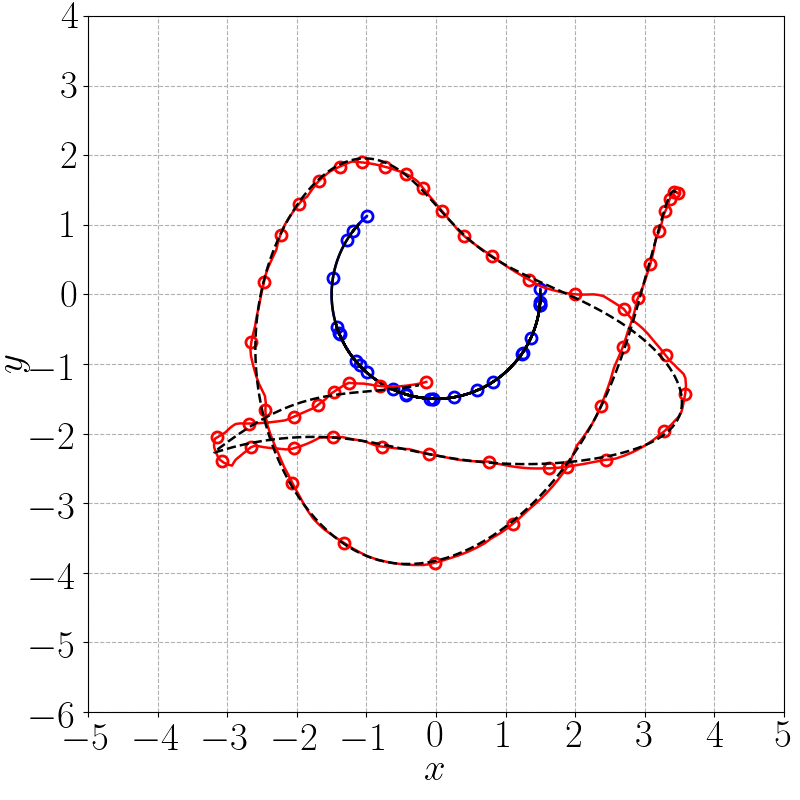}
  \includegraphics[width = \figsmall\textwidth]
  {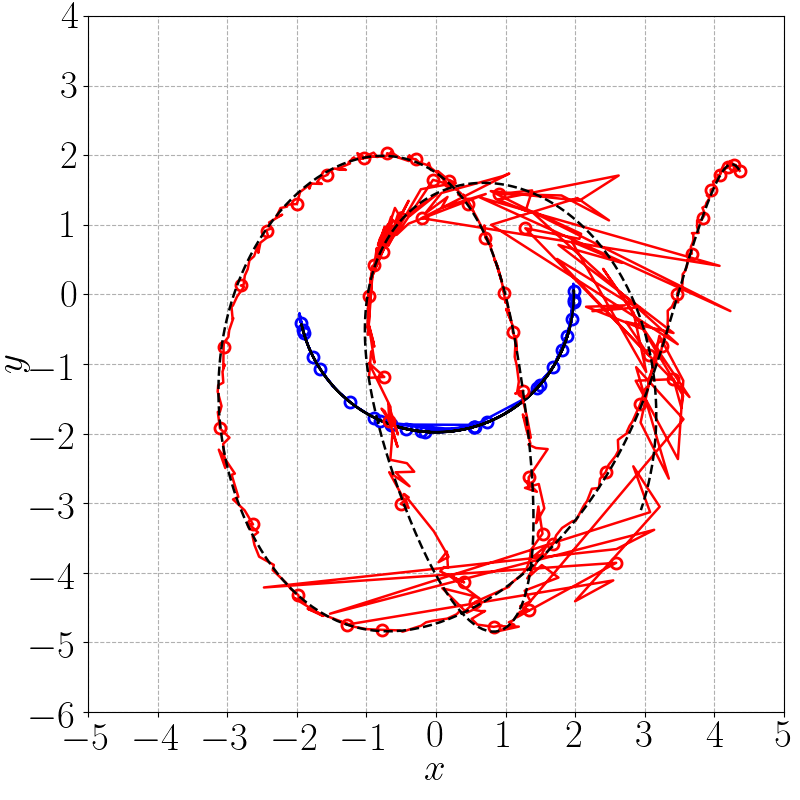}
  \includegraphics[width = \figsmall\textwidth]
  {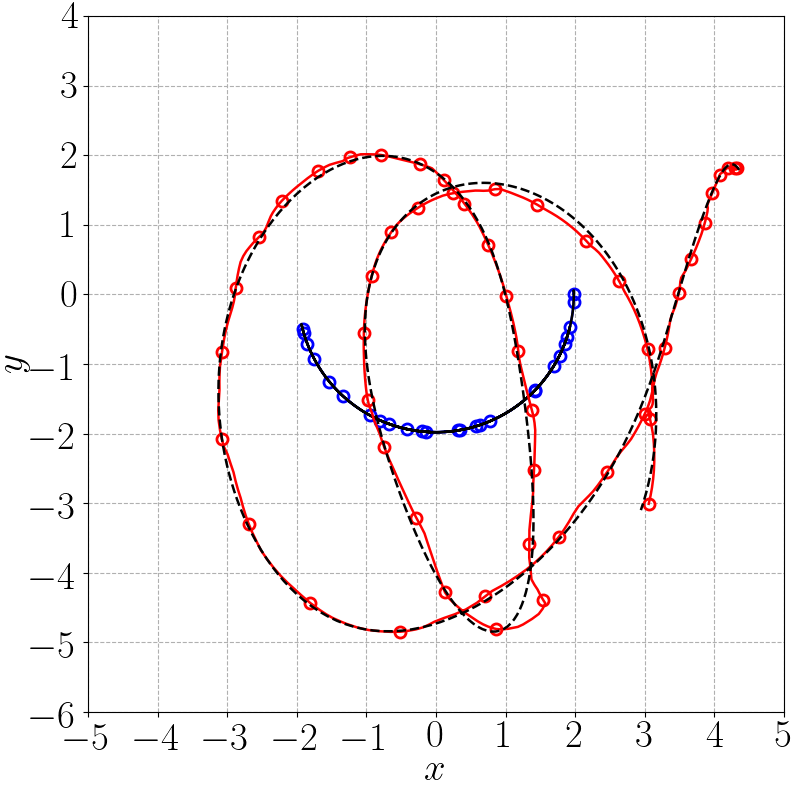}
  \includegraphics[width = \figsmall\textwidth]
  {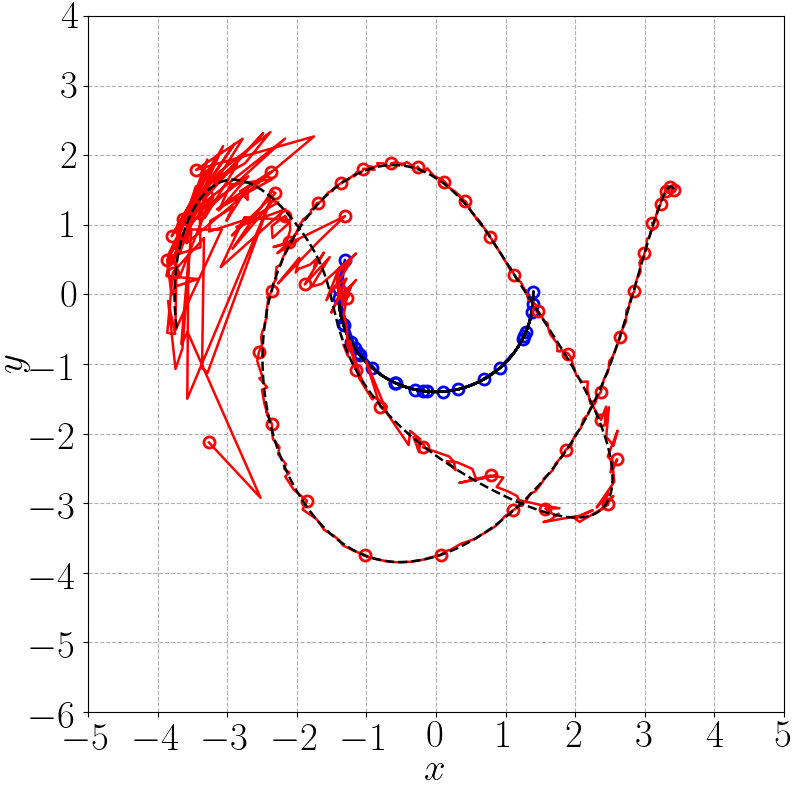}
  \includegraphics[width = \figsmall\textwidth]
  {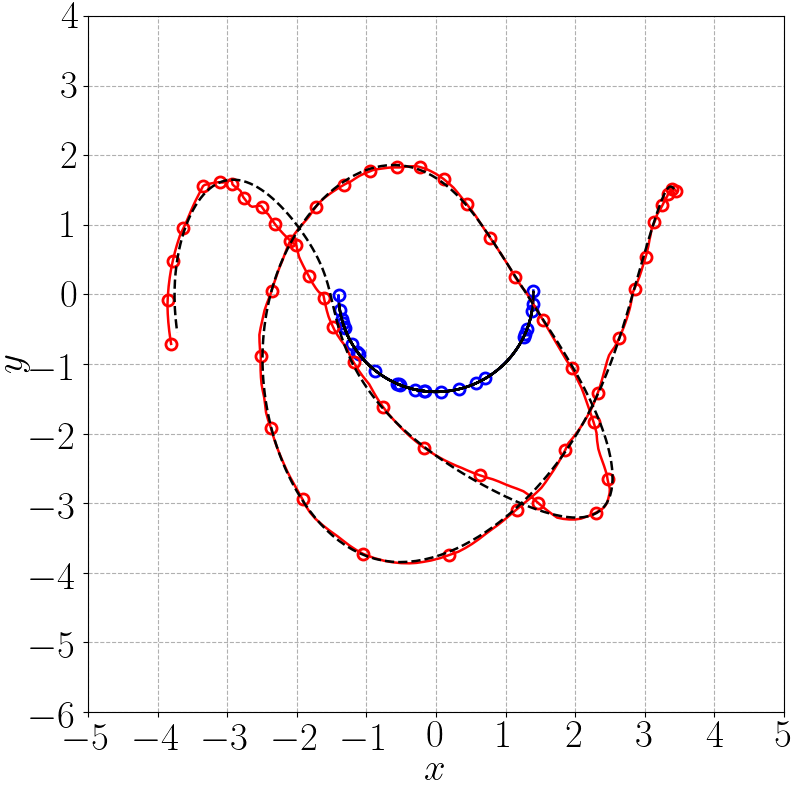}
\caption{Trajectories of masses $m_1$ and $m_2$ for double pendulum problems: Labels($m_1$: black solid, $m_2$: black dashed) vs. Predictions($m_1$:blue solid,circles, $m_2$:red solid,circles) for multiple inputs 
\inputDouble{1.500}{2.410}{0.03}{0.330} (Top), 
\inputDouble{1.980}{2.940}{0.087}{0.470} (Middle), 
\inputDouble{1.400}{2.500}{0.060}{0.380} (Bottom).
Results from two types of training set \Sfixed (Left) and \Sfull (Right) are compared.  
}
\label{fig:double:traj}
\end{figure}

Difference between two training data set \Sfixed\, and \Sfull\, is shown more clearly in Fig. \ref{fig:double:traj}, where trajectories of two masses $m_1$ and $m_2$ are shown. Labels ($m_1$: black solid, $m_2$: black dashed) and predictions ($m_1$:blue solid,circles, $m_2$:red solid,circles) are given for the results from \Sfixed (Left) and \Sfull (Right). Each plot is from a particular input parameters: 
\inputDouble{1.500}{2.410}{0.03}{0.330} (Top), 
\inputDouble{1.980}{2.940}{0.087}{0.470} (Middle), 
\inputDouble{1.400}{2.500}{0.060}{0.380} (Bottom).

\begin{table*}
\centering \ra{1.1}
\begin{tabular}{@{}lllll@{}}
\toprule
& Parameters
& Ranges
& \begin{tabular}[c]{@{}c@{}}Meshsizes for \\ Training Data\end{tabular} 
& \begin{tabular}[c]{@{}c@{}}Meshsizes for \\ Test Data\end{tabular} 
\\
\midrule
Fixed constants
& $g~[m/s^2]$ & $9.81$ & $\cdot$ & $\cdot$\\
& $m_1~[kg]$ & $2.0$ & $\cdot$ & $\cdot$\\
& $m_2~[kg]$ & $1.0$ & $\cdot$ & $\cdot$\\
& $\theta^0_1~[rad]$ & $1.6$ & $\cdot$ & $\cdot$\\
& $\theta^0_2~[rad]$ & $1.6$ & $\cdot$ & $\cdot$\\
Inputs
& $L_1~[m]$ & $[1, ~2]$ & $\Delta{L_1}=0.1$ & 
arbitrary(not uniform)\\
& $L_2~[m]$ & $[2, ~3]$ & $\Delta{L_2}=0.1$ & 
arbitrary(not uniform)\\
& $\dot{\theta}^0_1~[rad/s]$ & $[0, ~0.1]$ & $\Delta{\dot{\theta}^0_1}=0.01$ & arbitrary(not uniform)\\
& $\dot{\theta}^0_2~[rad/s]$ & $[0.3, ~0.5]$ & $\Delta{\dot{\theta}^0_2}=0.02$ & arbitrary(not uniform)\\
Time instants 
& $\{t_n\}~[s]$ & $[0, 5]$ & $\Delta{t}=0.01$\,$(t_0 = 0)$ & $\Delta{t}=0.01$\,$(t_0 = 0)$ \\
\bottomrule
\end{tabular}
\caption{Summary on parameters of double pendulum problem. In \Sfixed, a fixed time instant is considered. In \Sfull, all the time instants are treated as inputs.}
\label{tab:double:param}
\end{table*}

\subsection{Slider Crank Mechanism}
\label{sec:slider}
\begin{figure}[H]
  \centering
  \includegraphics[width = 0.7\textwidth]
  {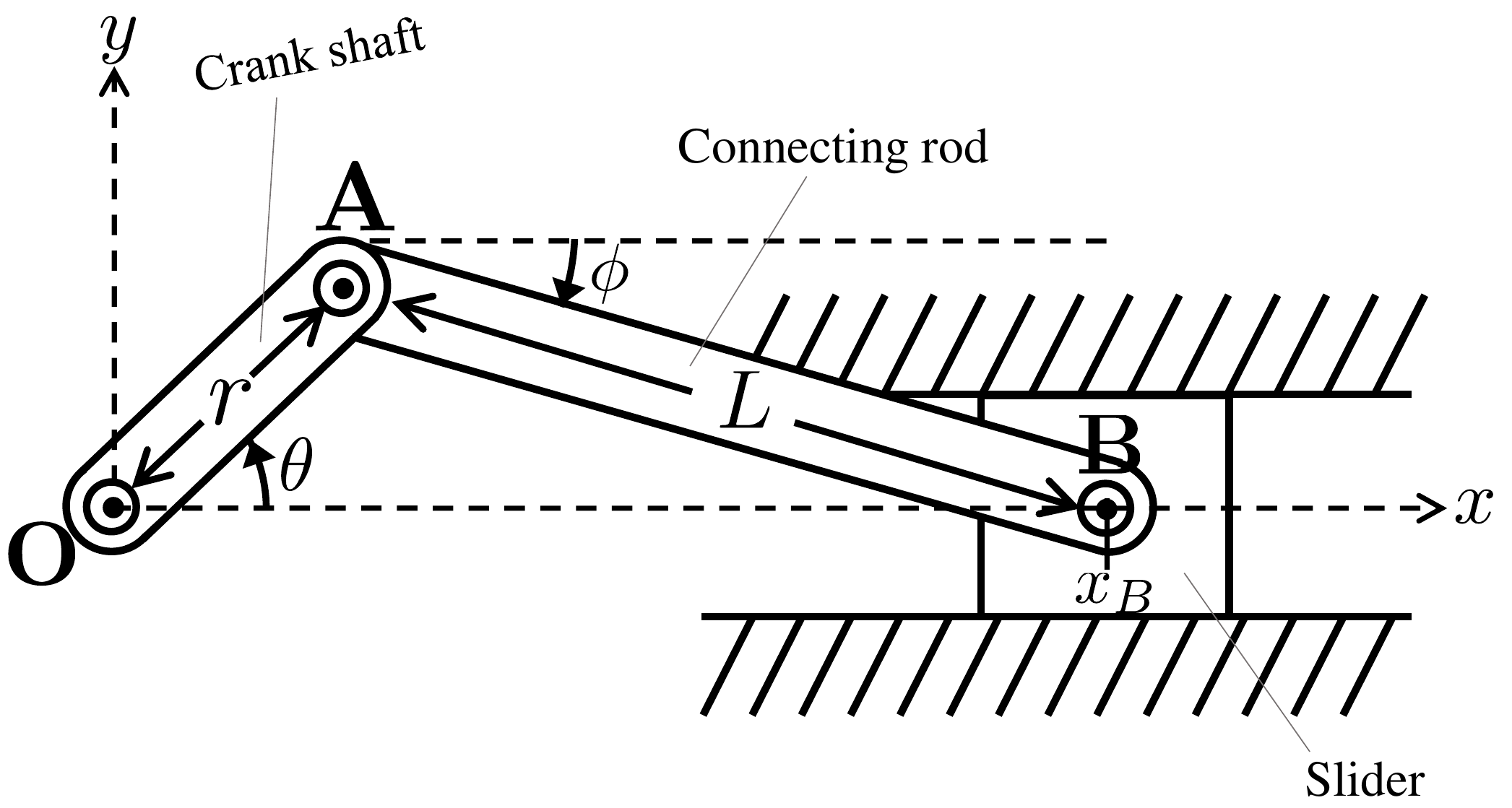}
\caption{Slider crank mechanism. 
Gravity acceleration $g$ is fixed to $g=9.81[m/s^2]$. The constant $\tau \in [1, 2]$, the length of crank shaft $r[m] \in [1, 3]$, the ratio of the lengths between the connecting rod and the crank shaft $L/r \in [2.5, 3.5]$ are arbitrarily determined within the given ranges. 
}
\label{fig:slider:diagram}
\end{figure}

Consider a slider crank in Fig. \ref{fig:slider:diagram}, where parameters $(r, L, \theta(t), \phi(t))$ represent, respectively, the length of the massless crank shaft$[m]$, the length of the massless connecting rod$[m]$, the angle of the crank shaft$[rad]$, and the angle of the connecting rod$[rad]$. The initial angle $\theta^0$$[rad]$ and the initial velocity $\dot{\theta}^0$$[rad/s]$ are assumed as zeros, and the angular acceleration of the crank shaft $\ddot{\theta}(t)$$[rad/s^2]$ is given as
\begin{equation}\begin{aligned}
  \theta(t) &= \theta^0 = 0,
  \quad \text{where}~ t = 0,
  \\
  \dot{\theta}(t) &= \dot{\theta^0} = 0,
  \quad \text{where}~ t = 0, 
  \\
  \ddot{\theta}(t) &= \sin(\tau\,t), 
  \quad \text{where}~ t \in [0, t_f],
\end{aligned}\end{equation}  
for some constant $\tau \in \mathbb{R}$. 

\noindent
Then the angle of the crank shaft $\theta(t)$ and its temporal derivatives can be rewritten explicitly, for $t \in [0, t_f]$, 
\begin{equation}\begin{aligned}
  \theta(t) 
  &= -\frac{1}{\tau^2} \sin(\tau\,t) + \frac{t}{\tau} + \theta^0
  = -\frac{1}{\tau^2} \sin(\tau\,t) + \frac{t}{\tau},
  \\
  \dot{\theta}(t) 
  &= -\frac{1}{\tau}\cos(\tau\,t) + \frac{1}{\tau} + \dot{\theta}^0
  = -\frac{1}{\tau}\cos(\tau\,t) + \frac{1}{\tau},
\end{aligned}\end{equation}
In DNN modeling, three independent parameters $(\tau, r, L/r)$ are considered as inputs, while time variable $t$ can be fixed to an instant (\Sfixed) or considered as an input (\Sfull). More details on ranges and mesh sizes of parameters are summarized in Table \ref{tab:slider:param}. 

Although the slider crank mechanism is not a dynamic problem, this kinematic example is a good example because the kinematics should be treated as a special case of dynamic problems. To describe kinematics of the slider crank, seven kinematic solutions $\theta$, $\phi$, $\dot{\phi}$, $\ddot{\phi}$, $x_B$, $\dot{x}_B$, and $\ddot{x}_B$ are considered as an output parameters, where $x_B$ denotes the $x$-directional translation of the slider. 

The output solutions other than $(\theta, \dot{\theta}, \ddot{\theta})$ can be found from kinematic equations as follows:
\begin{equation}\begin{aligned}
\label{eq:slider:governing}
  \begin{bmatrix}
  \phi \\ x_B
  \end{bmatrix}
  = \begin{bmatrix}
  \sin^{-1}\left(-(r/L) \sin\theta\right)
  \\
  r\cos\theta + L\cos\phi
  \end{bmatrix},
\end{aligned}\end{equation}

\begin{equation}\begin{aligned}
  \begin{bmatrix}
  \dot{\phi} \\ \dot{x_B}
  \end{bmatrix}
  = \begin{bmatrix}
  L\sin\phi && 1 \\ -L\cos\phi && 0 
  \end{bmatrix}^{-1}
  \begin{bmatrix}
  -r\dot{\theta} \sin\theta \\  r\dot{\theta}\cos\theta
  \end{bmatrix},
\end{aligned}\end{equation}
and 
\begin{equation}\begin{aligned}
  \begin{bmatrix}
  \ddot{\phi} \\ \ddot{x_B}
  \end{bmatrix}
  = \begin{bmatrix}
  L\sin\phi && 1 \\ -L\cos\phi && 0 
  \end{bmatrix}^{-1}
  \left(
  \begin{bmatrix}
  -L\, \dot{\phi} \cos\phi && 0 
  \\ -L\, \dot{\phi} \sin\phi && 0 
  \end{bmatrix}
  \begin{bmatrix}
  \dot{\phi} \\ \dot{x_B}
  \end{bmatrix}
  + 
  \begin{bmatrix}
  -r\dot{\theta}^2 \cos\theta \\  -r\dot{\theta}^2\sin\theta
  \end{bmatrix}
  + 
  \begin{bmatrix}
  -r\ddot{\theta}\sin\theta \\ r\ddot{\theta}\cos\theta
  \end{bmatrix}
  \right).
\end{aligned}\end{equation}

Two meta-models are generated from \Sfixed\, and \Sfull\, types of training data, by employing the hyper-parameters found from grid searches for the case of \Sfull shown in Table \ref{tab:slider:hyper-param}. 
\begin{table*}[h]
\centering \ra{1.15}
\begin{tabular}{@{}lc@{}}
\toprule
~~Hyper-parameters~
& ~~Choice~~~
\\
\midrule
The number of hidden layers & 2\\
The number of nodes in each layer & 128\\
The size of batch & 64\\
The number of epochs & 200\\
Loss function & $\MSE$\\
Optimizer & Adam\\
\bottomrule
\end{tabular}
\caption{Hyper-parameters for the slider crank problem }
\label{tab:slider:hyper-param}
\end{table*}

\begin{figure}
  \centering
  \includegraphics[width = \figsize\textwidth]
  {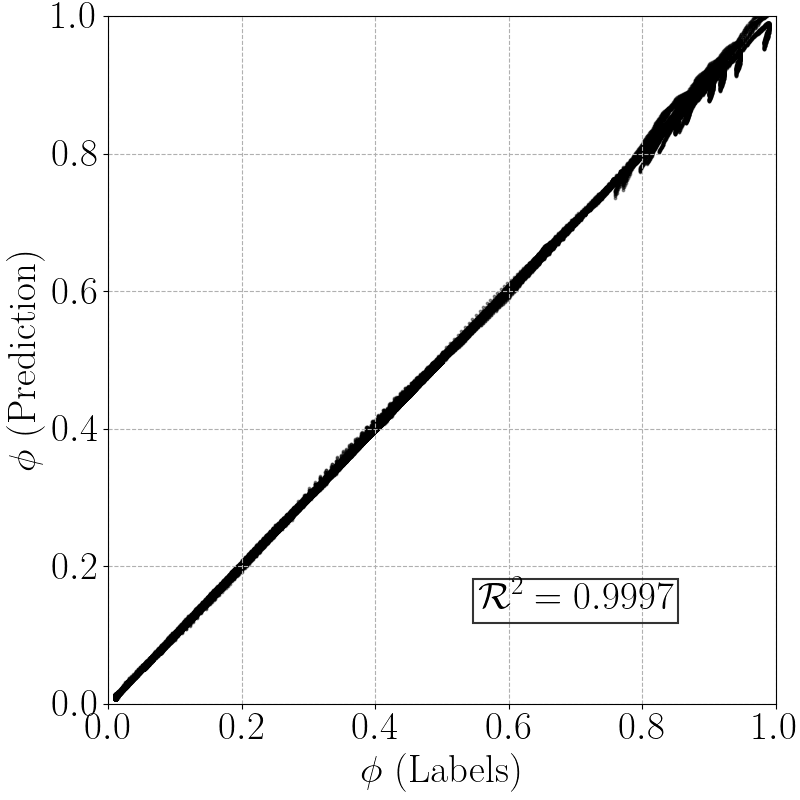}
  \includegraphics[width = \figsize\textwidth]
  {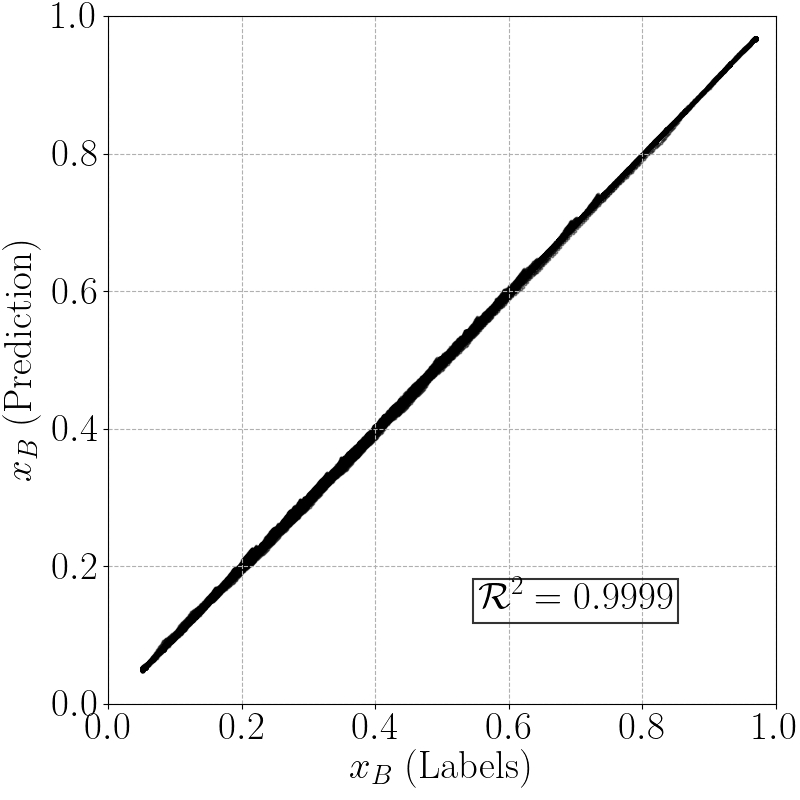}
  \includegraphics[width = \figsize\textwidth]
  {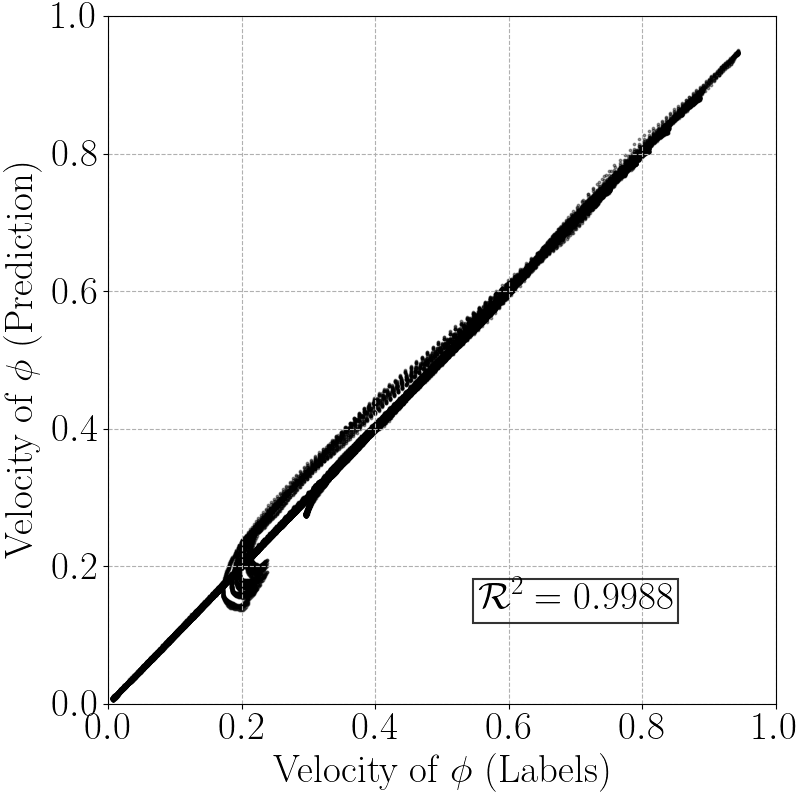}
  \includegraphics[width = \figsize\textwidth]
  {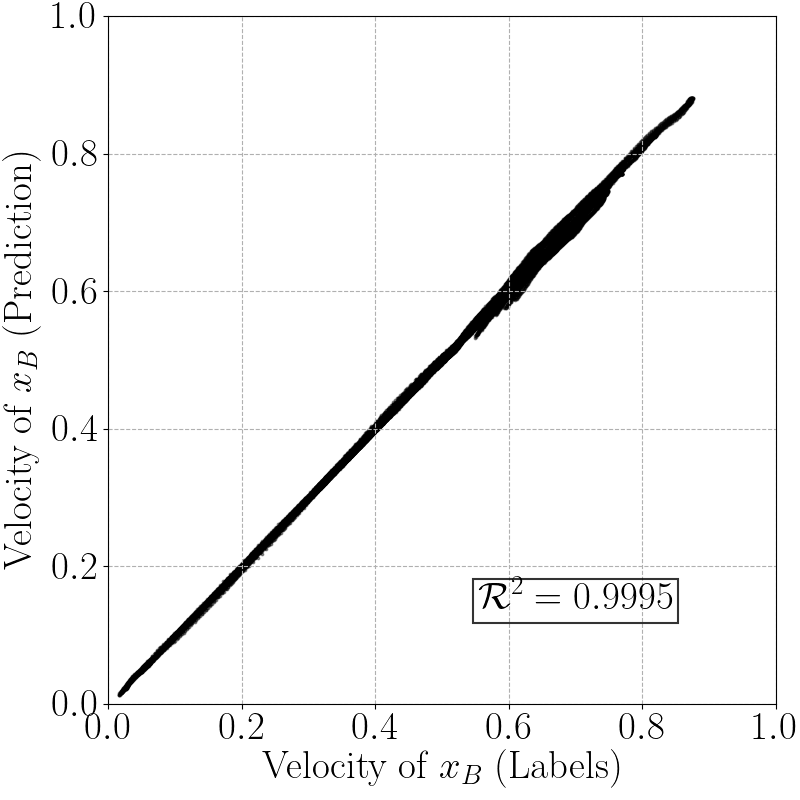}
  \includegraphics[width = \figsize\textwidth]
  {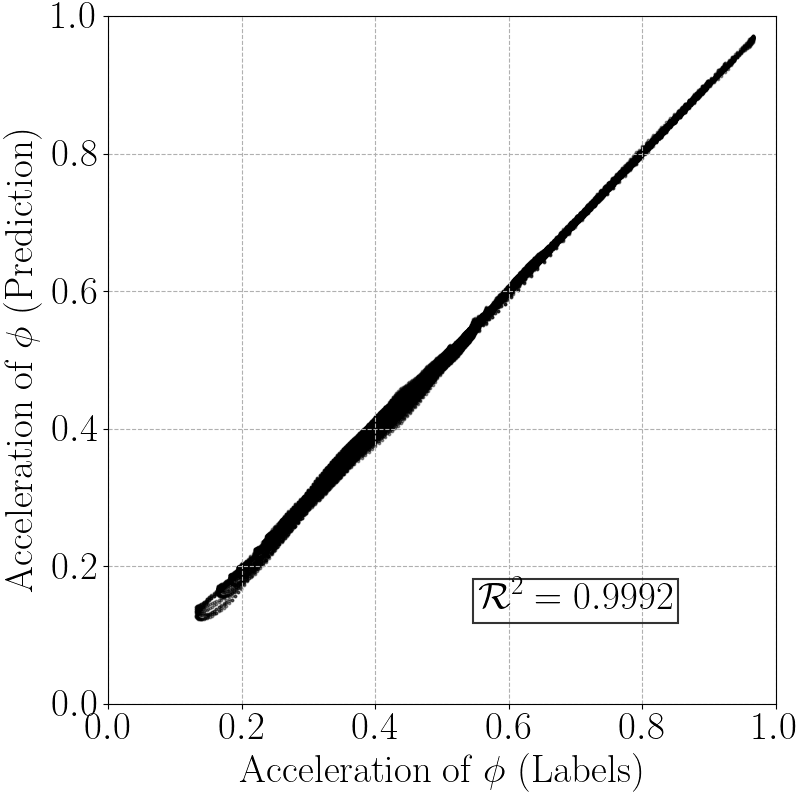}
  \includegraphics[width = \figsize\textwidth]
  {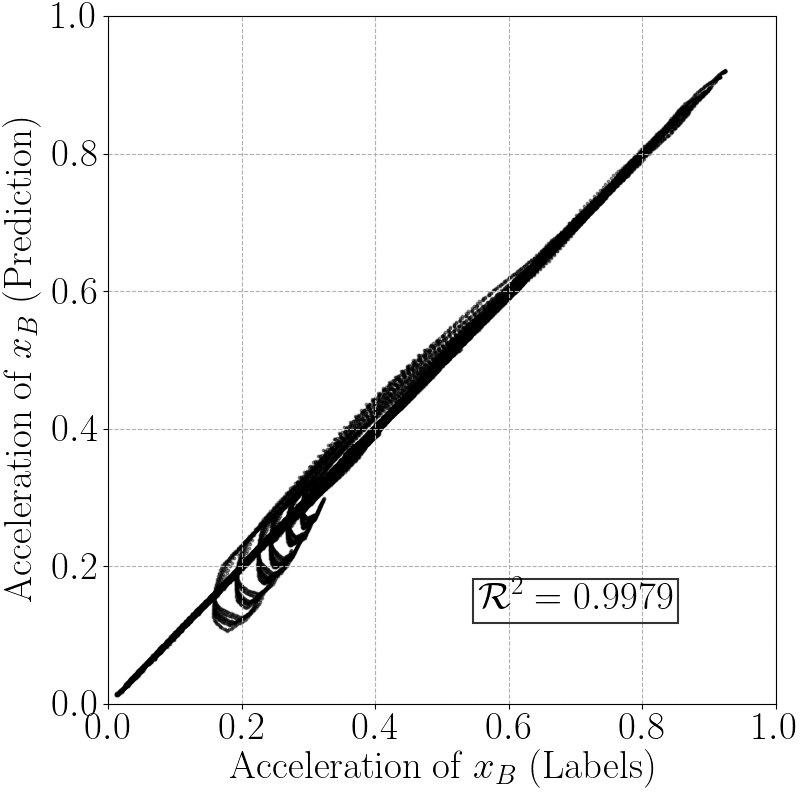}
\caption{
Labels vs. Predictions for normalized test data. The meta-model for the slider crank problem is generated from \Sfull\, type of training set. Test data are {\it{unseen}} from training. The $\R$ scores are almost 1, which implies that the DNN model predicts output solutions with high accuracy.
}
\label{fig:slider:scatter}
\end{figure}

The scatter plots in Fig. \ref{fig:slider:scatter} compares labels and predictions of the meta-model from \Sfull, and verifies that the meta-model produces almost accurate results. Its performance is much better than the other meta-models of previous examples, which seems to be caused by a simple form of kinematic equations \eqref{eq:slider:governing} and a sufficient training data set. The $\R$ values are over $0.999$ for the kinematic responses $\theta$, $\phi$, $\ddot{\phi}$, $x_B$, and $\dot{x}_B$.

\begin{figure}
  \centering
  \includegraphics[width = \figsize\textwidth]
  {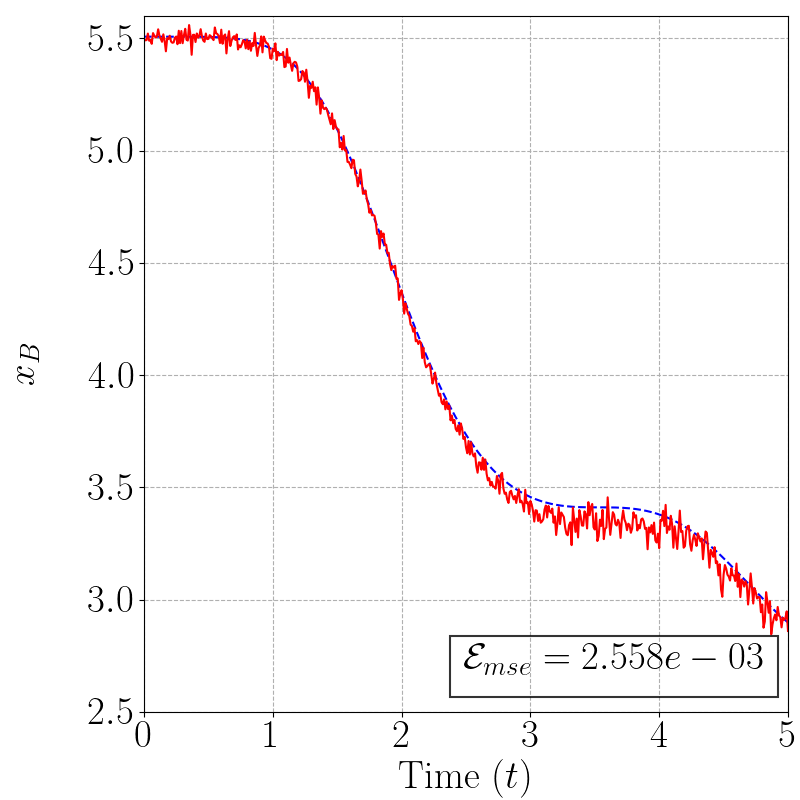}
  \includegraphics[width = \figsize\textwidth]
  {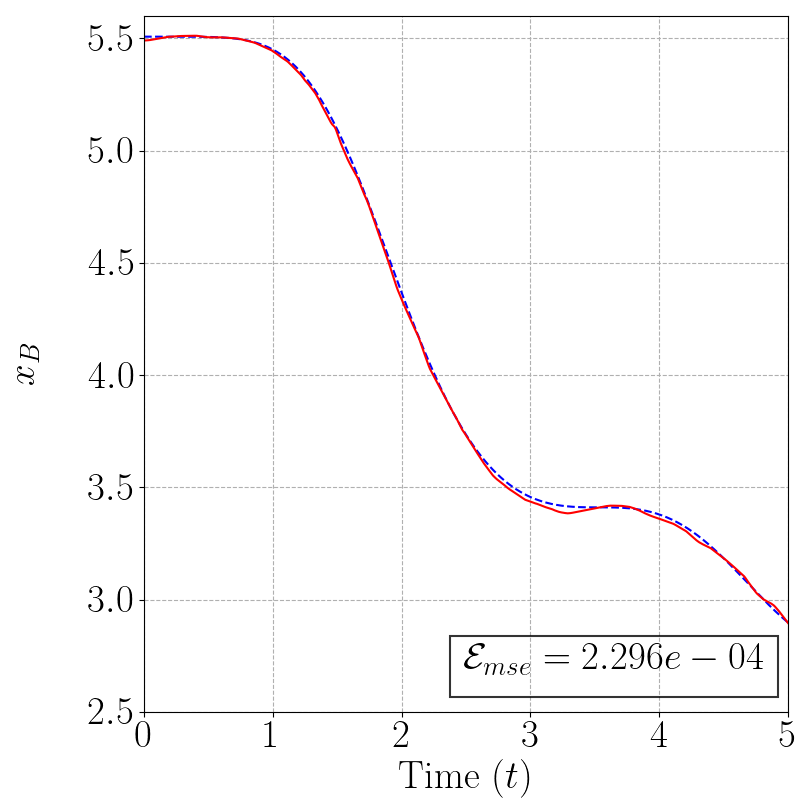}
  \includegraphics[width = \figsize\textwidth]
  {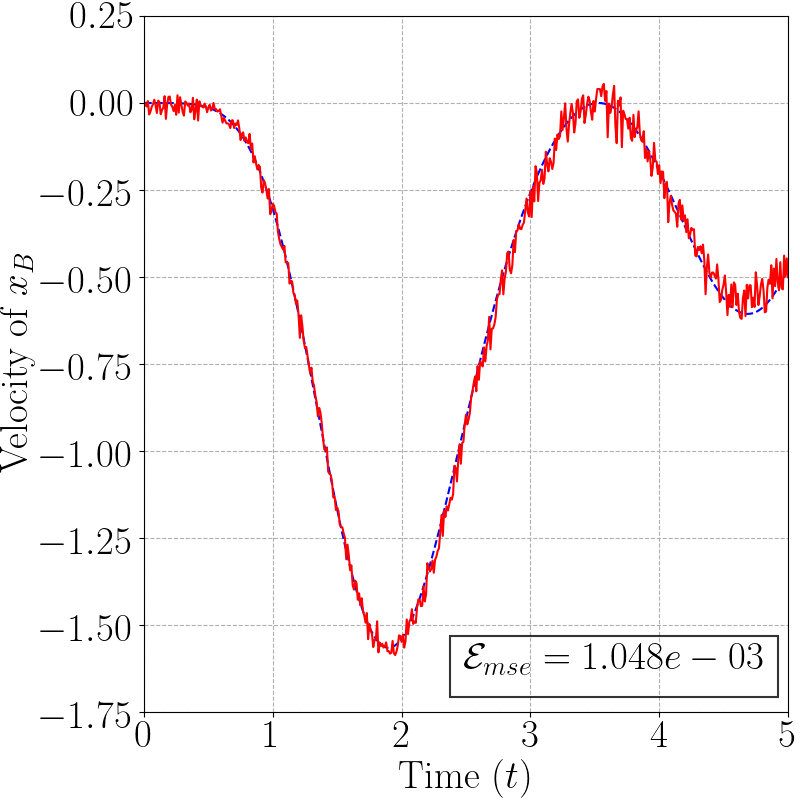}
  \includegraphics[width = \figsize\textwidth]
  {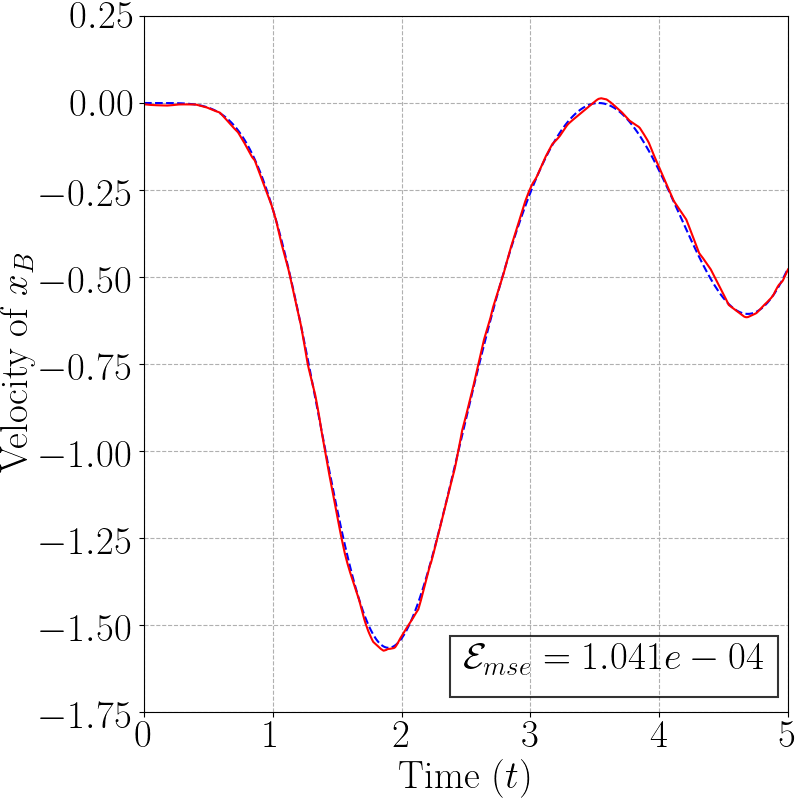}
  \includegraphics[width = \figsize\textwidth]
  {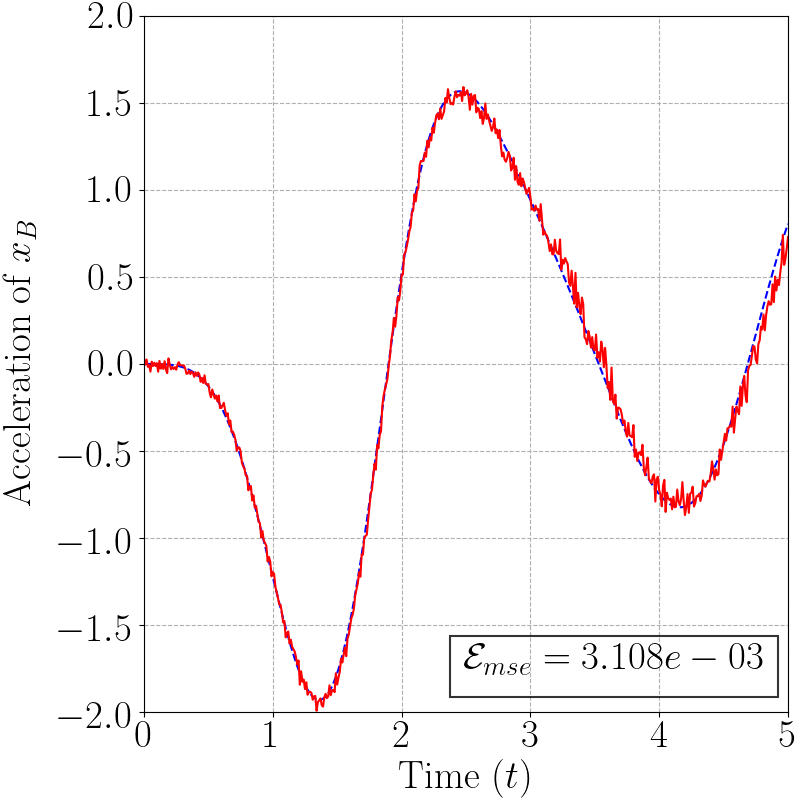}
  \includegraphics[width = \figsize\textwidth]
  {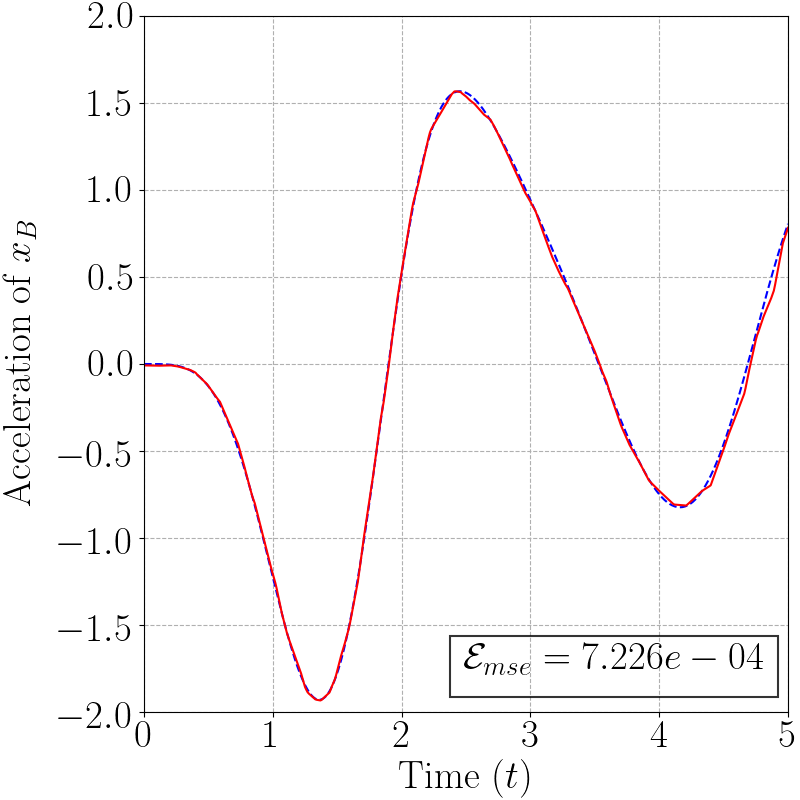}
\caption{
Dynamic responses of slider crank: Labels(blue dashed) vs. Predictions(red solid) for specific input $\tau=1.780$, $r=1.360$, and $L/r=3.050$. Left:$\#\{t_n\}$ numbers of meta-models are generated for each fixed time $t = t_n$ (\Sfixed).  Some oscillations are observed. Right:When time variable $t$ is considered as an input parameter (\Sfull). Relatively smooth solutions are achieved.
}
\label{fig:slider:specific}
\end{figure}

\begin{figure}
  \centering
  \includegraphics[width = \figsize\textwidth]
  {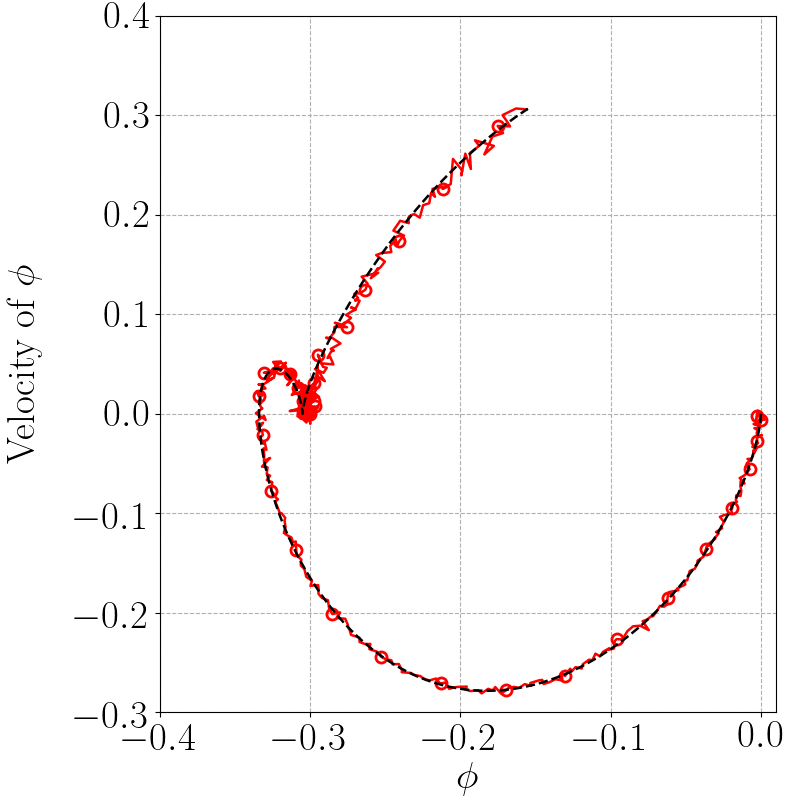}
  \includegraphics[width = \figsize\textwidth]
  {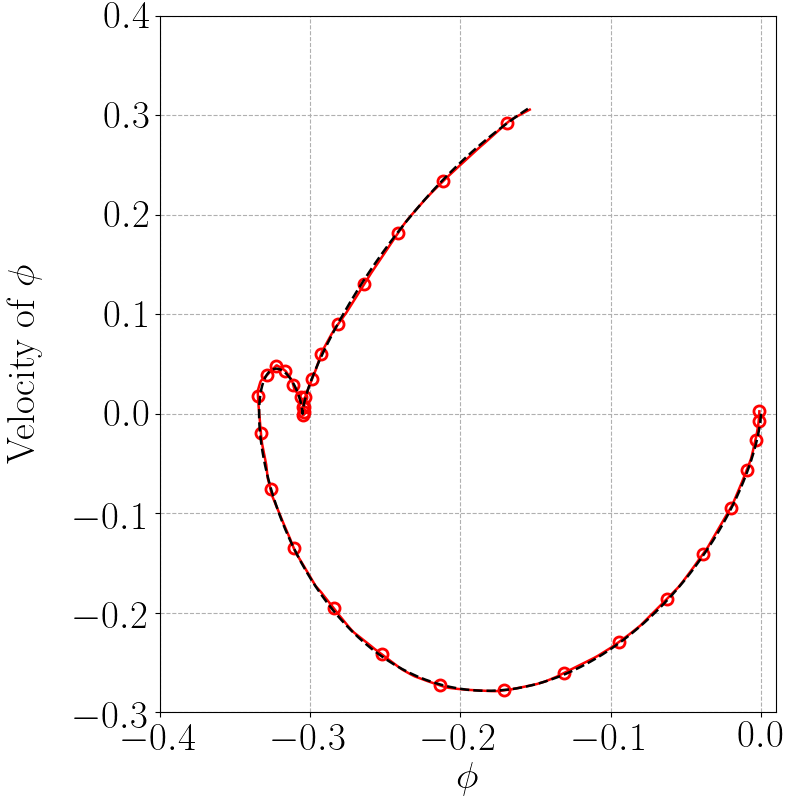}
  \includegraphics[width = \figsize\textwidth]
  {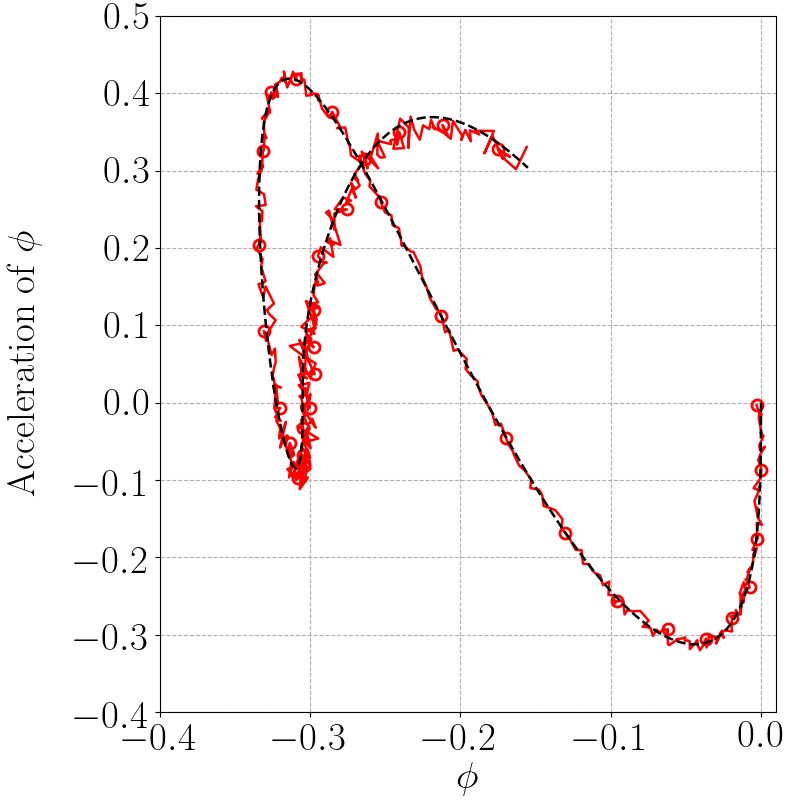}
  \includegraphics[width = \figsize\textwidth]
  {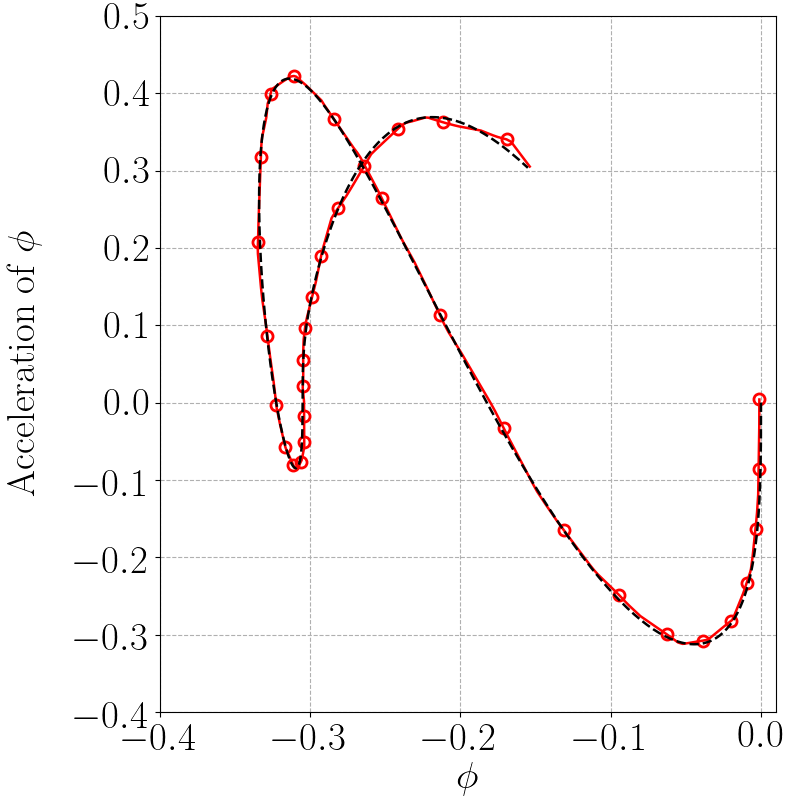}
\caption{Relations between dynamic responses of slider crank problem when $\tau=1.780$, $r=1.360$, and $L/r=3.050$: Labels(black dashed) and predictions (red solid, circles) are given. Results from different types of training data set \Sfixed (Left) and \Sfull (Right) are compared. \Sfull\, yields more smooth and accurate dynamic results.  
}
\label{fig:slider:traj:phi}
\end{figure}

\begin{figure}
  \centering
  \includegraphics[width = \figsize\textwidth]
  {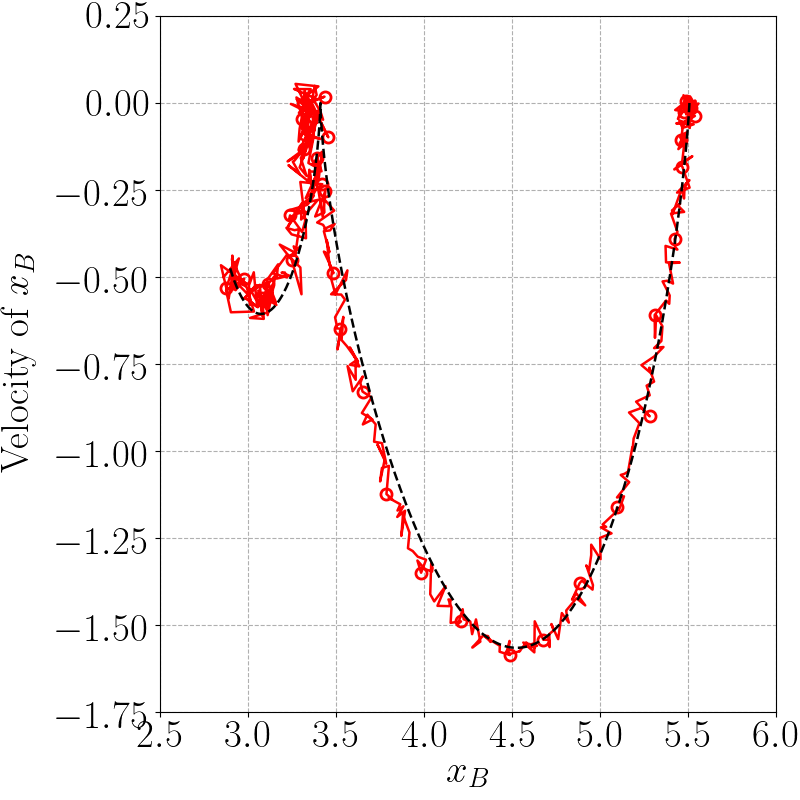}
  \includegraphics[width = \figsize\textwidth]
  {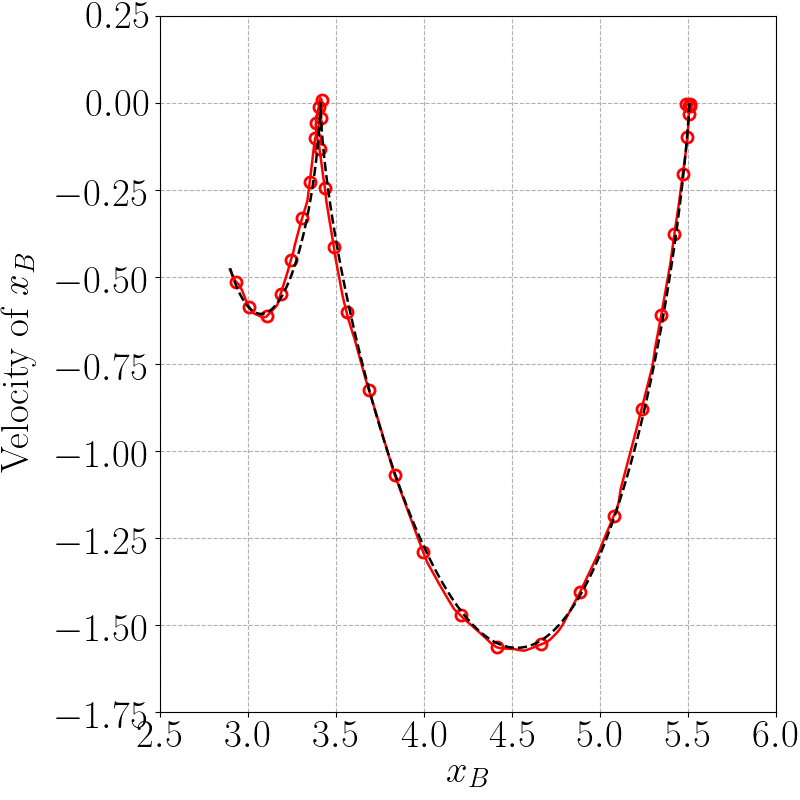}
  \includegraphics[width = \figsize\textwidth]
  {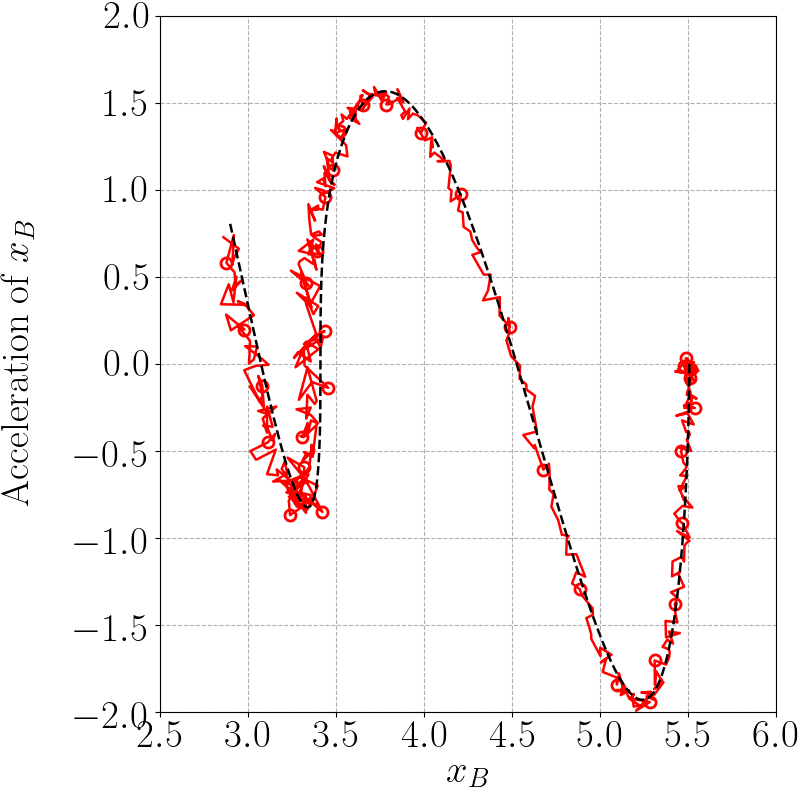}
  \includegraphics[width = \figsize\textwidth]
  {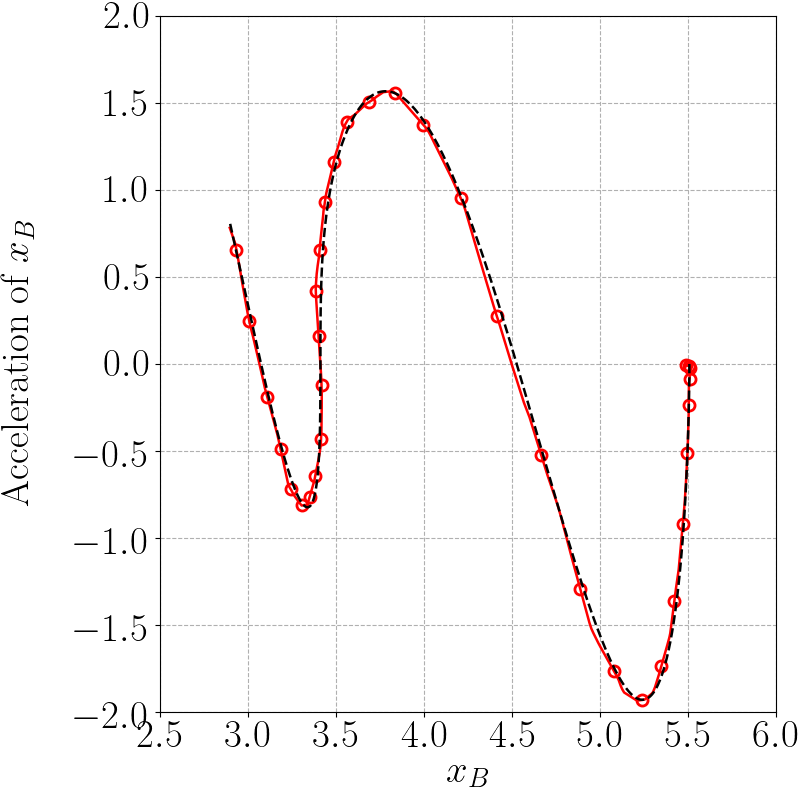}
\caption{Relations between dynamic responses of slider crank problem when $\tau=1.780$, $r=1.360$, and $L/r=3.050$: Labels(black dashed) and predictions (red solid, circles) are given.
Results from different types of training data set \Sfixed (Left) and \Sfull (Right) are compared. \Sfull\, yields more smooth and accurate dynamic results.  
}
\label{fig:slider:traj:xB}
\end{figure}

Since the predictions for test data are highly accurate as confirmed in Fig. \ref{fig:slider:scatter},  Fig. \ref{fig:slider:specific}, \ref{fig:slider:traj:phi}, and \ref{fig:slider:traj:xB} present results only for a specific case of test data: $\tau=1.780, r=1.360, L/r=3.050$. 
Fig. \ref{fig:slider:specific} shows changes of translation and velocities of the slider mass $B$ in time $t$. As shown in previous Sections \ref{sec:single} and \ref{sec:double}, \Sfixed (Left) shows oscillatory waves, while \Sfull\, yields smooth solutions. The error $\MSE$ compares the difference of their accuracies more clearly. 

Fig. \ref{fig:slider:traj:phi} displays time-varying relations between the angle of connecting rod $\phi(t)$ and its temporal derivatives $(\dot{\phi}(t), \ddot{\phi}(t))$. The oscillations from the case of \Sfixed\, (Left) are observed. Fig. \ref{fig:slider:traj:xB} shows relations between the displacement of slider $x_B$ and its derivatives. Performance of two training data set \Sfixed\, (Left) and \Sfull\, (Right) is more clear than Fig. \ref{fig:slider:traj:phi}.  \Sfull\, yields more smooth and accurate results than \Sfixed. 

\begin{table*}
\centering \ra{1.1}
\begin{tabular}{@{}lllll@{}}
\toprule
& Parameters
& Ranges
& \begin{tabular}[c]{@{}c@{}}Meshsizes for \\ Training Data\end{tabular} 
& \begin{tabular}[c]{@{}c@{}}Meshsizes for \\ Test Data\end{tabular} 
\\
\midrule
Fixed constants
& $\theta^0[rad]$ & $0$ & $\cdot$ & $\cdot$\\
Inputs
& $\tau$ & $[1, ~2]$ & $\Delta{\tau}=0.1$ & 
arbitrary(not uniform)\\
& $r[m]$ & $[1, ~3]$ & $\Delta{r}=0.2$ & 
arbitrary(not uniform)\\
& $L/r$ & $[2.5, ~3.5]$ & $\Delta{(L/r)}=0.1$ & 
arbitrary(not uniform)\\
Time instants 
& $\{t_n\}[s]$ & $[0, 5]$ & $\Delta{t}=0.01$\,$(t_0 = 0)$ & $\Delta{t}=0.01$\,$(t_0 = 0)$ \\
\bottomrule
\end{tabular}
\caption{Summary on parameters of slider crank problem. In \Sfixed, a fixed time instant is considered. In \Sfull, all the time instants are treated as inputs. }
\label{tab:slider:param}
\end{table*}

\section{Conclusions}
\label{sec:conclusions}
The present study introduces a procedure to combine a machine learning and solution of general purpose multibody dynamics. The paper contributes to data-driven modeling for multibody systems in two meaningful aspects. 
The first is that Deep Neural Network learning is applied, not to a specified particular type, but a {\it{general}} multibody dynamic problem. 
The generality makes it possible for the proposed DNN algorithm to be employed for other multibody system problems in future research. 
The second is that the present work analyzes and suggests how training data need to be structured for more effective DNN learning. In particular, it is found out that treating time variable as an input parameter enhances accuracy and smoothness of resulting predictions. The observation is worthwhile to notice, since the smoothness of physical variables in time direction is significant in dynamic problems. 
The paper demonstrates that the accurate solution of general purpose multibody dynamics can be achieved by DNN procedure. Despite the introduced numerical results, the present data-based learning algorithm can be improved through further studies. For one thing, performing {\it{smart sampling}} which decides more suitable ranges and non-uniform mesh sizes of data will improve computational efficiency in generating a meta-model. 
Moreover, to make fundamental progress in data-driven design of MBD, 
further studies are required on other various subjects, from theories on probability, uncertainties, and physics, to brand-new data-handling techniques.

\begin{acknowledgements}
This research is supported by 2018-2019 KyungHee University Research Support Program. 
\end{acknowledgements}




\begin{thebibliography}{}

\bibitem{KinBa14}
Kingma, Diederik P., and Ba, Jimmy Lei, 
Adam: A method for stochastic optimization, 
arXiv:1412.6980v9, 
(2014)

\bibitem{RumHinWil86}
Rumelhart, David E., Hinton, Geoffrey E., Williams, Ronald J.,
Learning Representations by Back-propagating Errors, 
Nature, 323, 533--536, (1986)

\bibitem{Hin12}
Hinton, G. 
Neural Networks for Machine Learning - Lecture 6a - Overview of mini-batch gradient descent, 
(2012)


\bibitem{Lanz06}
Lanz, O.,
Approximate Bayesian Multibody Tracking,
IEEE TRANSACTIONS ON PATTERN ANALYSIS AND MACHINE INTELLIGENCE, 28, 9, 
(2006)

\bibitem{Sha05}
Shabana, A.A., Dynamics of Multibody Systems, Cambridge University Press, Cambridge (2005)

\bibitem{PonAmoBalPaiFer16}
Pontes, F. J., Amorim, G. F., Balestrassi, P. P., Paiva, A. P., Ferreira, J. R.,  
Design of experiments and focused grid search for neural network parameter optimization, 
Neurocomputing, 186, 22-34, 
(2016)

\bibitem{HuaLeeLinHua07}
Huang, C. M., Lee, Y. J., Lin, D. K., Huang, S. Y.,  
Model selection for support vector machines via uniform design, 
Computational Statistics $\&$ Data Analysis, 52(1), 335-346, 
(2007)


\bibitem{FeuEggKalLinHut18}
Feurer, M., Eggensperger, K., Falkner, S., Lindauer, M., Hutter, F, 
Practical automated machine learning for the automl challenge 2018, 
In International Workshop on Automatic Machine Learning at ICML (pp. 1189-1232), 
(2018, July)


\bibitem{GodBenCou16}
Goodfellow, I., Bengio, Y., Courville, A, 
Deep learning, 
MIT press, 
(2016)

\bibitem{BerBen12}
Bergstra, J., Bengio, Y, 
Random search for hyper-parameter optimization, 
Journal of Machine Learning Research, 
13(Feb), 281-305, 
(2012) 

\bibitem{LeBen15}
LeCun, Y., Bengio, Y., Hinton, G, 
Deep learning, 
Nature, 521(7553), 436, 
(2015)


\bibitem{Dom12}
Domingos, P. M., 
A few useful things to know about machine learning, 
Commun. acm, 55(10), 78-87, 
(2012)

\bibitem{BlaTorGim15}
Blanco-Claraco, J.L., Torres-Moreno, J.L., 
Giménez-Fernández, A., 
Multibody dynamic systems as Bayesian networks:
Applications to robust state estimation of mechanisms, 
Multibody Syst Dyn, 34, 103-128,
(2015)

\bibitem{LiWuTedTenTor19}
Li, Y., Wu, J., Tedrake, R., 
Tenenbaum, J.B., Torralba, A., 
Learning particle dynamics for manipulating rigid bodies, deformable objects, and fluids, 
ICLR 2019.


\bibitem{TinMisPetSch07}
Ting, J-A.,  Mistry, M., Peters, J., Schaal, S., Nakanishi, J., 
A Bayesian Approach to Nonlinear Parameter Identification for Rigid Body Dynamics, 
Robotics: Science and Systems II., 
247-254, 
(2007)


\bibitem{TutBroWan12}
Tutsoy, O., Brown, M., Wang, H., 
Reinforcement learning algorithm application and multi-body system design by using MapleSim and Modelica, 
International Journal of Advanced Mechatronic Systems, 
(2012)

\bibitem{LinHafQueFre10}
Lin, Y-C., Haftka, R.T., Queipo, N.V., Fregly, B.J., 
Surrogate articular contact models for computationally efficient multibody dynamic simulations, 
Medical Engineering and Physics, 
32, 6, 584-594,
(2010)


\bibitem{HalErdBog09}
Halloran, J.P., Erdemir, A., van den Bogert, A.J., 
Adaptive Surrogate Modeling for Efficient Coupling of Musculoskeletal Control and Tissue Deformation Models, 
J. Biomech. Eng., 
131(1) 
(2009)

\bibitem{AnsTupDatNeg18}
Ansari, H., Tupy, M., Datar, M., Negrut, D., 
Construction and Use of Surrogate Models for the
Dynamic Analysis of Multibody Systems, 
SAE International by Columbia Univ, 
(2018)

\bibitem{KraCauMar18}
Kraft,S., Causse, J., Martinez, A., 
Black-box modelling of nonlinear railway vehicle
dynamics for track geometry assessment using
neural networks,
International Journal of Vehicle Mechanics and Mobility, 
(2018)


\bibitem{FalMalMel11}
Falomi, S., Malvezzi, M., Meli, E., 
Multibody modeling of railway vehicles: Innovative algorithms for the detection
of wheel-rail contact points, 
Wear, 
271, 453-461, 
(2011)


\bibitem{MarZaaWhiTaj07}
Martin, T.P., Zaazaa, K.E., Whitten, B., Tajaddini, A.,
USING A MULTIBODY DYNAMIC SIMULATION CODE WITH NEURAL NETWORK TECHNOLOGY TO PREDICT RAILROAD VEHICLE-TRACK INTERACTION PERFORMANCE IN REAL TIME, 
Proceedings of the ASME 2007 International Design Engineering Technical Conferences $\&$ Computers and Information in Engineering Conference, 
(2007)


\bibitem{ByrFox17}
Byravan, A., Fox, D., 
SE3-nets: Learning rigid body motion using deep neural networks, 
IEEE International Conference on Robotics and Automation (ICRA), 
(2017)



\end{thebibliography}



\end{document}